\def\year{2022}\relax
\documentclass[letterpaper]{article} % DO NOT CHANGE THIS
\usepackage{aaai22}  % DO NOT CHANGE THIS
\usepackage{times}  % DO NOT CHANGE THIS
\usepackage{helvet}  % DO NOT CHANGE THIS
\usepackage{courier}  % DO NOT CHANGE THIS
\usepackage[hyphens]{url}  % DO NOT CHANGE THIS
\usepackage{graphicx} % DO NOT CHANGE THIS
\usepackage{natbib}  % DO NOT CHANGE THIS AND DO NOT ADD ANY OPTIONS TO IT
\usepackage{caption} % DO NOT CHANGE THIS AND DO NOT ADD ANY OPTIONS TO IT
\DeclareCaptionStyle{ruled}{labelfont=normalfont,labelsep=colon,strut=off} % DO NOT CHANGE THIS
\usepackage{algorithm}
\usepackage{algorithmic}

\usepackage{newfloat}
\usepackage{listings}
\floatstyle{ruled}
\newfloat{listing}{tb}{lst}{}
\floatname{listing}{Listing}
\usepackage{times}
\usepackage{latexsym}

\usepackage[T1]{fontenc}
\usepackage[utf8]{inputenc}

\usepackage{subcaption}
\usepackage{graphicx}
\usepackage{natbib}
\usepackage{caption}
\usepackage{microtype}
\usepackage{booktabs}
\usepackage{amsmath}
\usepackage{amssymb}
\usepackage{soul, color}
\usepackage{xspace}
\usepackage{pifont}% http://ctan.org/pkg/pifont
\usepackage{hyperref}
\usepackage{nameref}
\usepackage{multirow}

\newcommand{\cmark}{\ding{51}}%
\newcommand{\xmark}{\ding{55}}%

\newcommand{\cam}{\textit{Coherence Aware Module}\xspace}
\newcommand{\cmcm}{\textit{Cross-Modal Coherence Model}\xspace}
\newcommand{\CMCM}{\texttt{CMCM}\xspace}
\newcommand{\cmca}{\textit{Cross-Modal Coherence Agnostic model}\xspace}
\newcommand{\CMCA}{\texttt{CMCA}\xspace}

\newcommand{\eg}{\textit{e.g., }}
\DeclareMathOperator{\ImgEnc}{E_{I}}
\DeclareMathOperator{\ImgFeat}{f_{I}}
\DeclareMathOperator{\TxtEnc}{E_{S}}

\title{Cross-Modal Coherence for Text-to-Image Retrieval}
\author{\textbf{Malihe Alikhani$^1$\thanks{These authors contributed equally. Author names ordered alphabetically based on last name.} \hspace{0.2in} Fangda Han$^2$\footnotemark[1] \hspace{0.2in} Hareesh Ravi$^2$\footnotemark[1] \hspace{0.2in} Mubbasir Kapadia$^2$ \hspace{0.2in} \\ Vladimir Pavlovic$^2$ \hspace{0.2in} Matthew Stone$^2$} \\\vspace{0.1in} $^1$University of Pittsburgh \hspace{0.2in} $^2$Rutgers University \\ \vspace{0.1in} \small{malihe@pitt.edu, fh199@rutgers.edu, hr268@scarletmail.rutgers.edu}, \{ mk1353, vladimir, mdstone\}@cs.rutgers.edu}

\begin{document}

\maketitle

\begin{abstract}
Common image-text joint understanding techniques presume that images and the associated text can universally be characterized by a single implicit model. However, co-occurring images and text can be related in qualitatively different ways, and explicitly modeling it could improve the performance of current joint understanding models. In this paper, we train a \cmcm for text-to-image retrieval task. Our analysis shows that models trained with image--text coherence relations can retrieve images originally paired with target text more often than coherence-agnostic models. We also show via human evaluation that images retrieved by the proposed coherence-aware model are preferred over a coherence-agnostic baseline by a huge margin. Our findings provide insights into the ways that different modalities communicate and the role of coherence relations in capturing commonsense inferences in text and imagery.

\end{abstract}

\section{Introduction}

When using text to retrieve an image, humans often rely on commonsense inference. 
Text and imagery can be related 
in obvious explicit ways, yet 
%with different kinds of implicit inferences.
the matter of a caption that accompanies an image frequently only indirectly overlaps with the content of the image. The text, for instance, can describe quantities that complement what is depicted in the image (e.g., add two cups of water) or a subjective reaction to what is depicted in an image (e.g., fantastic view). 
Retrieving imagery is therefore not just finding an image that portrays the text content but discovering an image that coherently fits with text to convey an integrated message.

These commonsense inferences can be modeled using representations and algorithms informed by approaches to natural language discourse, particularly coherence relations \cite{hobbs1985coherence,asher2003logics,taboada2006applications}. Coherence relations characterize the inferential links (such as temporal, causal, and logical) that connect the content of text and imagery.

Clues from text and from the typical relations of text and imagery provide important evidence about what kinds of visual content is coherent. Therefore, coherence agnostic methods don’t necessarily deliver images that fit naturally with text. By modeling coherence in text and imagery, we can supply images to text that human raters prefer by a large margin. This paper describes models of these broader associations between text and imagery for the task of image retrieval.

\begin{figure}[t]
\centering
    \caption*{{\textbf{Caption}: The start of the race.}}
    \begin{tabular}{cc}
         \includegraphics[width=0.45\linewidth, height=3cm]{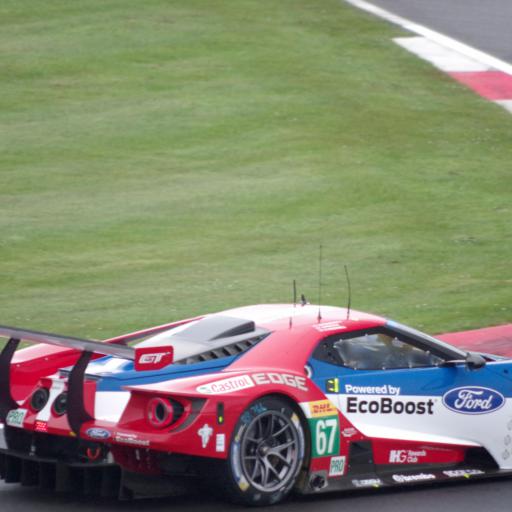}
         & \includegraphics[width=0.45\linewidth, height=3cm]{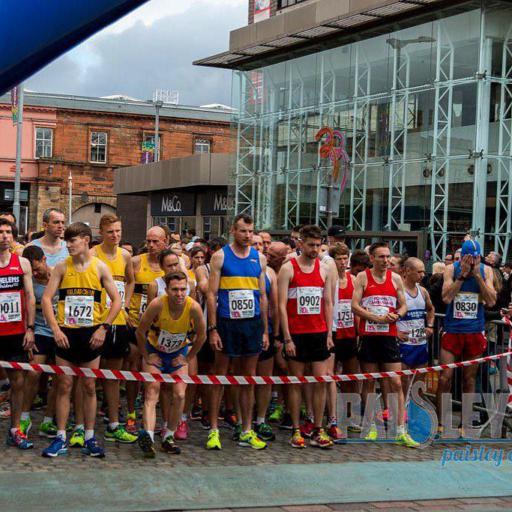} \\
         \large{\texttt{CMCA}} & \large{\texttt{CMCM}}
    \end{tabular}
    \caption{Example retrieved image by the proposed \cmcm (right) vs \cmca (left) for input caption (top).}
    \label{fig:intro_example}
    \vspace{-0.1in}
\end{figure}

We hypothesize that bringing in coherence relations \cite{alikhani2020cross} into the retrieval process, in contrast to personalities defined in \cite{personality}, should better improve the performance of text-to-image retrieval in a more generalizable way. We build on \citet{salvador2017learning} and \citet{chen2018deep} and introduce a new framework that integrates coherence relations in text-to-image retrieval task by extracting features for each modality separately then building a lower-dimensional common representation space. Our proposed framework introduces a \cam that learns to predict coherence relations that characterize an input image--text pair during training, and predictions from the module are applied during testing through a \emph{Selective Similarity Refinement} technique to further improve the retrieval performance. 
% This module helps the retrieval model focus on the intent of the users and the potential effects of image--text combination.

The examples of Figure~\ref{fig:intro_example} illustrate our approach. They contrast the output of our baseline \cmca (\texttt{CMCA}) taken from \citet{han2020cookgan} and that of the proposed \cmcm (\texttt{CMCM}) trained on image-text pairs with \emph{Story} coherence relations. We observe that the proposed \texttt{CMCM} provides more importance to the words \emph{the start} compared to \texttt{CMCA} that concentrates on visually grounded words like \emph{race}. Thus, \cam provides more interpretable and robust results, by virtue of explicitly modelling image-text coherence. During inference, the model leverages this knowledge to retrieve relevant images. We evaluate our systems on two image-text coherence datasets namely \textbf{CITE++} \cite{alikhani2019cite} and \textbf{CLUE} \cite{alikhani2020cross}. Each of these datasets correspond to different domains and are annotated with different coherence relations as shown in \autoref{tab:cite_questions} and \autoref{tab:clue_relations}. We also analyze the effect of each coherence relation in the datasets by modifying the \cam in the proposed \texttt{CMCM} model to detect only the presence of a single relation. These models $\texttt{CMCM}_c$ show which coherence relations improve/reduce the performance when compared with the baselines.\footnote{Code and data: \url{https://github.com/klory/Cross-Modal-Coherence-for-Text-to-Image-Retrieval}}

\section{Related Work}
%%%%%%%%%%%%%%%%%%%%%%%%%
Text-to-image retrieval models have been used in several multimodal NLP tasks and applications. \citet{saggion2003nlp} extract syntactic relations from captions for indexing and retrieving photographs of crime scenes. \citet{elliott-etal-2014-query} use image retrieval as a testbed for learning spatial relationships between image regions using Visual Dependency Representations. 
% Several works have shown that including images in information retrieval tasks such as document retrieval can improve the performance of the models \cite{funaki2015image,chowdhury2019know2look}.
% Most recent visual dialogue systems include image retrieval models to present images with text in response to user's needs and to better fulfill the dialogue goals \cite{zhang2019neural}. 
None of the previous works in this line have studied a discourse-aware approach for text-to-image retrieval which would best suit the context of the dialogue, inferences between text and imagery in multimodal documents, and the role of coherence in learning better models of image-text alignments.

% The majority of the previously proposed techniques propose a two-stage framework (e.g., \cite{frome2013devise}). In the first step, features for each modality are extracted separately, and then a lower-dimensional common representation space is built using canonical correlation analysis \cite{shi2019image} or cosine similarity \cite{salvador2017learning, chen2018deep, han2020cookgan, ravi2018show}. These techniques assume a single overall relation over all image-text pairs. Although some works \cite{wang2019learning} leverage meta-tags to regularize features from different domains, they still do not explicitly model different coherence relations that characterize image-text pairs.

Inspired by the success of coherence theory that has been applied to other forms of multimodal communication such as gesture \cite{lascarides2009formal} and comics \cite{mccloud1993understanding}, \citet{alikhani2019cite, alikhani2020cross} characterized coherence relations in text and imagery. Examples of these relations include \textit{elaboration}, when the text include information that is not depicted in the image (e.g., leave it in the oven for 30 minutes) or \textit{subjective} when the text evaluates or reacts to the content of the image 
(e.g., a delicious pizza).
They evaluated the effectiveness of coherence relations on a controlled caption generation task.
% where an image and a coherence relation are given as input while the model generates a caption that adheres to both the image and the relation. 
We do not train a controllable model as we hypothesize that not all relations equally characterize the image and text in a pair. Though the relations are defined for joint image-text discourse, some coherence relations like ``Subjective'' in the CLUE dataset characterize how the caption relates to the image and not the other way around. Hence, conditioning image retrieval on the relation is not reasonable. The proposed method evaluates the effectiveness of coherence relations by comparing \texttt{CMCA} with \texttt{CMCM}. Note that the proposed \cmcm is not the same as in \citet{alikhani2020cross}. Instead, our model learns to predict the coherence relation during training. 

With the advent of the Transformer \cite{att-all} architecture, there have been large pretrained multimodal transformers \cite{lu2019vilbert, chen2020uniter, li2020oscar} that train on large datasets like MSCOCO and others on muiltiple joint image--text learning tasks such as cross modal retrieval. Though they obtain state of the art performance, they do not directly support the addition of our proposed \cam. We hence leave the exploration of such architectures for the proposed setting as future work.

% To the best of our knowledge, this is the first work that comprehensively analyzes the effect of coherence relations for image retrieval and trains text-to-image retrieval models with cross-modal coherence. 

%%%%%
%uncomment for image to text 
%%%%
% \textbf{Image-to-Text: }
% Image-to-text retrieval \cite{vendrov2015order} and generation are well studied in the literature \cite{xu2015show, updown, dense} using datasets like MSCOCO \cite{lin2014microsoft}. Other works like storytelling, consider generation of \emph{story-like} text for a sequence of images \cite{wang2018show, liu2017let, yu2017hierarchically}. These datasets and techniques consider only \emph{descriptive} or \emph{story} text associated with an image ignoring other types of associations. In \cite{personality, chunseong2017attend}, controlled caption generation task is proposed, where a \emph{personality} type or \emph{hashtags} are used to condition the caption generation.  However, these signals only characterize how the text relates to the image while the image-text joint coherence is ignored. \\
%%%%%%%%%%%%%%%%%%%%%%%%%
\section{Methodology} 
\label{sec:method}

\begin{figure}[th]
    \centering
    \includegraphics[width=0.99\linewidth]{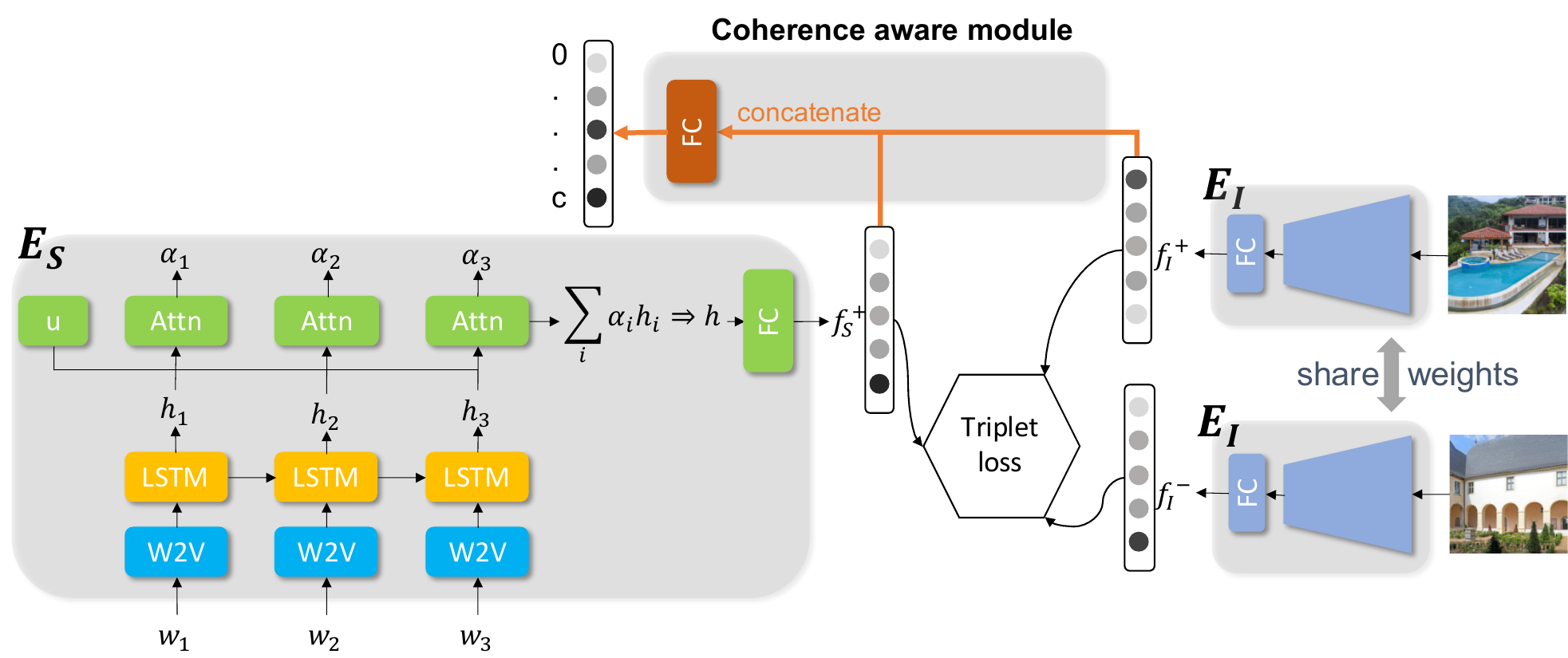}
    \caption{Framework of our proposed \cmcm. $\TxtEnc$ stands for text encoder, $\ImgEnc$ stands for image encoder, $\alpha_i$ stands for the attention of word embedding $h_i$}
    \label{fig:framework}
    \vspace{-0.2in}
\end{figure}

In this Section we describe the details of our proposed model. We argue that coherence relations characterize the data for multimodal discourse comprehension and hypothesize that a model with coherence (\texttt{CMCM}) will better retrieve relevant images compared to \texttt{CMCA}. \autoref{fig:framework} shows our framework for \texttt{CMCM} that consists of Image and Text Encoders that project the two modalitied onto a common embedding space optimized over cosine similarity, followed by a \cam that predicts the image-text coherence relations that characterize the input image-text pair. We show that addition of \cam regularizes the latent space and improves the performance of text-to-image retrieval by modelling the different coherence relations that characterize an image-text pair. To further explicitly use the predictions from \cam, we propose a Selective Similarity Refinement technique to refine and rank the retrieval result. 

To further analyze the performance of each coherence relation on the overall model, we train separate  $\texttt{CMCM}_{c}$ models that are aware of only one relation. The \cam is modified to predict only the presence of a particular relation $c$ (through binary classification in contrast to multi--label classification in the overall model) in these models.

% Further details about the model architecture are described in Sec. \ref{subsec:model_architecture}.

%In this section, we first describe the retrieval system (\autoref{fig:framework} consisted of an image encoder $E_I$, text encoder $E_s$ and the proposed \cam (CAM). CAM utilizes the additional coherence information to regularize the latent space. We continue elaborate Selective Similarity Refinement (SSR), which uses the confidence score of coherence prediction to further refine the retrieval result. We note our model as \cmcm (\CMCM) and the model without using coherence as \cmca (\CMCA).

\subsection{Model Architecture}
\label{subsec:model_architecture}
In order to train \texttt{CMCM} for text-to-image retrieval,% we leverage a state-of-the-art image retrieval model for recipes proposed in \citet{han2020cookgan}. 
we write $\mathbf{S} = [w_1, w_2, ..., w_m]$ for the input natural language text composed of $m$ words. (In principle $w_i$ could be words, phrases, sentences or any other semantic unit of text.) Similarly, we write $\mathbf{I}$ for the corresponding image. 
% Then the objective of an image retrieval model is to find the model parameter $\theta$, such that $\theta = \underset{\theta}{\arg\max}\, Pr_{\theta}(\mathbf{I} \mid \mathbf{S})$. 
Given text $\mathbf{S}_i$, the objective of an image retrieval model is to retrieve the paired image $\mathbf{I}_i$ from an image pool $\{\mathbf{S}_j\}, j \in [1,...,N]$, where N is the number of images in pool.

\noindent\textbf{Image Encoder} The image encoder $\ImgEnc$ is a pretrained Resnet-50 \cite{resnet} followed by a bottleneck layer to transform image features to the shared latent space. Each image is first resized to $224 \times 224$, and then forwarded through $\ImgEnc$ to get the image embedding $\ImgFeat \in \mathbb{R}^{300}$.

\noindent\textbf{Text Encoder} The text encoder $\TxtEnc$ starts from a pretrained word2vec model that embeds each word into a $300$ dimensional vector. 
The word2vec model is trained using Gensim \cite{rehurek_lrec}.
The maximum length of the text sequence considered is $200$ for CITE++ and $40$ for Clue based on the longest sentences in the dataset. 
% \begin{table*}[t]
%     \centering
%     \footnotesize
%     \begin{tabular}{ccp{0.55\textwidth}cc}
%     \toprule
%     Relation & Question & Description & Positive rate &  Entropy \\
%     \midrule
%     Expansion & Q2 & The image gives visual information about the step described in the text. & $0.821$ & $0.162$ \\
%     ImageNeeded & Q3 & You need to see the image in order to be able to carry out the step properly. & $0.115$ & $0.249$ \\
%     Elaboration$_{t}$&Q4 & The text provides speciﬁc quantities (amounts, measurements, etc.) that you would not know just by looking at the picture. & $0.329$ & $0.366$ \\
%     Elaboration$_{i-tool}$&Q5 & The image shows a tool used in the step but not mentioned in the text. & $0.193$ & $0.318$ \\
%     Temporal$_{i<t}$&Q6 & The image shows how to prepare before carrying out the step. & $0.158$ & $0.291$ \\
%     Temporal$_{i>t}$&Q7 & The image shows the results of the action that is described in the text. & $0.588$ & $0.312$ \\
%     Temporal$_{i=t}$&Q8 & The image depicts an action in progress that is described in the text. & $0.313$ & $0.364$ \\
%     \bottomrule
%     \end{tabular}
%     \caption{Coherence relations, their distribution and entropy in CITE++ dataset. We use the question identifier and the relation name interchangeably in the paper. \textit{Positive rate} is the percentage of samples that are labeled as `Yes' for that question. \textit{Entropy} is described in~\autoref{subsec:contribution_of_coherence_relations}.}
%     \label{tab:cite_questions}
% \end{table*}
\begin{table*}[t]
    \centering
    \footnotesize
    \begin{tabular}{ccp{0.55\textwidth}cc}
    \toprule
    Relation & Question & Description & Positive rate \\
    \midrule
    Expansion & Q2 & The image gives visual information about the step described in the text. & $0.821$  \\
    ImageNeeded & Q3 & You need to see the image in order to be able to carry out the step properly. & $0.115$  \\
    Elaboration$_{t}$&Q4 & The text provides speciﬁc quantities (amounts, measurements, etc.) that you would not know just by looking at the picture. & $0.329$  \\
    Elaboration$_{i-tool}$&Q5 & The image shows a tool used in the step but not mentioned in the text. & $0.193$  \\
    Temporal$_{i<t}$&Q6 & The image shows how to prepare before carrying out the step. & $0.158$   \\
    Temporal$_{i>t}$&Q7 & The image shows the results of the action that is described in the text. & $0.588$  \\
    Temporal$_{i=t}$&Q8 & The image depicts an action in progress that is described in the text. & $0.313$ \\
    \bottomrule
    \end{tabular}
    \caption{Coherence relations, their distribution and entropy in CITE++ dataset. We use the question identifier and the relation name interchangeably in the paper. \textit{Positive rate} is the percentage of samples that are labeled as `Yes' for that question}
    \label{tab:cite_questions}
    % \vspace{-0.2in}
\end{table*}
Then, the word embeddings are given as input to a Long Short Term Memory (LSTM) network to get each word representation.
We next apply an \textit{attention mechanism} \cite{att-all} to the LSTM representations, which learns the attention for each word and helps the model attend to key words that are important to our task.
Finally a fully-connected layer is applied to encode the joined representation of all words $h$ into the shared latent space.

The outputs of the text and image encoders are then used with a triplet objective using cosine similarity trained with hard negative mining. Hard negative mining targets on the most difficult negative image for each query in a batch based on the similarities to improve performance~\cite{hermans2017defense}. 
Let $s(a, b) = a^Tb / \sqrt{(a^Ta)(b^Tb)}$ measure the cosine similarity between two vectors $a$ and $b$, then the objective for the retrieval task per sample is given by \autoref{eq:triplet},
\begin{equation}
    \begin{aligned}
    \noindent trip\>(a, p, n)\> &= \>s(a,p)\>-\>s(a,n)-\>\alpha, \\
    \noindent \mathcal{L}_{ret} = &  \> \min\{0,\>trip(f_{S}^{+}, \>f_I^{+}, \>f_I^{-})\>\} \\
    + & \noindent\ \min \{0,\>trip(f_I^{+}, \>f_{S}^{+}, \>f_{S}^{-})\>\},
    \end{aligned}
    \label{eq:triplet}
\end{equation}
where $L_{ret}$ is the retrieval loss, $f_S^{+}$ and $f_I^{+}$ are outputs of text and image encoder for a pair of text and image while $f_S^{-}$ is a text output that does not correspond to current image and $f_I^{-}$ is an image output that does not correspond to current text. The margin $\alpha$ is set to $0.3$ by cross-validation.\\

\noindent \textbf{Coherence Aware Module}
Instead of relying only on the encoders, we also leverage coherence relations labelled by humans. We add a \cam that takes the normalized features from both text and image encoders as input and then passes them through a multi-layer perceptron to predict the relations.

The dimension of the final linear layer is equal to the number of relations in the dataset when trained with all relations (i.e. multi-label classification) and $1$ when trained with a single relation (i.e. single-label classification). We use Binary Cross Entropy (BCE) as the loss function and the objective of \cam for one sample is,
\begin{equation}
\begin{array}{l@{}l}
    \mathcal{L}_{cls} = \sum_{c} w_{c} \left( y_{c}\log(x_{c}) + (1 - y_{c})\log (1 - x_{c}) \right),
\end{array}
\label{eq:prediction}
\end{equation}
where $x_{c}$ is the probability assigned to relation $c$ by the model while $y_{c}$ is the ground truth binary value. Since the relations are not equally distributed in the dataset, we balance the training of different relations by giving a weight $w_{c}$ for each relation that is reciprocal to its proportion in the dataset. For $\texttt{CMCM}_{c}$ models, the summation is removed as there is only one relation that is predicted. 

The model is thus trained in a multi-task setting where the coherence predictor is the auxiliary task. 
The final objective over the entire batch with batch size $N$ is given in Equation \ref{eq:final},
\begin{equation}
    \mathcal{L}_{total} = \frac{1}{N} \sum_{n=1}^{N} \left( \mathcal{L}_{ret}^n + \lambda_{cls} \mathcal{L}_{cls}^n \right),
\label{eq:final}
\end{equation}
where $\lambda_{cls}$ is the weight associated with the coherence aware module and is chosen empirically as described later.

\subsection{Selective Similarity Refinement}
%In this section, we will first introduce how to use similarity to perform retrieval in testing, then define confidence score and illustrate how to apply it to refine the similarity, and finally shows a selective refinement technique that makes the refinement more effective.
%\noindent \textbf{Similarity}
The performance of the retrieval model depends on the similarities between a query caption $S_i$ and all possible images $\{I_j\}, j \in [1,...,N]$ (including the ground-truth image). We use cosine similarity (though any other valid similarity metric can be used) and notate the similarity between query $S_i$ and one image $I_j$ as $\theta_{i,j} = cosine(S_i, I_j)$. %, the similarity can be any possible similarity metric, in this paper, we use cosine similarity as shown below,
%\begin{align}
%    \theta_i = cosine(S, I_i)
%    \label{eq:similarity}
%\end{align}

%Ideally, we want the similarity between the query caption and the paired image to be the largest, however, due to the natural complexity of the two domains, limited data and model capacity, this may not always be true during testing.

\begin{figure}[h]
    \begin{minipage}[t]{.48\linewidth}
        \centering
        \includegraphics[width=\textwidth]{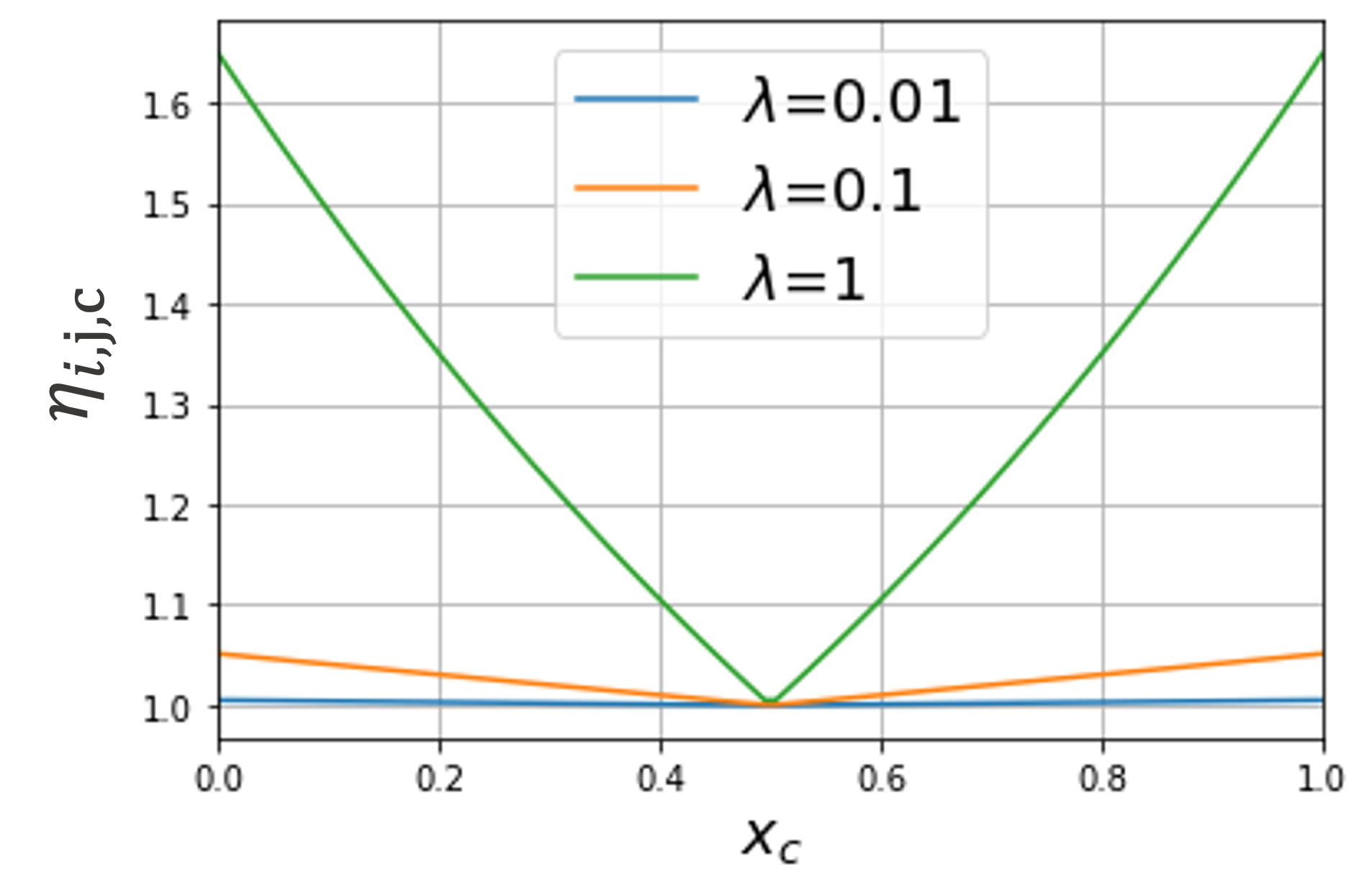}
        \captionsetup{width=\textwidth,font=footnotesize}
        \caption*{(a)}
    \end{minipage}
    \hfill
    \begin{minipage}[t]{.48\linewidth}
        \centering
        \includegraphics[width=\textwidth,height=3cm]{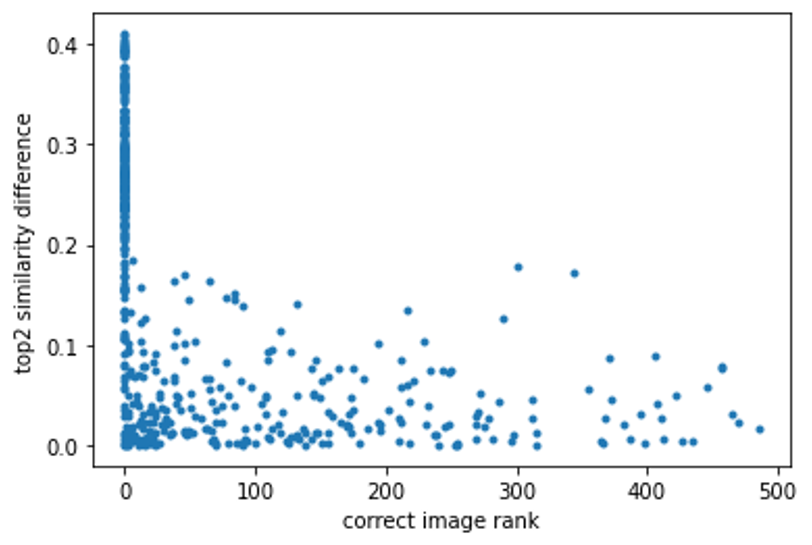}
        \captionsetup{width=\textwidth,font=footnotesize}
        \caption*{(b)}
    \end{minipage}
    \vspace{-0.1in}
    \caption{(a) Confidence function $\eta_{i,j,c}$ with different $\lambda$. (b) Correct image rank vs. the difference between the similarities of the top 2 retrieved images on CITE++ validation set}
    \label{fig:selective_similarity_refinement}
    \vspace{-0.2in}
\end{figure}

\noindent \textbf{Leveraging Confidence Score}
%We focus on using the prediction of coherence from \cam to refine the similarity. Note that we don't know the ground-truth coherence between the caption and image, but intuitively, if \cam is well trained, a query caption paired with the ground-truth image should predict coherence with more confidence. So the confidence function should return higher score when the predicted probability is close to zero or one, and lower score when the predicted probability approaches $0.5$. We tried different confidence function designs and find exponential function works best in the retrieval task, formally, for a query caption $S$ and one possible image $I_i$, the confidence function is computed as, 
We use the coherence prediction from \cam to refine the similarity between an image--text pair for retrieval during inference. Note that we do not know the coherence relation characterizing a ground truth image--text pair. However, a well trained \cam is expected to predict coherence for a ground truth image--text pair with high confidence. We define a confidence function for a query caption $S_i$ and one possible image $I_k$ as
%Note that we do not know the ground-truth coherence between the caption and image, but intuitively, if \cam is well trained, a query caption paired with the ground-truth image should predict coherence with more confidence. So the confidence function should return higher score when the predicted probability is close to zero or one, and lower score when the predicted probability approaches $0.5$. We tried different confidence function designs and find exponential function works best in the retrieval task, formally, for a query caption $S$ and one possible image $I_i$, the confidence function is computed as, 
\begin{align}
    \eta_{i,j,c} = e^{\lambda|x_{i,j,c}-0.5|}, \\
    \eta_{i,j} = \sum_c \eta_{i,j,c}, 
\end{align}
where $x_c$ is defined in \autoref{eq:prediction}, and $\lambda$ is a hyperparameter decided by cross validation. Confidence function with different $\lambda$ are shown in \autoref{fig:selective_similarity_refinement} (a). We can see that lower $\lambda$ decreases the impact of the confidence function. We set $\lambda=0.13$ for CITE++ and $\lambda=0.12$ for Clue datasets empirically. The refined similarity is defined as,
%Our experiments show that the confidence score only slightly impacts the performance
\begin{equation}
    \bar{\theta}_{i,j} = \theta_{i,j} * \eta_{i,j}
\label{eq:refined_similarity}
\end{equation}
%, we can see a lower $\lambda$ decreases the impact of the confidence function. 
\vspace{-0.2in}
\vspace{0.1in}

\noindent \textbf{Selective Refinement}
%During experiments we found although confidence score helps in general, it is still a much weaker indicator compared with similarity $\theta_i$, \eg retrieving using confidence score only gives result slightly better than random picking an image, base on this fact, we want to limit the usage of confidence score only on more difficult query captions.
Though confidence score helps, by itself the score is a weak indicator performing only slightly better than random. We hence limit the use of confidence score to difficult examples. We hypothesize that similarity between a correct image--text pair should on average be ``$\alpha$'' larger than that of a wrong image--text pair because of ~\autoref{eq:triplet}. In \autoref{fig:selective_similarity_refinement} (b), we verify this hypothesis by plotting the rank of ground truth image vs. the difference between the similarities of the top 2 retrieved images with the query caption. We observe that when the difference between the similarities of the top 2 images ($\Delta$) is large enough (\eg $\ge 0.2$), the retrieval is always successful (\eg ground truth image rank = 1). Based on this analysis, we select difficult query captions as those with $\Delta<T$, where $T$ is a hyperparameter chosen as $0.1$ empirically. We use the refined similarity \autoref{eq:refined_similarity} for ''difficult'' examples during inference.

\section{Image-Text Coherence Datasets}
\label{sec:dataset}
We study the efficacy of \texttt{CMCM} for image-retrieval by leveraging two image-text datasets CITE++ and Clue \cite{alikhani2020cross} that are annotated with image-text coherence relations. 
% We also propose an extended version of CITE dataset, CITE++, that adds $2242$ image-text pairs annotated with coherence relations.
CITE++ is extended by us from CITE~\cite{alikhani2019cite} adding $2242$ image-text pairs annotated with coherence relations.

%%%%%%figure
\begin{figure}[h]
    \begin{minipage}[t]{.48\linewidth}
        \centering
        \includegraphics[width=0.6\textwidth,height=3cm]{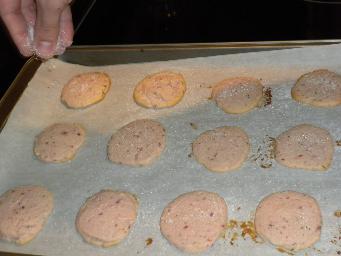}
        \captionsetup{width=\textwidth,font=footnotesize}
        \caption*{(a) Once they have baked remove them from the oven and sprinkle lightly with sugar. After you have dressed them allow them to cool for about 5 minutes and serve}
    \end{minipage}
    \hfill
    \begin{minipage}[t]{.48\linewidth}
        \centering
        \includegraphics[width=\textwidth,height=3cm]{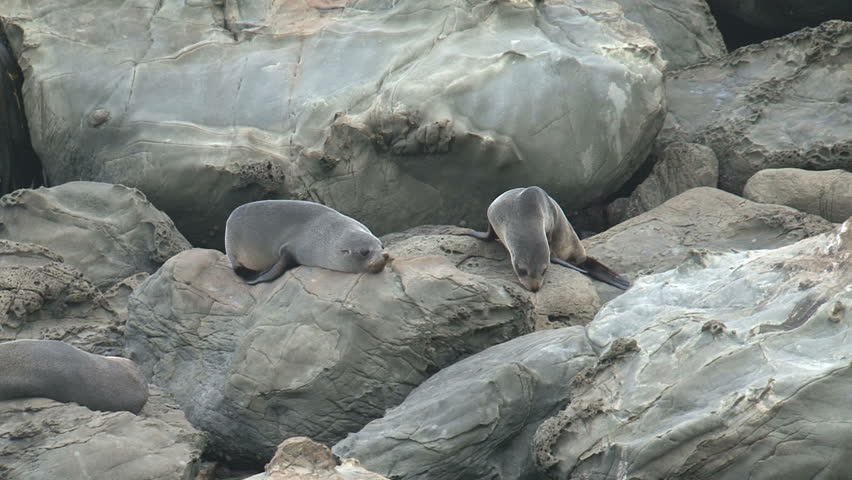}
        \captionsetup{width=\textwidth,font=footnotesize}
        \caption*{(b) Seals fighting for a spot to sleep on the rocks}
    \end{minipage}
    \caption{Example image-text pairs from CITE++ (a) and Clue (b) datasets. Image-text pair on the left has relations \emph{Expansion}, \emph{Elaboration} and \emph{Temporal$_{i>t}$} while the one on the right has relations  \emph{Action} as \emph{Visible}}
    \label{fig:dataset-examples}
    \vspace{-0.2in}
\end{figure}
\subsection{CITE++}
We extend the CITE dataset which is a subset of a popular recipe dataset RecipeQA \cite{yagcioglu-etal-2018-recipeqa}. The RecipeQA dataset consists of multimodal recipes that contains textual instructions accompanied by one or more images. CITE leveraged recipes that have one-to-one correspondence between instruction and image, \eg every instruction in the text has one image that visualizes it. Using Amazon Mechanical Turk, the authors obtained answers to $10$ questions that help characterize the relationship between image and text. We choose the questions that are best suited to train \texttt{CMCM} as described in \autoref{tab:cite_questions}. The original dataset has $2057$ image-text pairs annotated with True/False answers to these questions indicating presence/absence of the coherence relation. To perform a more comprehensive experiment, we collected $2242$ more pairs using the same annotation protocol, giving us a total of $4299$ image-text pairs. The distribution of relations in the entire dataset is given in \autoref{tab:cite_questions}. \autoref{fig:dataset-examples} [a] shows an example from CITE++ dataset.
% \begin{table}[h]
%     \centering
%     \footnotesize
%     \begin{tabular}{ccc}
%     \toprule
%     Relation & Positive rate &  Entropy \\
%     \midrule
%     Visible & $0.674$ & $0.266$ \\
%     Subjective & $0.066$ & $0.180$ \\
%     Action & $0.157$ & $0.290$ \\
%     Story & $0.243$ & $0.344$ \\
%     Meta & $0.391$ & $0.367$ \\
%     Irrelevant & $0.087$ & $0.212$ \\
%     \bottomrule
%     \end{tabular}
%     \caption{Coherence relations and their distribution in Clue dataset \cite{alikhani2020cross}}
%     \label{tab:clue_relations}
% \end{table}
\begin{table}[h]
     \centering
     \footnotesize
     \begin{tabular}{c|llllll}
     \toprule
     Relation & Visible & Subj. & Action & Story & Meta & Irr. \\
     \hline \\
     \% Positive & $67.4$& $6.6$ & $15.7$ & $24.3$ & $39.1$ & $08.7$ \\
     \bottomrule
     %\midrule
     %Visible &  \\
     %Subjective  \\
     %Action \\
     %Story  \\
     %Meta \\
     %Irrelevant \\
     %\bottomrule
     \end{tabular}
     \caption{Coherence relations (Subj. is Subjective and Irr. is Irrelevant) and their distribution in Clue dataset \cite{alikhani2020cross}}
    \label{tab:clue_relations}
    \vspace{-0.2in}
 \end{table}
\subsection{CLUE}
The Clue dataset \cite{alikhani2020cross} is constructed using the much larger Conceptual Captions dataset \cite{sharma-etal-2018-conceptual} which is primarily an image captioning dataset like COCO \cite{lin2014microsoft}.
Clue annotated $7559$ image-caption pairs with six coherence relations to summarize the structural, logical and purposeful relationships between the contributions of texts and images. 
Example image-caption pair with coherence relations are shown in \autoref{fig:dataset-examples} (b). %The distribution of relations in the Clue dataset is as follows: Visible: $0.674$,  Subjective: $0.066$, Action: $0.157$, Story: $0.243$,    Meta: $0.391$, Irrelevant: $0.087$.

% given in \autoref{tab:clue_relations}.

\section{Experimental Setup}
\label{sec:experiment}

\noindent \textbf{Network Details.} The backbone of image encoder $E_i$ is ResNet-50, with one additional batch normalization layer and one fully-connected layer to transform the feature into the shared space ($\mathbb{R}^{1024}$). The word2vec model encodes each word into a vector of $\mathbb{R}^{300}$, text encoder $E_t$ takes the vector as input and  forwards it through a bidirectional, one-layer LSTM module following an attention layer~\cite{att-all}, and finally the attention-weighted summation of word features is also transformed into the shared space ($\mathbb{R}^{1024}$) by a batch normalization layer and a fully-connected layer. \cam contains one fully-connected layer, adding more layers does not improve performance.

% Through \cam we propose to use coherence relations during training so that models are knowledgeable about other types of relation that can exist for the same image-text pair essentially regularizing the joint learned space. In Fig. \ref{fig:intro_example}, the \texttt{CMCA} models retrieves the other image as there are more images of \lq{}races\rq{} in general than there are of \lq{}start of a race\rq{} in the dataset. Also, the text \lq{}start of a race\rq{} communicates a story rather than factually describe elements in an image. Existing coherence agnostic models ignore the different characteristics by which an image-text pair can be related thereby producing the most commonly found semantically similar image. We address this concern using \cam. We could also perform \lq{}controlled\rq{} image retrieval (using the relation during inference) as done in \cite{alikhani2020cross}. However, this is ill-defined as some relations like \lq{}subjective\rq{} denote how text relates to image and not the reverse. In this section, we describe the metrics and baselines used as part of our experiments.

\noindent \textbf{Evaluation Metrics.} We evaluate the retrieval performance of all the models using the median retrieval rank (MedR) and the recall at K (R@K) metrics following existing works on text--to--image retrieval \cite{han2020cookgan, frome2013devise}. 
The retrieval range is set to be $500$. 
% We believe that a model trained with coherence relations would achieve higher Recall and lower Median Rank when compared with an agnostic model. 
% However, unlike datasets like MSCOCO where text descriptions are very specific to the paired image, both CITE++ and Clue have image-text pairs that exhibit complex relationships. 
% Given the diverse nature of image-text relations (e.g., \emph{Subjective} and \emph{Action}), we believe that the above quantitative metrics that measure if the model retrieves the original image is stricter. 
% Hence, we also perform comprehensive user study to evaluate the performance of the model. 
Since CITE++ and Clue have image-text pairs that exhibit complex relationships, we also perform a comprehensive user study to evaluate the performance of the model. 
\textbf{MedR} ($0\le$ MedR $\le 1$) is computed as the median rank of the true positive over all queries, a lower MedR suggests better performance. 
\textbf{R@K} ($0\le$ R@K $\le 100$) computes the percentage of true positives recalled among the top-K retrieved candidates, higher indicates better performance.
Here we only report the results of retrieving image by using the caption as query.

\noindent \textbf{Dataset Partition.} In our experiments, we evaluate the model and the coherence relations on CITE++ and Clue datasets independently. We split the CITE++ dataset as $3439/860$ for training/testing while the Clue dataset as $6047/1512$ for training/testing. 10\% of the training data is used as validation. Further training and hyperparameter details are given in the appendix.

\noindent \textbf{Comparative Evaluation.}
For both the datasets, we train the proposed model and compare with various baselines as shown in \autoref{tab:comparison}. The baseline \texttt{CMCA} \cite{han2020cookgan} is similar to existing CNN-RNN architectures such as \cite{xu2015show, ravi2018show, yang2020constrained}. Note we also compare with \texttt{CMCM}$_{c}$, which only uses one specific relation to train the system. We perform this experiment primarily to analyze the effect of each relation as not all relations contribute equally to the retrieval system. This also helps us better understand the influence of different relations on the proposed \cmcm model. Though it is possible to develop transformer based models for the proposed setting, we use GRUs and CNNs because of the low cardinality of the datasets and the necessity of large datasets for transformer based models \cite{inan-etal-2021-cosmic-coherence, ganesh2021compressing, crawford2021atlas}.

\begin{table}[h]
    \centering
    \scriptsize
    \begin{tabular}{lp{0.4in}cc}
    \toprule
     Model & \cam & Relations & Attention\\
    \midrule
     Base & \xmark & - & \xmark\\
     \texttt{CMCA}& \xmark & - & \cmark \\
     \texttt{CMCM}-NoAttn & \cmark & All & \xmark\\
%     \texttt{CMCM}-NoConf & \cmark & All & \xmark & \cmark \\
     \texttt{CMCM} & \cmark & All  & \cmark \\
     \texttt{CMCM}$_{c}$ & \cmark & $c$ & \cmark \\
    \bottomrule
    \end{tabular}
    \caption{Description of the models used for comparison. -NoAttn means removing the attention module from the proposed model.%, -NoConf means removing the confidence score refinement mentioned in \autoref{eq:refined_similarity}. 
    \lq{}All\rq{} relations indicate that the \cam is trained with all the relations in a multi-label multi-class setting. $c$ indicates only one relation is used with the \cam in a binary classification setting. }
    \label{tab:comparison}
    \vspace{-0.2in}
\end{table}

\section{Results and Discussion}
\label{sec:results}
\subsection{\texttt{CMCM} vs \texttt{CMCA}}
The results on CITE++ dataset are shown in \autoref{tab:cite_quantitative_evaluation}. As can be seen, having attention over the text clearly improves retrieval performance. This can be attributed to the lengthy texts in CITE++ dataset. Moreover, we observe that \texttt{CMCM} model performs better than \texttt{CMCA} and $Base$ across all metrics though with variable significance. For example, MedR for \texttt{CMCA} model is $5.4$ but all \texttt{CMCM} models achieve average MedR of less than $5.0$. Moreover the standard deviation is also lower indicating more robust performance. 
The results on the Clue dataset are given in \autoref{tab:clue_quantitative_evaluation}. We observe that both the attention mechanism and the coherence-aware module improve the performance. 
We use the example in Fig. \ref{fig:intro_example} to intuitively explain the effect of \cam. Note \texttt{CMCA} retrieves the incorrect image as there are more images of ``races'' in general than there are of ``start of a race'' in the dataset. Also, the text ``start of a race'' communicates a story rather than factually describe elements in an image, \CMCA ignores the different characteristics by which an image-text pair can be related thereby producing the most commonly found semantically similar image. \CMCM resolves this concern by considering coherence relations between the two modalities and retrieves the correct image.
We observe that all per-relation $\texttt{CMCM}_{c}$ models perform better than \CMCA. In some instants, per-relation models perform better than \texttt{CMCM}, confirming the conjecture that not all relations contribute in increasing the performance of the retrieval model. We perform additional analysis on per-relation contribution to \texttt{CMCM}s performance in the Appendix. 

%Moreover, we observe that \texttt{CMCM}$_{c}$ trained with one coherence relation improves the performance of the model by 1\% R@1, 3\% R@5 and 1\% R@10 on an average for all relations. 
%\cam increases the performance on retrieval compared with the baseline even without attention (cf. \texttt{CMCM}-NoAttn vs \texttt{CMCA}). However, having all the relations \texttt{CMCM} does not perform better than the variants with only one relation, which confirms the conjecture that not all relations contribute in increasing the performance of the retrieval model. 
%We believe this is because, recall metrics are stricter for both CITE++ and Clue datasets, since some captions can adequately be visualized by more than one image.

\begin{figure}
    \centering
    \includegraphics[width=\linewidth]{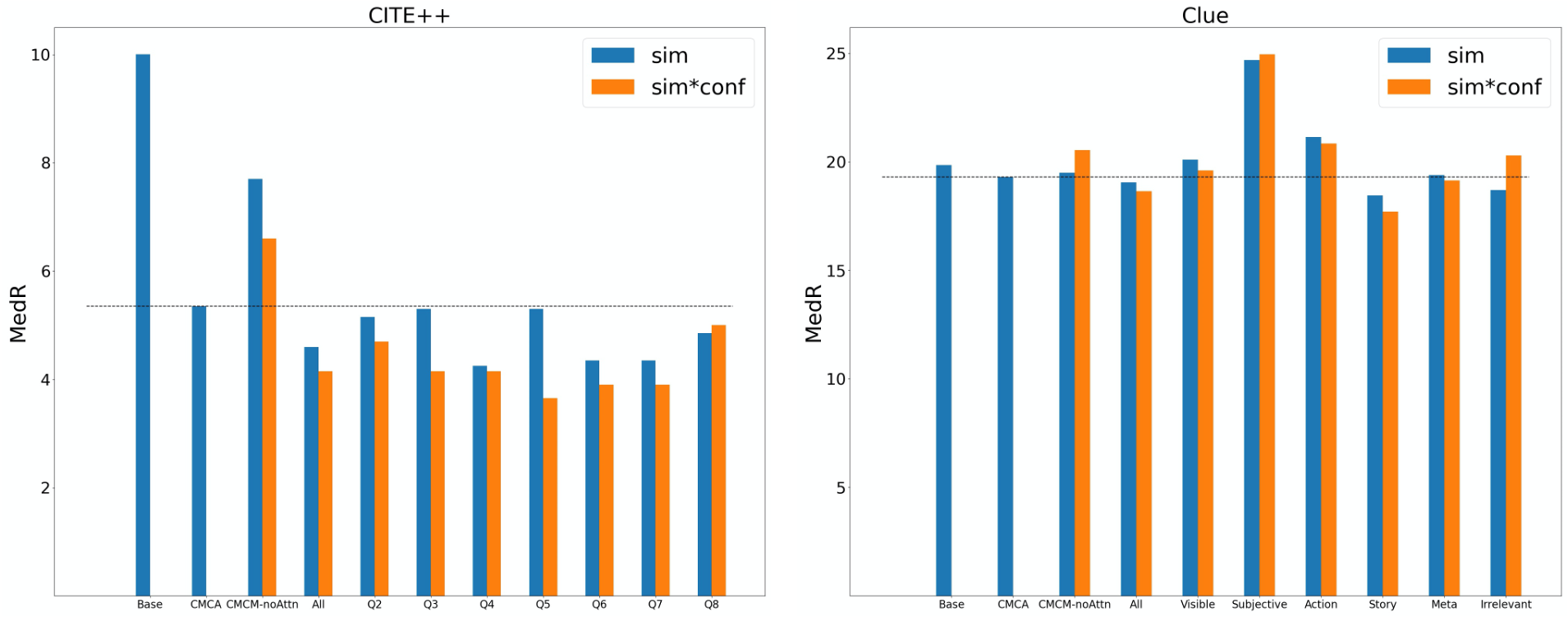}
    \caption{Comparison MedR between baseline, CMCA and different CMCM variants; as well as the comparison between the same model with and without selective similarity refinement. Left: CITE++ dataset. Right: Clue dataset}
    \label{fig:compare_medR}
\end{figure}

\noindent \textbf{Impact of Similarity Refinement.}
To evaluate the contribution of selective similarity refinement, we compare MedR based on $\theta_i$ and $\bar{\theta_i}$ of the same model in \autoref{fig:compare_medR}. The \texttt{CMCM} (except \lq{}NoAttn\rq{}) variants clearly outperform \texttt{CMCA} and baseline. Moreover, the selective refinement technique improves the result of almost all the \texttt{CMCM} models even further by a large margin as can be seen by the difference between the blue and orange bars.
In Clue dataset, in most cases, model using selective similarity refinement performs better than the same model without refinement, proving the effectiveness of the refinement technique. 
For $\texttt{CMCM}_{Irrelevant}$ model on Clue (last two bars on \autoref{fig:compare_medR} right), applying the refinement severely degrades the performance. We believe that \lq{}Irrelevant\rq{} relation does not effectively characterize the relationship between an image--text pair on top of its low positivity score.

\begin{table}[th]
    \footnotesize
    \centering
    \begin{tabular}{lccccccc}
    \toprule
      & MedR$\downarrow$ &  R@1$\uparrow$ & R@5$\uparrow$ & R@10$\uparrow$ \\
    \cmidrule{2-5}
    \texttt{Base} & $10.0^{\pm 3.7}$ & $45.7$ & $48.4$ & $50.6$ \\
    \texttt{CMCA}& $5.4^{\pm 2.3}$ & $46.0$ & $50.1$ & $53.8$ \\
    \texttt{CMCM}-NoAttn & $6.6^{\pm 2.5}$ & $46.0$ & $49.5$ & $52.0$ \\
    \texttt{CMCM} & $\mathbf{4.2^{\pm 1.2}}$ & $\mathbf{46.5}$ & $\mathbf{51.4}$ & $\mathbf{53.9}$ \\
    \midrule
    \texttt{CMCM}$_{Q_2}$ & $4.7^{\pm 2.0}$ & $46.4$ & $50.6$ & $53.4$ \\
    \texttt{CMCM}$_{Q_3}$ & $4.2^{\pm 1.3}$ & $46.2$ & $51.1$ & $54.2$ \\
    \texttt{CMCM}$_{Q_4}$ & $4.2^{\pm 1.3}$ & $46.2$ & $51.2$ & $54.2$ \\
    \texttt{CMCM}$_{Q_5}$ & $\mathbf{3.7^{\pm 1.3}}$ & $46.6$ & $\mathbf{51.5}$ & $\mathbf{54.4}$ \\
    \texttt{CMCM}$_{Q_6}$ & $4.6^{\pm 1.4}$ & $45.9$ & $50.8$ & $53.4$ \\
    \texttt{CMCM}$_{Q_7}$ & $3.9^{\pm 1.7}$ & $\mathbf{46.9}$ & $51.2$ & $54.1$ \\
    \texttt{CMCM}$_{Q_8}$ & $5.0^{\pm 1.7}$ & $46.4$ & $50.8$ & $53.8$ \\
    \bottomrule
    \end{tabular}
    \caption{Quantitative comparison in CITE++ dataset. The relations corresponding to each $Q_i$ are shown in \autoref{tab:cite_questions}. $\downarrow$ indicates that lower the better and $\uparrow$ indicates that higher the better.}
    \label{tab:cite_quantitative_evaluation}
    \vspace{-0.2in}
\end{table}

\begin{figure*}[t]
  \centering
  \footnotesize
  \begin{tabular}{cccccccccccc}
    %GT & Retrieved & GT & Retrieved\\ 
    GT & \multicolumn{5}{c}{\texttt{CMCM}} && \multicolumn{5}{c}{\texttt{CMCA}} \\
    \hline\\
    (a) \textcolor{blue}{Action}&\multicolumn{11}{c}{Horse grazing on a summer meadow in the forest outdoors.}\\ \\
    \begin{minipage}{0.08\textwidth}
      \includegraphics[width=\textwidth]{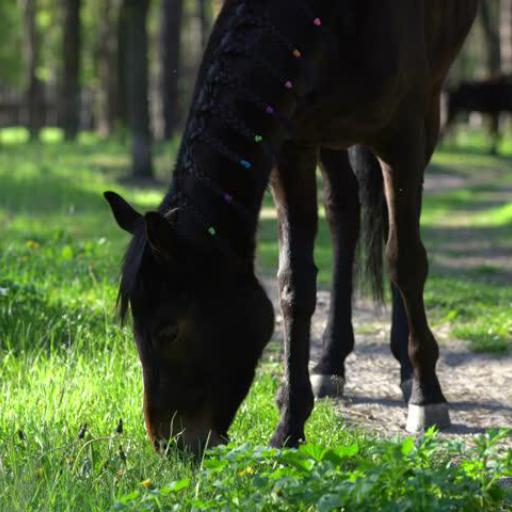}
    \end{minipage}&\multicolumn{5}{c}{
    \begin{minipage}{0.4\textwidth}
      \includegraphics[width=0.19\textwidth]{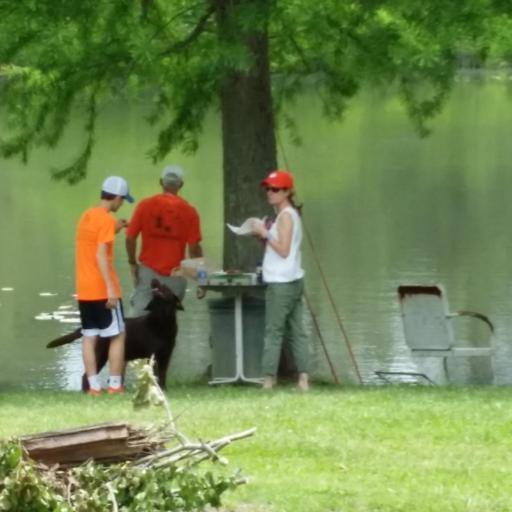}
      \includegraphics[width=0.19\textwidth]{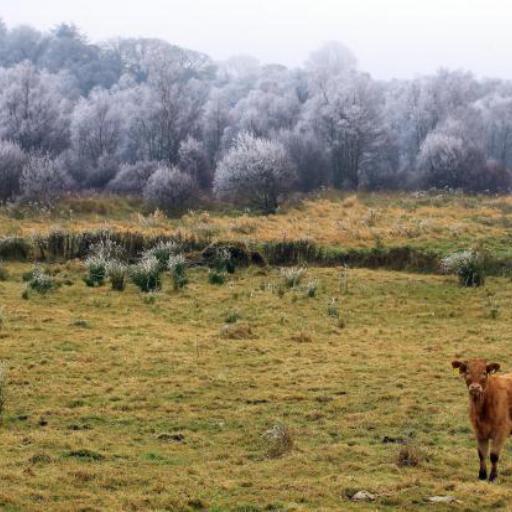}
      \includegraphics[width=0.19\textwidth]{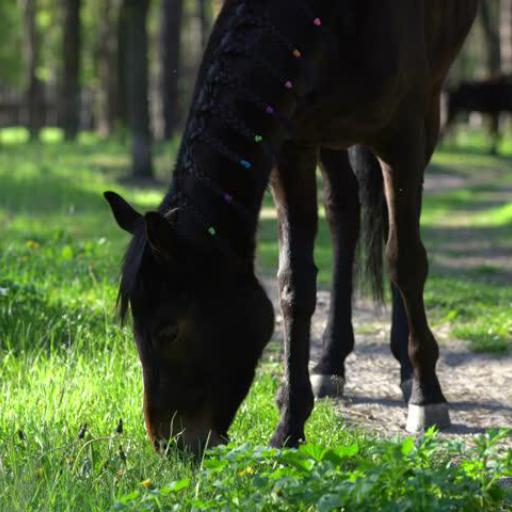}
      \includegraphics[width=0.19\textwidth]{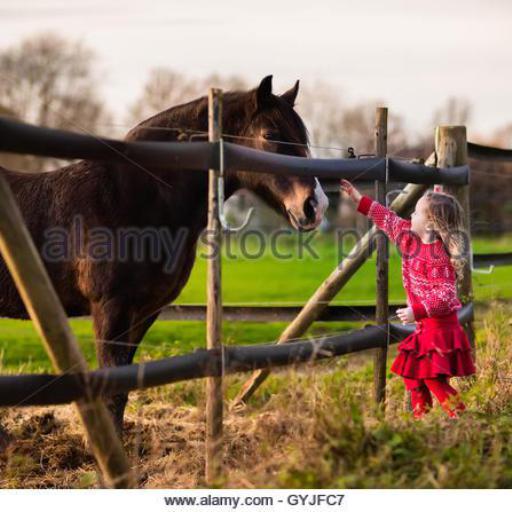}
      \includegraphics[width=0.19\textwidth]{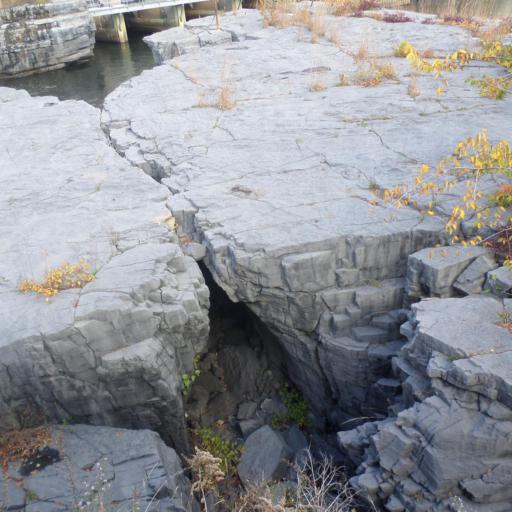}
    \end{minipage}}
    &&\multicolumn{5}{c}{
    \begin{minipage}{0.4\textwidth}
      \includegraphics[width=0.19\textwidth]{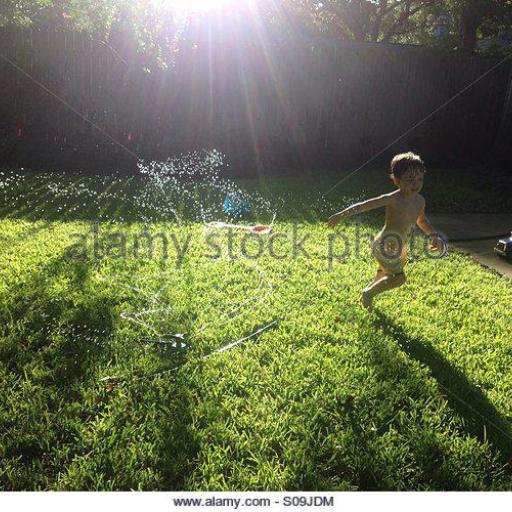}
      \includegraphics[width=0.19\textwidth]{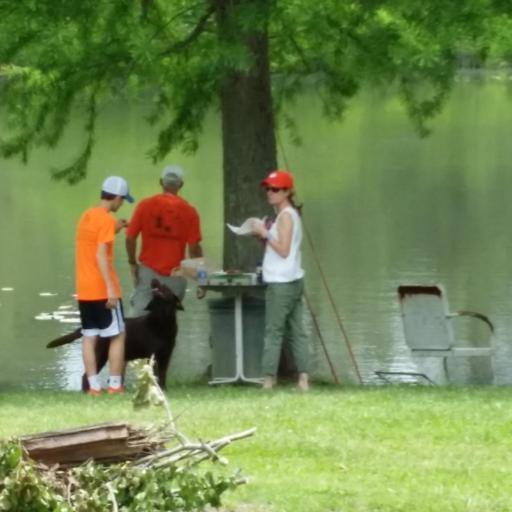}
      \includegraphics[width=0.19\textwidth]{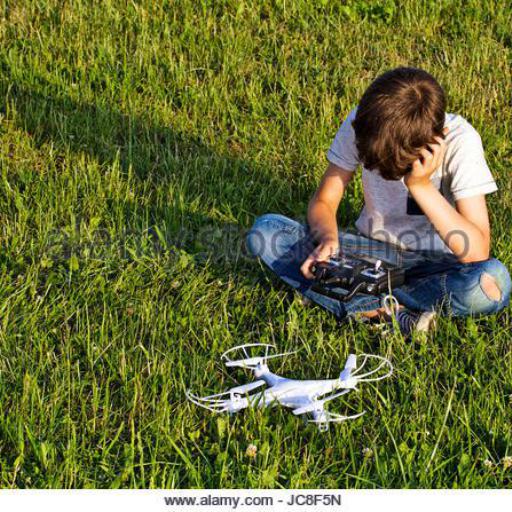}
      \includegraphics[width=0.19\textwidth]{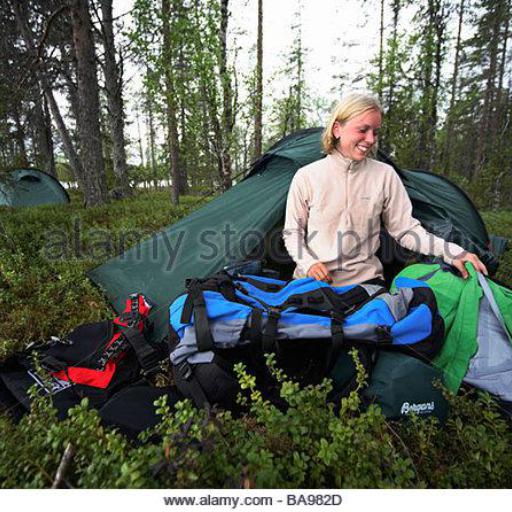}
      \includegraphics[width=0.19\textwidth]{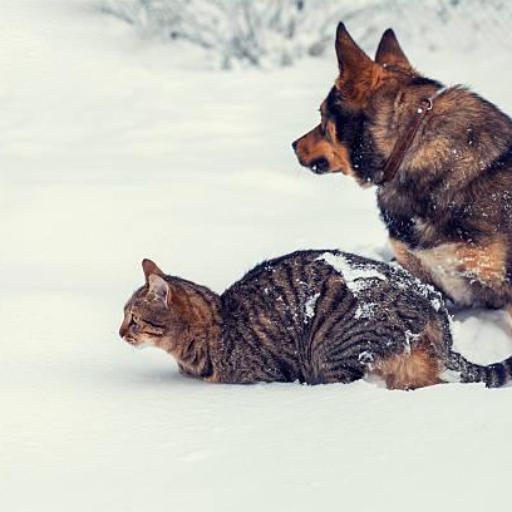}
    \end{minipage}} \\ \\
     \hline \\
     (b) \textcolor{blue}{Visible}&\multicolumn{11}{c}{A vector illustration of a happy male golfer.}\\ \\
     \begin{minipage}{0.08\textwidth}
      \includegraphics[width=\textwidth]{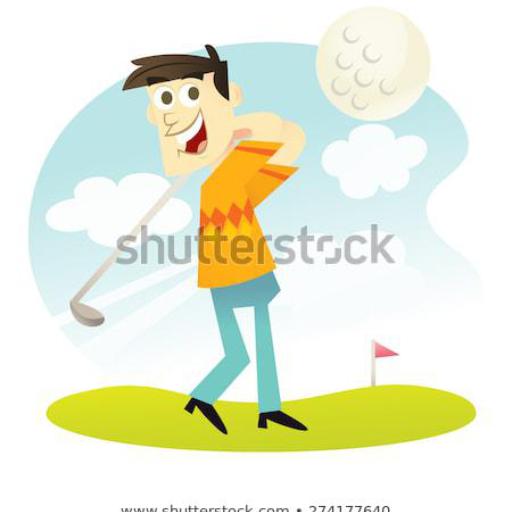}
    \end{minipage}&\multicolumn{5}{c}{
    \begin{minipage}{0.4\textwidth}
      \includegraphics[width=0.19\textwidth]{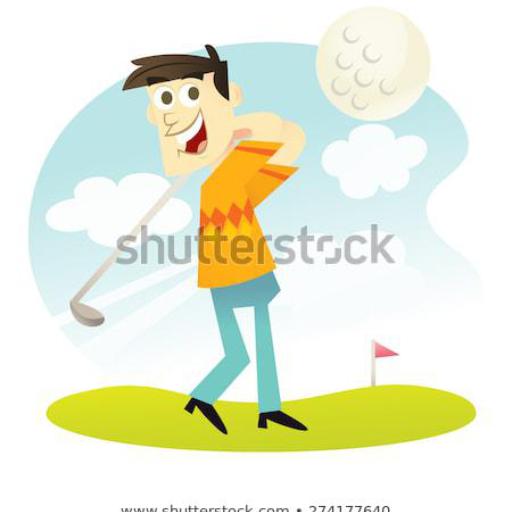}
      \includegraphics[width=0.19\textwidth]{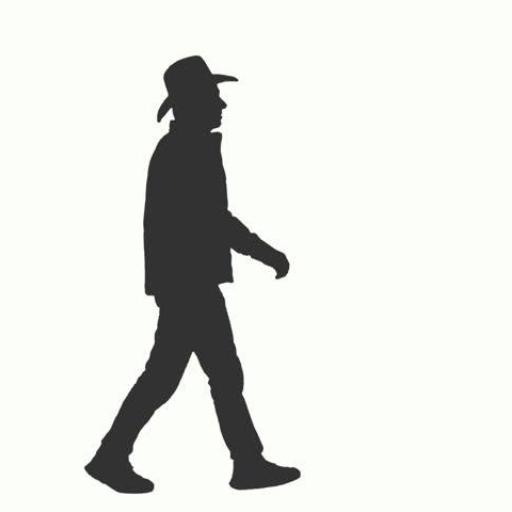}
      \includegraphics[width=0.19\textwidth]{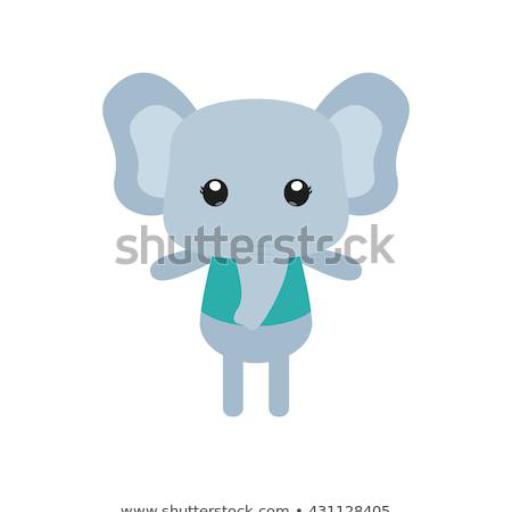}
      \includegraphics[width=0.19\textwidth]{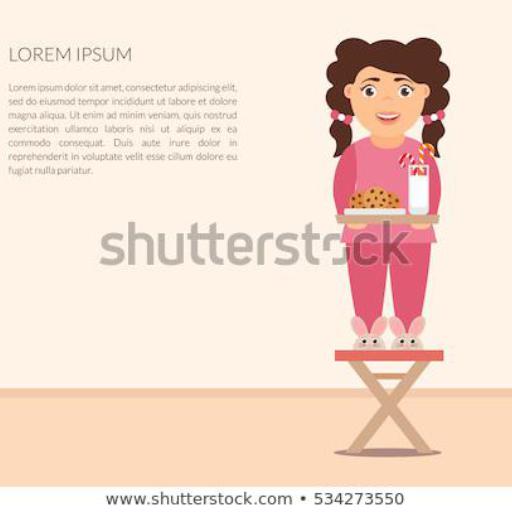}
      \includegraphics[width=0.19\textwidth]{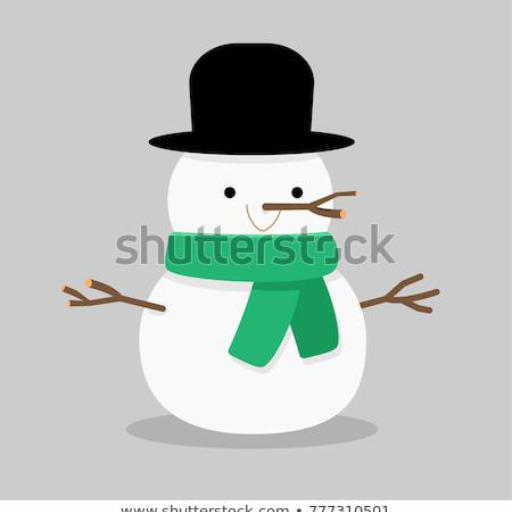}
    \end{minipage}}
    &&\multicolumn{5}{c}{
    \begin{minipage}{0.4\textwidth}
      \includegraphics[width=0.19\textwidth]{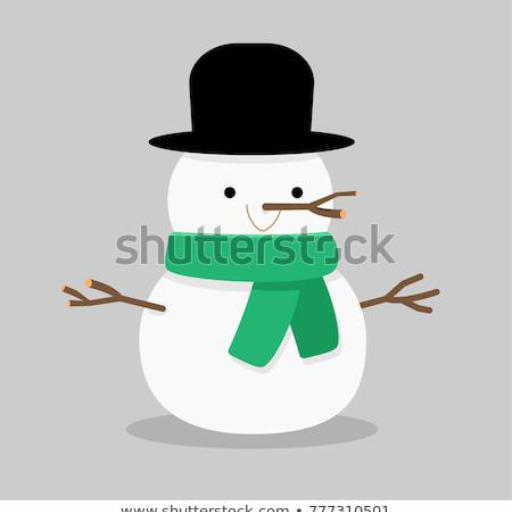}
      \includegraphics[width=0.19\textwidth]{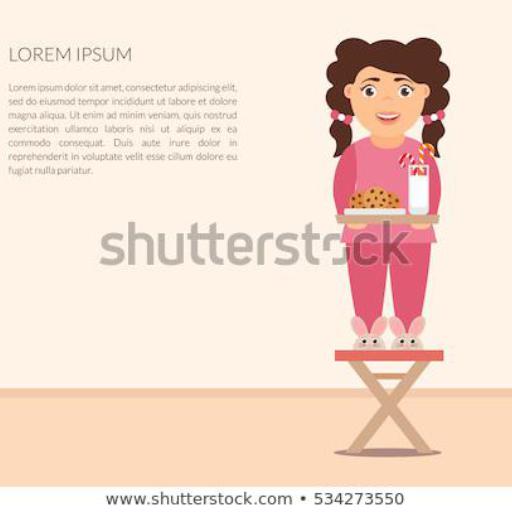}
      \includegraphics[width=0.19\textwidth]{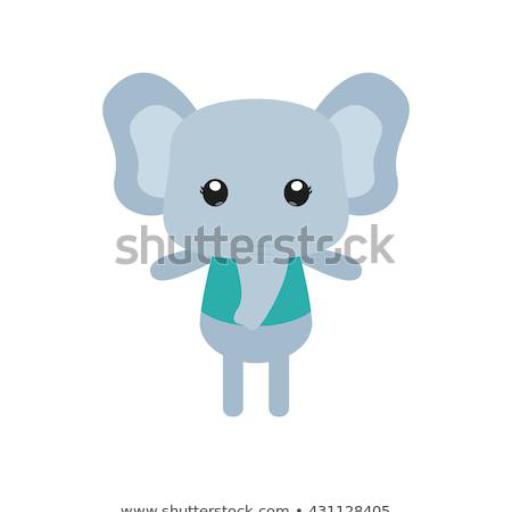}
      \includegraphics[width=0.19\textwidth]{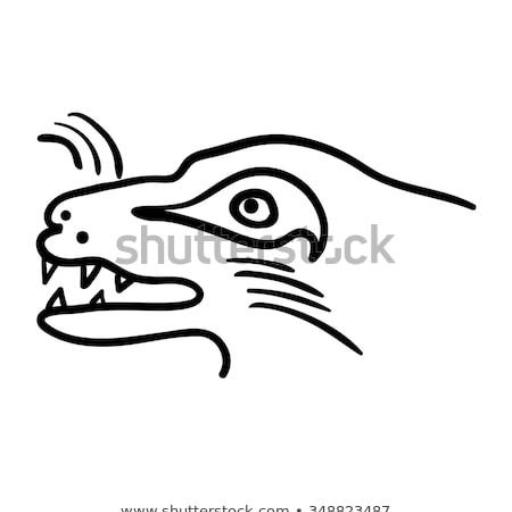}
      \includegraphics[width=0.19\textwidth]{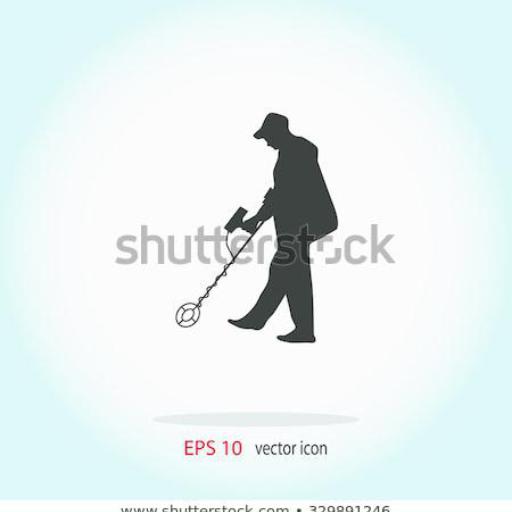}
    \end{minipage}} \\ \\
     \hline \\
     (c) \textcolor{blue}{Temporal$_{i>t}$}&\multicolumn{11}{c}{Finishing - Paint all the black parts except the door on the locomotive with gold food paint... add more details.}\\ \\
     \begin{minipage}{0.08\textwidth}
      \includegraphics[width=\textwidth]{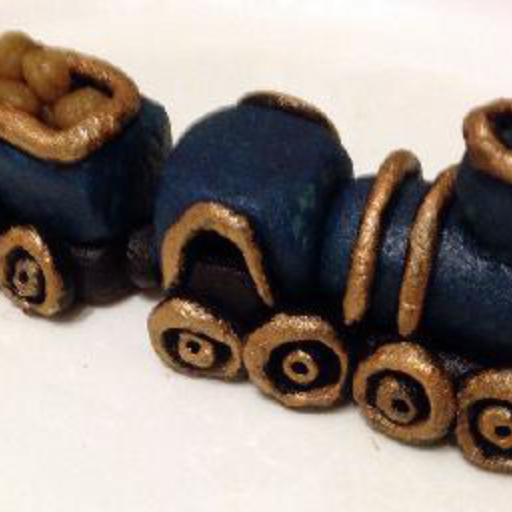}
    \end{minipage}&\multicolumn{5}{c}{
    \begin{minipage}{0.4\textwidth}
      \includegraphics[width=0.19\textwidth]{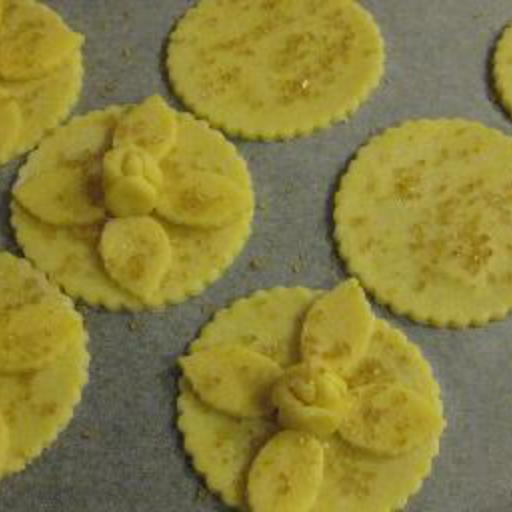}
      \includegraphics[width=0.19\textwidth]{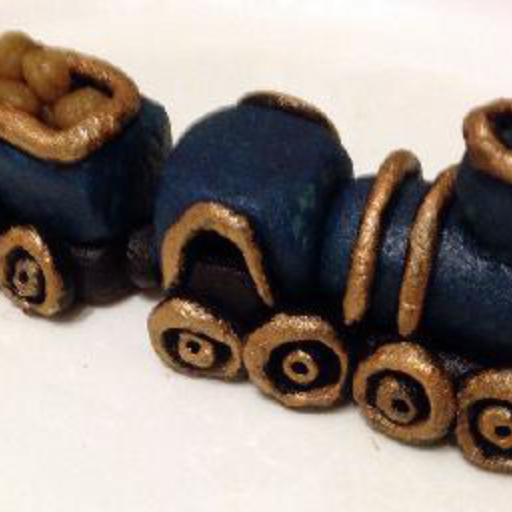}
      \includegraphics[width=0.19\textwidth]{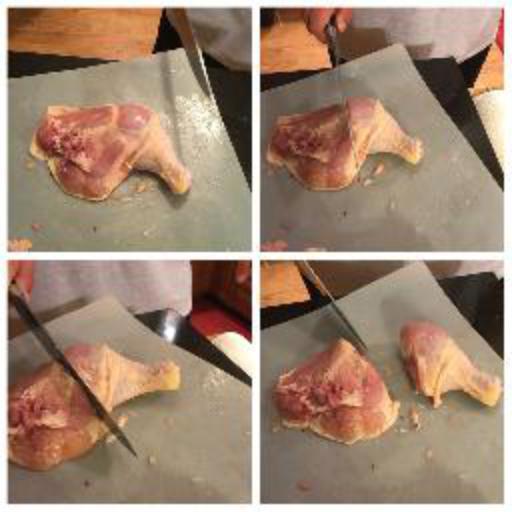}
      \includegraphics[width=0.19\textwidth]{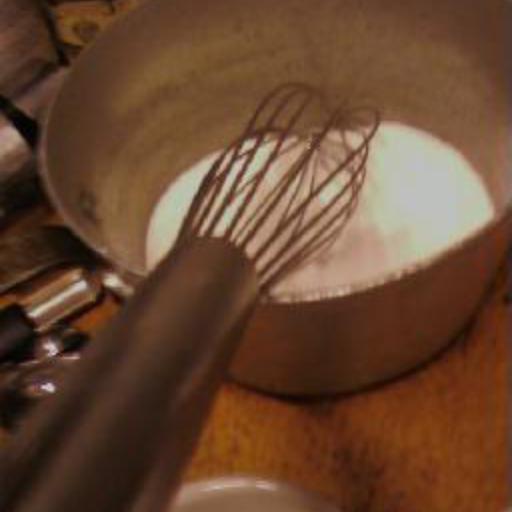}
      \includegraphics[width=0.19\textwidth]{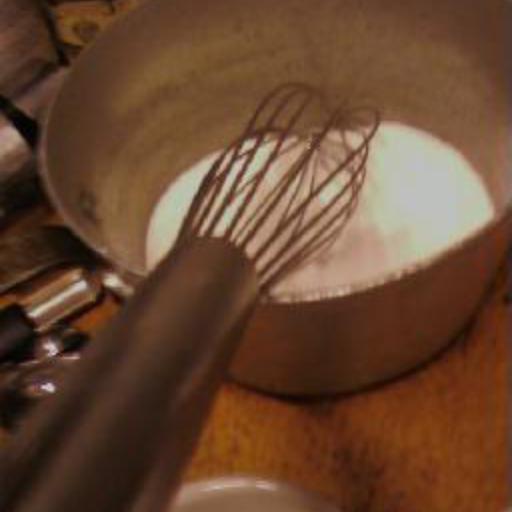}
    \end{minipage}}
    &&\multicolumn{5}{c}{
    \begin{minipage}{0.4\textwidth}
      \includegraphics[width=0.19\textwidth]{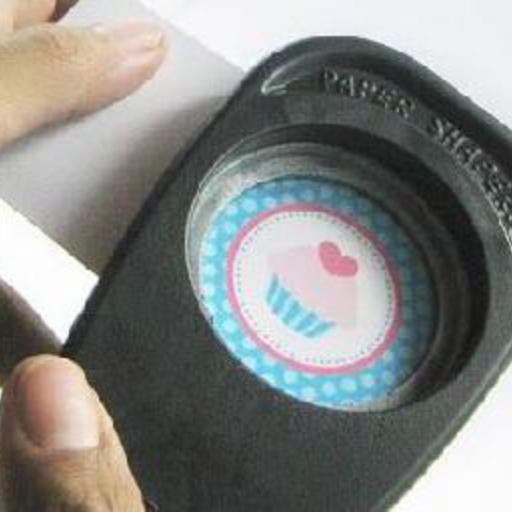}
      \includegraphics[width=0.19\textwidth]{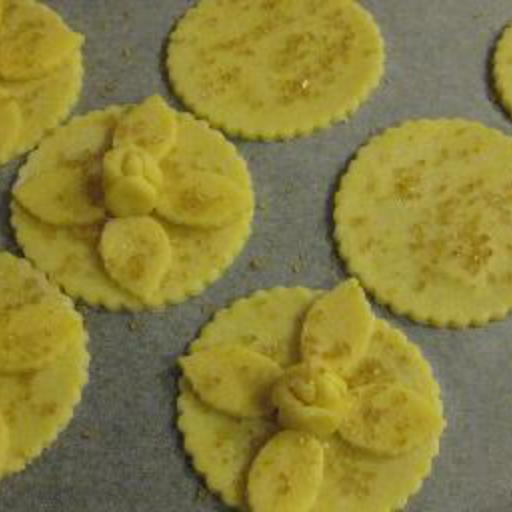}
      \includegraphics[width=0.19\textwidth]{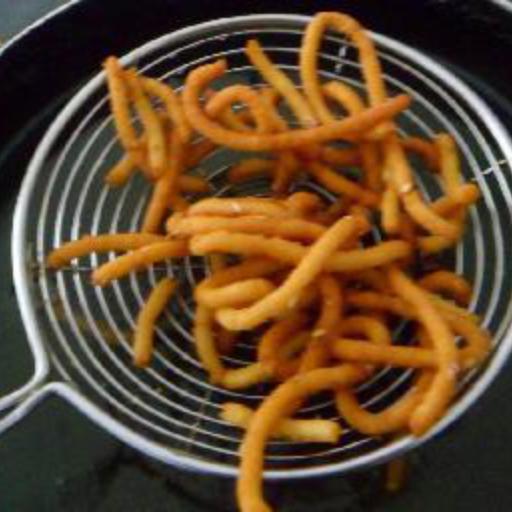}
      \includegraphics[width=0.19\textwidth]{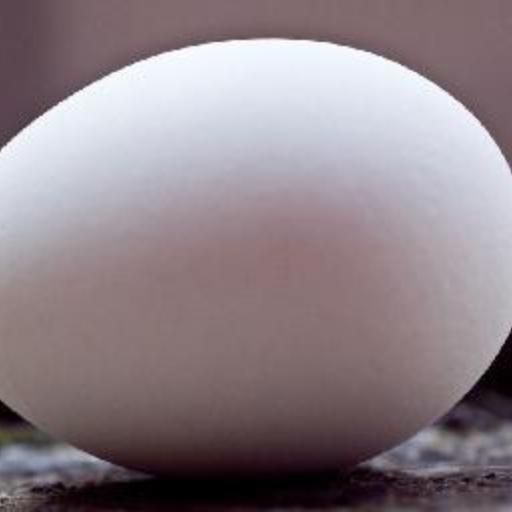}
      \includegraphics[width=0.19\textwidth]{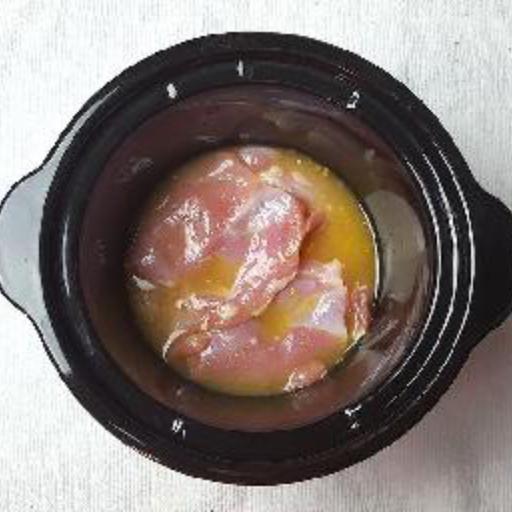}
    \end{minipage}} 
     %\begin{minipage}{0.07\textwidth}
     % \includegraphics[width=\textwidth]{557_result_visible/real.jpg}
    %\end{minipage}
    %&
    %\begin{minipage}{0.35\textwidth}
    %  \includegraphics[width=\textwidth]{557_result_visible/whole_image.png}
    %\end{minipage}\\ &&&\\
    %\multicolumn{2}{c}{\footnotesize{(a) \textcolor{blue}{Action:} Horse grazing on a summer meadow in the}}
    %&
    %\multicolumn{2}{c}{\footnotesize{(b) \textcolor{blue}{Visible:} A vector illustration of a happy male golfer}}\\
    %\multicolumn{2}{c}{\footnotesize{forest outdoors.}}&&\\
    \end{tabular}
    \caption{The ground truth image (Left) and the top 5 retrieved images by the \texttt{CMCM} and the \texttt{CMCA} models for two examples. The coherence relation (in blue) and caption are given above the images. The image-text pair in example (a) has \emph{Action} relation while in example (b) has \emph{Visible} relation. In example (a) the \texttt{CMCM} model leverages the \emph{Action} coherence relation to retrieve images that depict some action in the top 5. Similarly in example (c) images retrieved by proposed our model with \texttt{CAM} retrieves images that depict the result of a process as given by the \emph{Temporal$_{i>4}$} relation, whereas the agnostic model shows images that depict action in progress.}
    \label{fig:result_1}
\end{figure*}

\begin{table}[h]
\footnotesize
    \centering
    \begin{tabular}{lccccccc}
    \toprule
      & MedR$\downarrow$ &  R@1$\uparrow$ & R@5$\uparrow$ & R@10$\uparrow$ \\
    \cmidrule{2-5}
    \texttt{Base} & $19.8^{\pm 1.9}$ & $11.6$ & $28.6$ & $38.3$ \\
    \texttt{CMCA}& $19.3^{\pm 2.0}$ & $13.2$ & $30.6$ & $40.0$ \\
    \texttt{CMCM}-NoAttn & $20.6^{\pm 2.6}$ & $12.4$ & $28.9$ & $38.8$ \\
    \texttt{CMCM} & $\mathbf{18.7^{\pm 1.6}}$ & $\mathbf{13.8}$ & $\mathbf{31.6}$ & $\mathbf{40.6}$ \\
    \midrule
    \texttt{CMCM}$_{Visible}$ & $19.6^{\pm 3.1}$ & $13.4$ & $\mathbf{31.7}$ & $\mathbf{41.1}$ \\
    \texttt{CMCM}$_{Subjective}$ & $25.0^{\pm 3.1}$ & $12.9$ & $29.4$ & $38.0$ \\
    \texttt{CMCM}$_{Action}$ & $20.9^{\pm 2.1}$ & $11.7$ & $28.4$ & $38.0$ \\
    \texttt{CMCM}$_{Story}$ & $\mathbf{17.7^{\pm 1.7}}$ & $13.0$ & $30.7$ & $41.5$ \\
    \texttt{CMCM}$_{Meta}$ & $19.2^{\pm 1.5}$ & $13.1$ & $31.0$ & $40.7$ \\
    \texttt{CMCM}$_{Irrelevant}$ & $20.3^{\pm 1.9}$ & $12.6$ & $31.1$ & $40.9$ \\
    \bottomrule
    \end{tabular}
    \caption{Quantitative comparison of the models trained and evaluated on Clue dataset.}
    \label{tab:clue_quantitative_evaluation}
    \vspace{-0.2in}
\end{table}

\subsection{Human Evaluation}
\label{sec:humaneval}

Both CITE++ and Clue have image-text pairs with complex coherence relations in contrast to datasets like MSCOCO that have predominantly just \emph{Visible} relations. Hence, considering the ground truth as a gold standard is not reasonable. Given the wide distribution of different relations in the datasets, the quantitative metrics (\eg MedR and Recalls) are unreliable for the proposed setting. Therefore, we perform human evaluation where the top 1 retrieved images by \texttt{CMCA} and \texttt{CMCM} models are shown for pairwise comparison. 

% \textbf{Protocol.} 
We recruit 250 participants through Amazon Mechanical Turk. All subjects were US citizens, agreed to a consent form approved by the University of Pittsburgh IRB review board, and were compensated at an estimated rate of USD 15 an hour. We showed subjects the caption, the top image retrieved by the coherence aware and the coherence agnostic model for five relations from both the datasets and asked them to choose one of the following options:\\
(1) I prefer image A (2) I prefer image B (3) The images are exactly the same (4) Neither of the images is a good match for this text.
The order of images is random and each example was ranked by three workers and the final rank is decided via majority voting. The results are shown in \autoref{tab:humaneval}. 
It can be seen that the images retrieved by the proposed model are preferred by humans. More importantly, the difference in preference is significant in contrast to the quantitative metrics. We can also see that the difference in preference between \texttt{CMCM} and \texttt{CMCA} is higher when the relation is \emph{Subjective} or \emph{Story} when compared to regular captions (see \emph{Visible}), indicating the importance of explicitly modeling coherence relations for cross-modal understanding. The results of the t-test shows that the differences observed in \texttt{CMCM} and \texttt{CMCA} category are all statistically significant ($p<0.01 , \text{t}>14.1$). The results of the sensitivity power analysis shows that our experiment detects effect sizes as small as 0.17 with a power and significance level of 95\%.  These results effectively show that the quantitative metrics such as MedR and Recall can not solely measure performance of the model especially given the nature of the dataset and coherence relations.\\ 

\begin{table}[ht!]
\centering
\footnotesize
\begin{tabular}{lcccc}
\toprule
            & Better & Worse & Both Good & Both Bad \\ 
\cmidrule{2-5}
\texttt{CMCM}$_{Visible}$   & \textbf{24\%}  & 17\%     & 46\% & 13\% \\
\cmidrule{2-5}
\texttt{CMCM}$_{Subjective}$ & \textbf{53\%}  & 10\%     & 7\%  & 30\% \\
\cmidrule{2-5}
\texttt{CMCM}$_{Story}$   & \textbf{40\%}  & 10\%     & 33\% & 17\% \\
\cmidrule{2-5}
\texttt{CMCM}$_{Meta}$    & \textbf{56\%}  & 9\%      & 25\% & 10\% \\
\cmidrule{2-5}
\texttt{CMCM}$_{Q_7}$      & \textbf{43\%}  & 17\%     & 27\% & 13\% \\
\bottomrule
\end{tabular}
\caption{Human evaluation results. Values indicate the percentage of samples for which humans voted the output of \texttt{CMCM} as \textbf{Better}, \textbf{Worse}, \textbf{Both Good}, \textbf{Both Bad} when compared with \texttt{CMCA}.}
\label{tab:humaneval}
\vspace{-0.3in}
\end{table}
\subsection{Qualitative Analysis}
\begin{figure}[t]
    \vspace{-0.1in}
    \centering
    \includegraphics[width=0.48\linewidth]{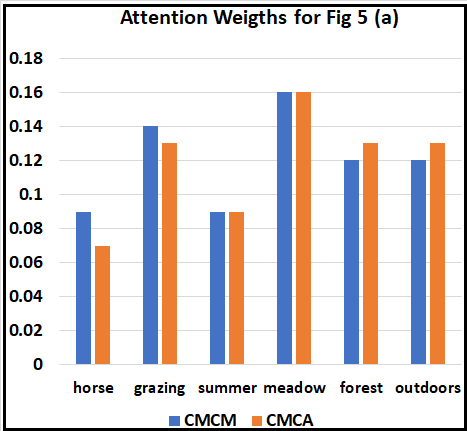} \hfill
    \includegraphics[width=0.465\linewidth]{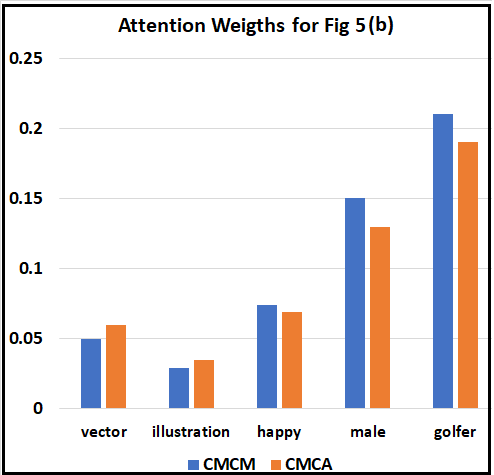}
    \caption{Attention weights for \texttt{CMCM} and \texttt{CMCA} models for example (a) and (b) in \autoref{fig:result_1}.}
    \label{fig:attn_weights}
    \vspace{-0.2in}
\end{figure}

To further understand the behavior of the model, we investigate the attention weights over input text for \texttt{CMCM} and \texttt{CMCA} models. In example \autoref{fig:result_1} (a), the proposed coherence-aware model retrieves the ground truth within the top 5 images. We can see from \autoref{fig:attn_weights} (left) that adding \cam increases the weight on words \emph{horse} and \emph{grazing} relative to the agnostic model. This can be attributed to the model's ability to predict the associated coherence relation to help retrieve the right image. The \texttt{CMCA} model, however, attends more to commonly visualized words like \emph{forest} and \emph{outdoors}.
Similarly, in \autoref{fig:result_1} (right), \texttt{CMCM} shows improved attention weights for words like \emph{male} and \emph{golfer}. The result is the model being able to retrieve the correct image in top 1 though both models retrieve images of \emph{Vector illustration} in the top 5~\autoref{fig:result_1} [b]. More examples for other relations are provided in the Appendix.
\iffalse
\begin{figure}[h!]
  \centering
  \footnotesize
  \begin{tabular}{cc}
    \begin{minipage}{0.07\textwidth}
      \includegraphics[width=\textwidth]{838_result_q7/real.jpg}
    \end{minipage}
    &
    \begin{minipage}{0.35\textwidth}
      \includegraphics[width=\textwidth]{838_result_q7/whole_image.png}
    \end{minipage} \\ & \\
    \hline
    \multicolumn{2}{c}{\footnotesize{\textcolor{blue}{Temporal$_{i>t}$:} Finishing - Paint all the black parts except}} \\
    \multicolumn{2}{c}{\footnotesize{the door on the locomotive with gold food paint. Fill the}} \\
    \multicolumn{2}{c}{\footnotesize{wagons with little marzipan pellets to add more details.}} \\
    \cline{1-2}
    \end{tabular}
    \caption{The ground truth image (Left) and the top 5 retrieved images by the \texttt{CMCM} (right, green box) and the \texttt{CMCA} (right, red box) models for an example from CITE++ dataset are given. The GT coherence relation (in blue) and caption are given at the bottom.}
    \label{fig:result_cite}
\end{figure}
\fi

In CITE++ dataset, we observe similar behavior as shown in \autoref{fig:result_1} [c]. The relation \emph{Temporal$_{i>t}$} characterizes the temporal correlation between an image and text where the image visualizes the result of the process described in the corresponding text. These relations are difficult to implicitly understand as the text is no different from any other step in the recipe. Training with \cam that explicitly models temporal relation improves the performance of image retrieval. For example, we can see in \autoref{fig:result_1} that all top 5 images retrieved by the \texttt{CMCM} are images that visualize the \emph{result of a process}, in contrast to \texttt{CMCA} model that shows images of \emph{the step being carried out} as well. 

\subsection{Predicting Coherence Relations} 
To further understand the effect and importance of coherence relations, we analyze the model's ability to predict the presence of a coherence relation given the ground truth image and text. For this, we use the models trained using the original objective in \autoref{eq:final}. We provide the ground truth image and text as input and calculate the Average Precision (AP) of coherence relation prediction. The results are provided in %\autoref{tab:coh_classification_cite} for CITE++ dataset. The results for Clue dataset are provided in 
the Appendix. We can see that in most cases as expected, the \texttt{CMCM}$_{c}$ algorithms perform reasonably well. We haven't provided the results for \texttt{CMCA} models as they were not trained with coherence relations. These results are comparable to similar experiments performed in \cite{alikhani2020cross} though in their experiment, classification was the only objective. Interestingly \emph{Subjective} relation has very low AP (cf. appendix) similar to retrieval performance but the proposed model obtains significance gain in performance in the human evaluation. 
% This reinforces the unreasonableness of the Recall and Rank metrics for this task. 

%%%%%%%%%%%%%%%%%%%%%%%%%
\section{Conclusion}
%%%%%%%%%%%%%%%%%%%%%%%%%
Automating the understanding and generation of multimodal discourse requires a joint understanding of co-occurring images and text. Our study shows the effectiveness of cross-modal coherence modeling for text-to-image retrieval tasks.
% 
% We propose a novel coherence-aware text-to-image retrieval model trained using an auxiliary \cam that learns to predict the coherence relation that characterizes a ground truth image-text pair during training. 
% 
% We release an extended version of the CITE dataset called CITE++ with coherence relation annotations for twice the number of data samples. 
% We show quantitatively that our model performs better than state-of-the-art retrieval models on CITE++ and Clue datasets. Moreover, c
Our evaluation shows that the performance of the coherence-aware model is significantly better compared to the agnostic models. We also observe that the existing Recall based quantitative metrics for text-to-image retrieval are unreliable and fail to meaningfully evaluate retrieval systems especially when image-text pairs can be characterized by different coherent relations. Future work involves developing new transformer-based coherence-aware metrics that can better measure the performance of retrieval models. Based on the evidence shown in this paper, an important extension is to annotate existing datasets with coherence relations to further improve semantic joint understanding of image and text.

The research presented in this paper has been supported by NSF awards IIS-1703883, IIS-1955404, IIS-1955365, IIS 1955404, RETTL-2119265, IIS-1526723, CCF-1934924, and EAGER-2122119, and through generous donations from Adobe.

\bibliography{naacl2021}
% \bibliographystyle{aaai22}

% \clearpage
\newpage
\section{Training Details} 
We train all the models using the Adam \cite{kingma2014adam} optimizer with an initial learning rate of $10^{-4}$. 
All models are trained for about $20$ epochs and the epoch with the lowest MedR on validation data is chosen as the best epoch for inference.
Throughout training, the model sees the text-image pair and the corresponding coherence relation as input.
During inference, we give the text as input and retrieve the closest image in the shared space. The coherence relations are not used during inference.

For all models, we repeat the testing experiments three times and report the standard deviation of the main metric. The word2vec model for CITE++ and Clue datasets are trained on the training set with $minimum\,frequency=1$ and $window\,size=10$, which outputs a $\mathbb{R}^{300}$ vector for all the words in the vocabulary. The vocabulary size is $8917$ for CITE++, whereas for Clue it is $5612$. Training the retrieval model takes about 20 minutes on CITE++ and 35 minutes on Clue datasets with an NVIDIA K80 GPU and all codes were written using PyTorch \cite{NEURIPS2019_9015} GPU library.

\subsection{Hyperparameter Search} 
\begin{figure}[th]
    \centering
    \includegraphics[width=\linewidth]{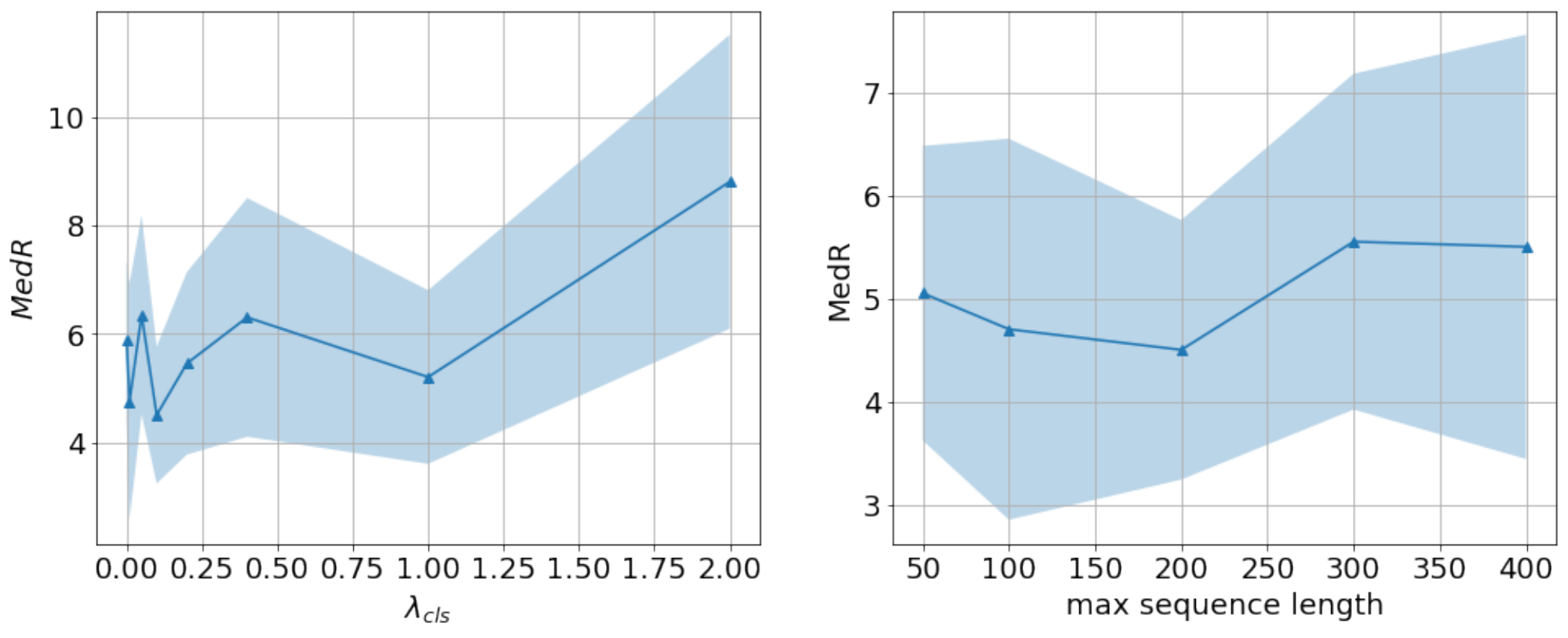}
    \caption{On CITE++ dataset, different values for $\lambda_{cls}$ vs MedR to determine the best value for $\lambda_{cls}$ as $0.1$ (Left); Text sequence length vs MedR to determine the best value for maximum sequence length as $200$ (Right). }
    \label{fig:cite_lambda_vs_medr}
\end{figure}

We determine the best value for the weight $\lambda_{cls}$ for \texttt{CAM} empirically as shown in \autoref{fig:cite_lambda_vs_medr} (Left), $\lambda_{cls}=0.10$ achieves the best MedR. 
Another key parameter is the maximum sequence length for the LSTM module in the text encoder, smaller value may lose information in the sentence and larger value may bring extra burden to the model.
We observe in \autoref{fig:cite_lambda_vs_medr} (Right) that $length=200$ achieves the best MedR for CITE++ dataset. 
We did similar experiments for Clue dataset and decide these value to be $\lambda_{cls}=0.10$  and $length=40$.

\section{Impact of Coherence Relations in CMCM}
We perform further analysis to evaluate the impact of each coherence relation on the performance of \texttt{CMCM} when compared with \texttt{CMCA}. We calculate all the quantitative metric values for the overall \texttt{CMCM} over subsets of the test set where all the samples have a particular relation as positive. The results are shown in \autoref{tab:per_relation_notrain}. We can observe that for most of the metrics, the proposed \texttt{CMCM} model performs better than or equivalent to \texttt{CMCA} for all relations. We also emphasize that the quantitative metrics are not sufficient to evaluate the performance of the models given the subjective nature of the task, dataset and coherence relations. However, the better performance is still encouraging. 

\begin{table*}[th]
\centering
    \begin{tabular}{llccccccc}
    \toprule
     Relation&Model & MedR$\downarrow$ &  R@1$\uparrow$ & R@5$\uparrow$ & R@10$\uparrow$ \\
    \midrule
    \multirow{2}{*}{Q2}&\texttt{CMCA}& $3.0$ & $37.0$ & $54.0$ & $58.0$ \\
    &\texttt{CMCM} & $3.0$ & $37.0$ & $\mathbf{55.0}$ & $58.0$ \\
    \midrule
    \multirow{2}{*}{Q3}&\texttt{CMCA}& $2.0$ & $43.0$ & $59.0$ & $61.0$ \\
    &\texttt{CMCM} & $2.0$ & $\mathbf{45.0}$ & $\mathbf{61.0}$ & $61.0$ \\
    \midrule
    \multirow{2}{*}{Q4}&\texttt{CMCA}& $\mathbf{4.0}$ & $35.0$ & $50.0$ & $\mathbf{55.0}$ \\
    &\texttt{CMCM} & $5.0$ & $35.0$ & $50.0$ & $52.0$ \\
    \midrule
    \multirow{2}{*}{Q5}&\texttt{CMCA}& $2.0$ & $48.0$ & $60.0$ & $65.0$ \\
    &\texttt{CMCM} & $2.0$ & $48.0$ & $\mathbf{64.0}$ & $\mathbf{67.0}$ \\
    \midrule
    \multirow{2}{*}{Q6}&\texttt{CMCA}& $2.0$ & $42.0$ & $52.0$ & $56.0$ \\
    &\texttt{CMCM} & $2.0$ & $42.0$ & $\mathbf{54.0}$ & $56.0$ \\
    \midrule
    \multirow{2}{*}{Q7}&\texttt{CMCA}& $5.0$ & $36.0$ & $48.0$ & $52.0$ \\
    &\texttt{CMCM} & $\mathbf{4.0}$ & $\mathbf{38.0}$ & $\mathbf{51.0}$ & $\mathbf{55.0}$ \\
    \midrule
    \multirow{2}{*}{Q8}&\texttt{CMCA}& $2.0$ & $42.0$ & $54.0$ & $57.0$ \\
    &\texttt{CMCM} & $2.0$ & $42.0$ & $\mathbf{56.0}$ & $\mathbf{60.0}$ \\
    \midrule \midrule
    \multirow{2}{*}{Visible}&\texttt{CMCA}& $12.0$ & $11.0$ & $34.0$ & $48.0$ \\
    &\texttt{CMCM} & $\mathbf{11.0}$ & $\mathbf{12.0}$ & $\mathbf{35.0}$ & $\mathbf{50.0}$ \\
    \midrule
    \multirow{2}{*}{Subjective}&\texttt{CMCA}& $28.0$ & $4.0$ & $24.0$ & $\mathbf{40.0}$ \\
    &\texttt{CMCM} & $\mathbf{22.0}$ & $\mathbf{5.0}$ & $\mathbf{21.0}$ & $29.0$ \\
    \midrule
    \multirow{2}{*}{Action}&\texttt{CMCA}& $\mathbf{12.0}$ & $\mathbf{17.0}$ & $\mathbf{36.0}$ & $\mathbf{47.0}$ \\
    &\texttt{CMCM} & $18.0$ & $10.0$ & $32.0$ & $42.0$ \\
    \midrule
    \multirow{2}{*}{Story}&\texttt{CMCA}& $\mathbf{25.0}$ & $5.0$ & $20.0$ & $30.0$ \\
    &\texttt{CMCM} & $28.0$ & $\mathbf{6.0}$ & $\mathbf{25.0}$ & $\mathbf{37.0}$ \\
    \midrule
    \multirow{2}{*}{Meta}&\texttt{CMCA}& $9.0$ & $12.0$ & $40.0$ & $52.0$ \\
    &\texttt{CMCM} & $9.0$ & $\mathbf{13.0}$ & $\mathbf{41.0}$ & $\mathbf{56.0}$ \\
    \midrule
    \multirow{2}{*}{Irrelevant}&\texttt{CMCA}& $15.0$ & $17.0$ & $43.0$ & $48.0$ \\
    &\texttt{CMCM} & $\mathbf{4.0}$ & $\mathbf{21.0}$ & $\mathbf{54.0}$ & $\mathbf{62.0}$ \\
    \bottomrule
    \end{tabular}
    \caption{Quantitative comparison in CITE++ and CLUE datasets over only examples in the test set that have a particular relation. $\downarrow$ indicates that lower the better and $\uparrow$ indicates that higher the better.}
    \label{tab:per_relation_notrain}
\end{table*}

\section{CITE++ Examples}
\begin{figure*}[t]
  \centering
  \footnotesize
  \begin{tabular}{cccccccccccc}
    %GT & Retrieved & GT & Retrieved\\ 
    GT & \multicolumn{5}{c}{\texttt{CMCM}} && \multicolumn{5}{c}{\texttt{CMCA}} \\
    \hline\\
    (a) \textcolor{blue}{Q2}&\multicolumn{11}{c}{Fill 30 cupcake tins with batter, 3/4 full. Bake in oven }\\
    &\multicolumn{11}{c}{for about 17 minutes or until toothpick comes out clean. }\\ 
     &\multicolumn{11}{c}{Cool in pans for about ten minutes and then finish cooling on cooling rack. }\\ \\
    \begin{minipage}{0.08\textwidth}
      \includegraphics[width=\textwidth]{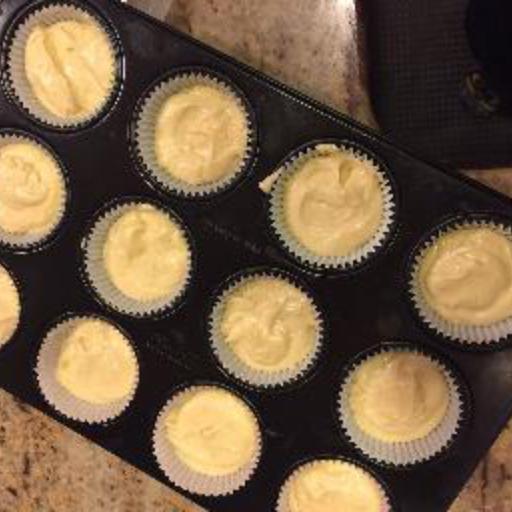}
    \end{minipage}&\multicolumn{5}{c}{
    \begin{minipage}{0.45\textwidth}
      \includegraphics[width=0.19\textwidth]{appendix/q2/q2_top0.jpg}
      \includegraphics[width=0.19\textwidth]{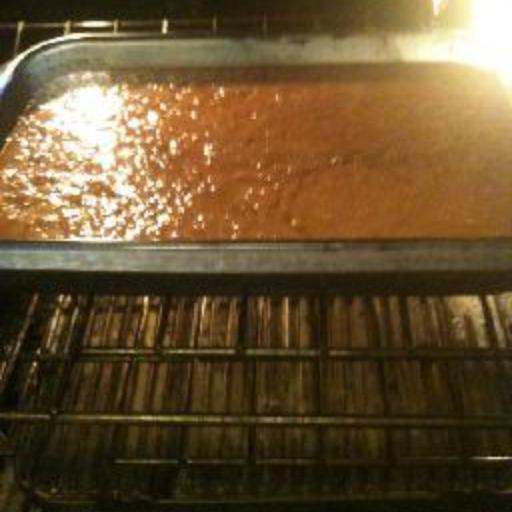}
      \includegraphics[width=0.19\textwidth]{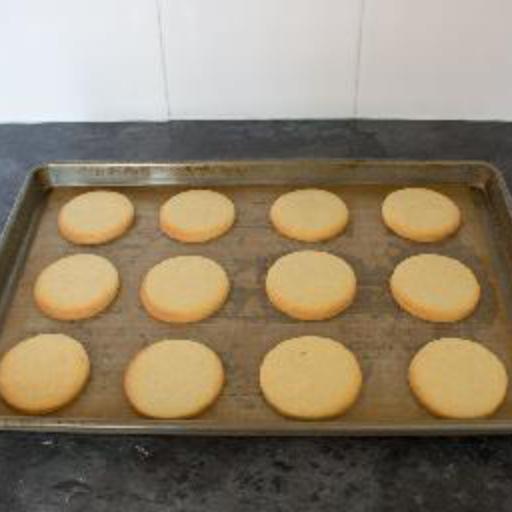}
      \includegraphics[width=0.19\textwidth]{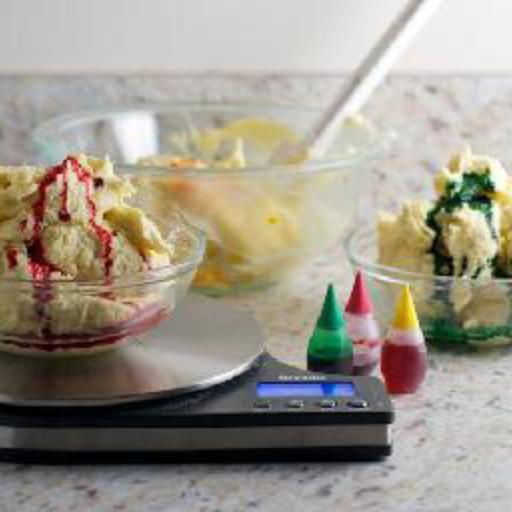}
      \includegraphics[width=0.19\textwidth]{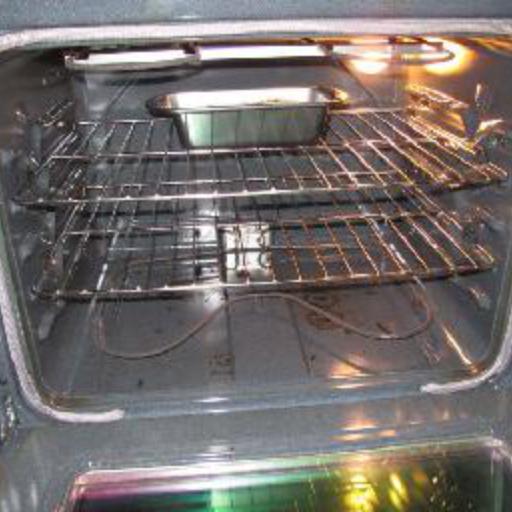}
    \end{minipage}}
    &&\multicolumn{5}{c}{
    \begin{minipage}{0.45\textwidth}
      \includegraphics[width=0.19\textwidth]{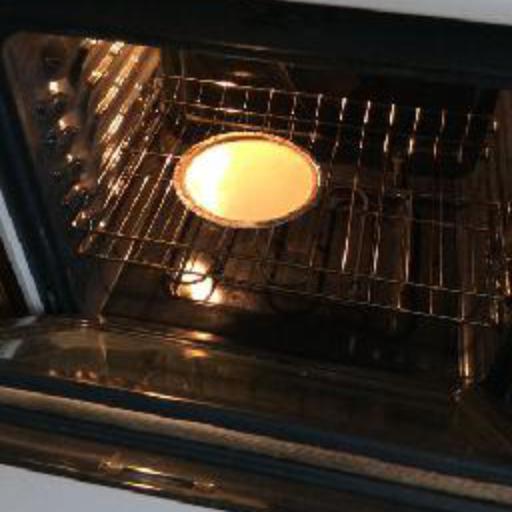}
      \includegraphics[width=0.19\textwidth]{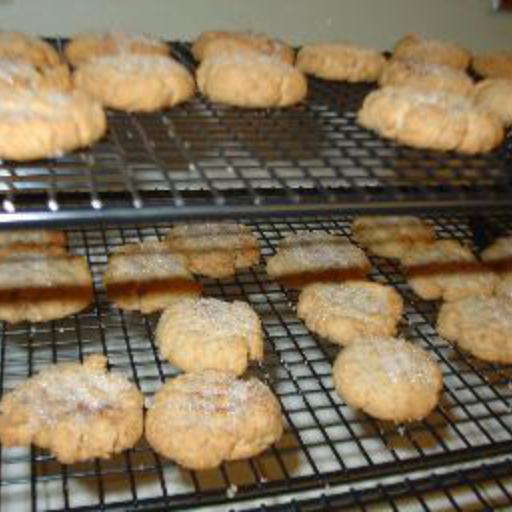}
      \includegraphics[width=0.19\textwidth]{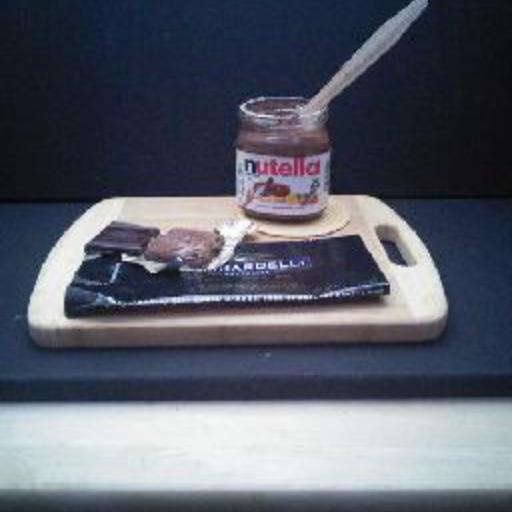}
      \includegraphics[width=0.19\textwidth]{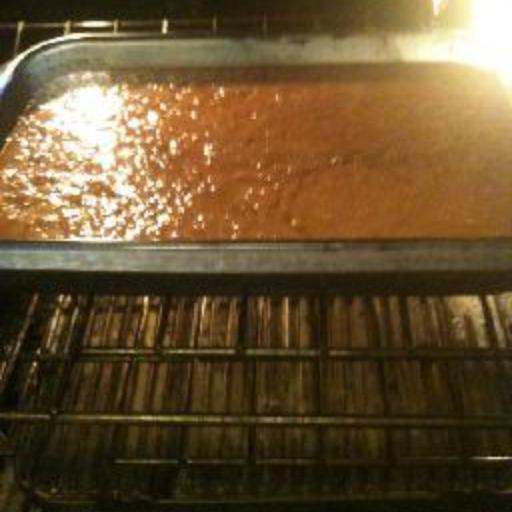}
      \includegraphics[width=0.19\textwidth]{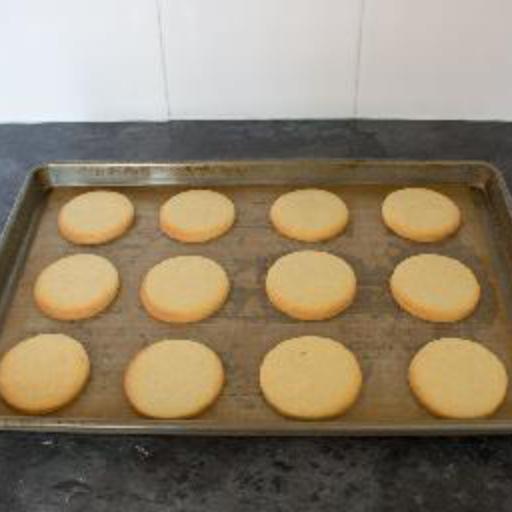}
    \end{minipage}} \\ \\
     \hline \\
     (b) \textcolor{blue}{Q3}&\multicolumn{11}{c}{Introduce sugar in a stream - Being sure to add it in a slow stream. }\\
    &\multicolumn{11}{c}{Whisk the sugar into the eggs espresso powder salt and vanilla mixture.}\\
    &\multicolumn{11}{c}{Be aware that too much sugar at same time may cause batter to be grainy.}\\ \\
     \begin{minipage}{0.08\textwidth}
      \includegraphics[width=\textwidth]{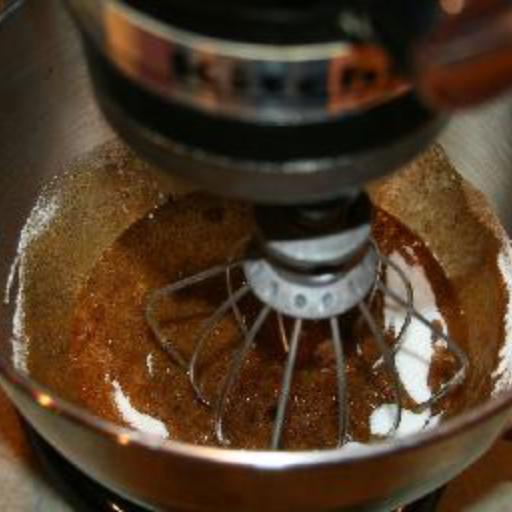}
    \end{minipage}&\multicolumn{5}{c}{
    \begin{minipage}{0.45\textwidth}
      \includegraphics[width=0.19\textwidth]{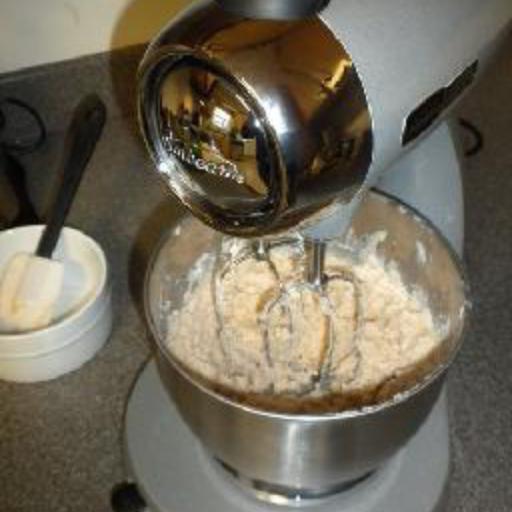}
      \includegraphics[width=0.19\textwidth]{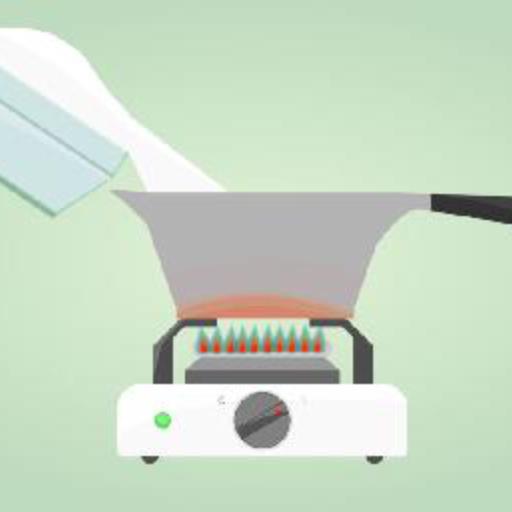}
      \includegraphics[width=0.19\textwidth]{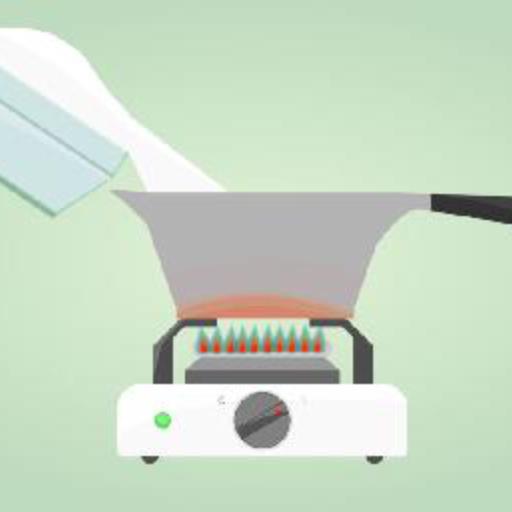}
      \includegraphics[width=0.19\textwidth]{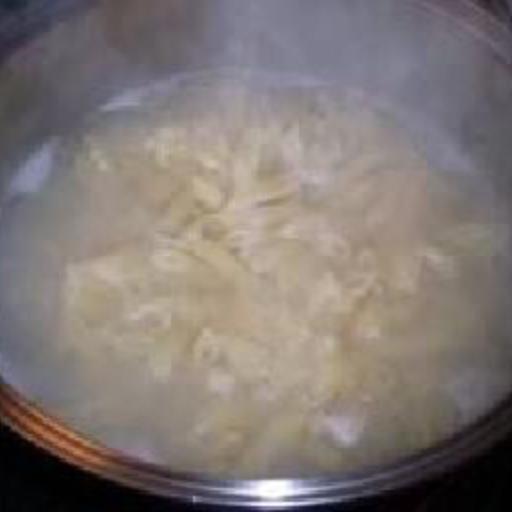}
      \includegraphics[width=0.19\textwidth]{appendix/q3/q3_top4.jpg}
    \end{minipage}}
    &&\multicolumn{5}{c}{
    \begin{minipage}{0.45\textwidth}
      \includegraphics[width=0.19\textwidth]{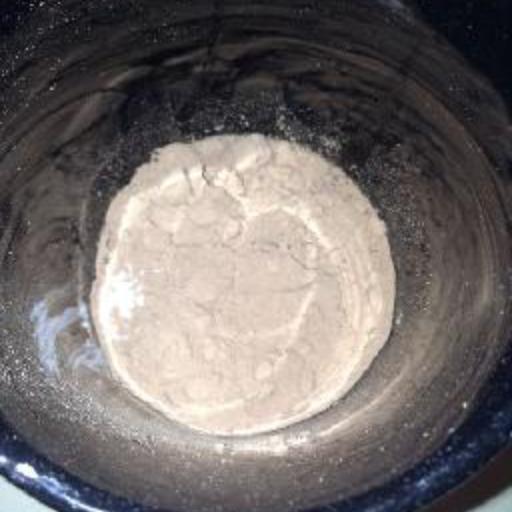}
      \includegraphics[width=0.19\textwidth]{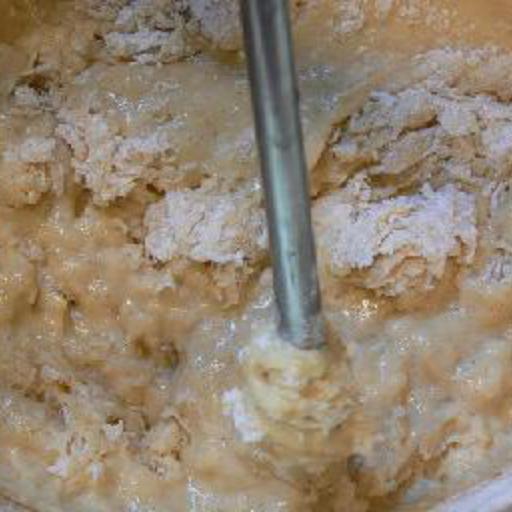}
      \includegraphics[width=0.19\textwidth]{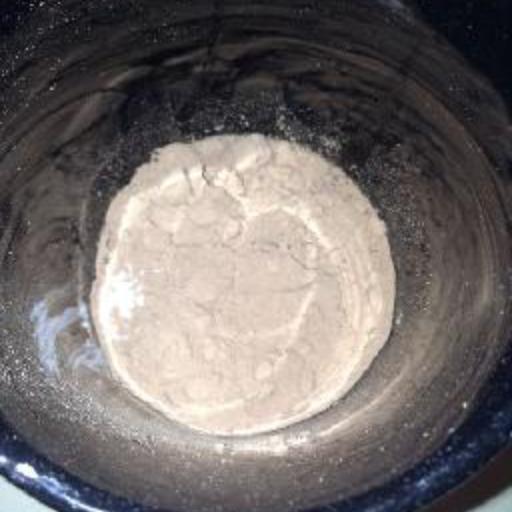}
      \includegraphics[width=0.19\textwidth]{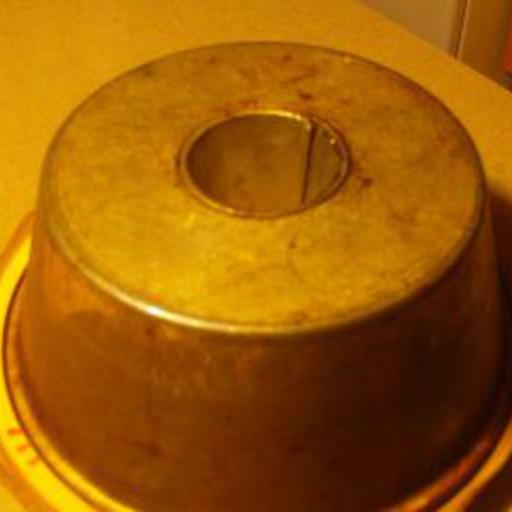}
      \includegraphics[width=0.19\textwidth]{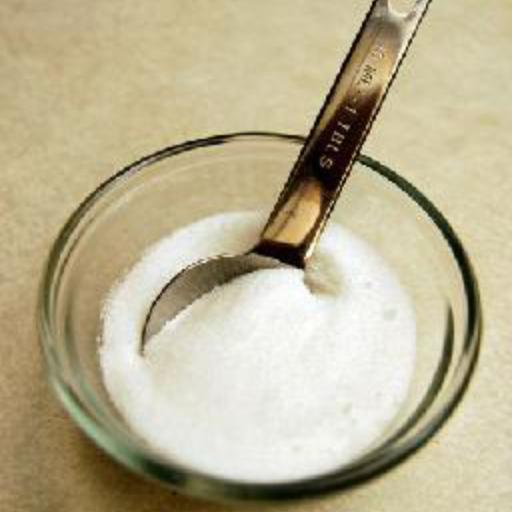}
    \end{minipage}} \\ \\
     \hline \\
     (c) \textcolor{blue}{Q4}&\multicolumn{11}{c}{Make sure ingredients are evenly distributed}\\
    &\multicolumn{11}{c}{but not over mized. 2. Bake at 35-40 min at 350 degrees. }\\ 
    &\multicolumn{11}{c}{3. A toothpick will come out clean}\\
      \begin{minipage}{0.08\textwidth}
      \includegraphics[width=\textwidth]{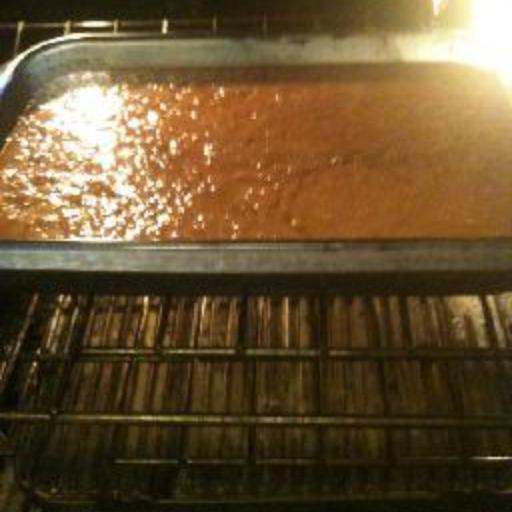}
    \end{minipage}&\multicolumn{5}{c}{
    \begin{minipage}{0.45\textwidth}
      \includegraphics[width=0.19\textwidth]{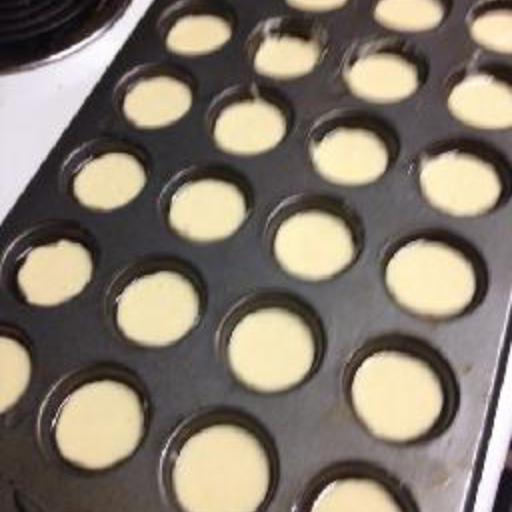}
      \includegraphics[width=0.19\textwidth]{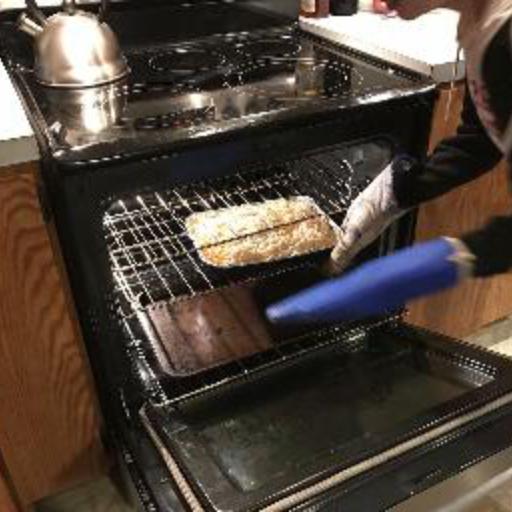}
      \includegraphics[width=0.19\textwidth]{appendix/q4/q4_top2.jpg}
      \includegraphics[width=0.19\textwidth]{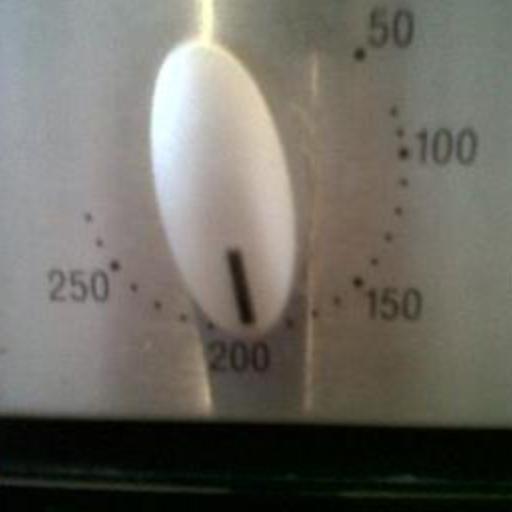}
      \includegraphics[width=0.19\textwidth]{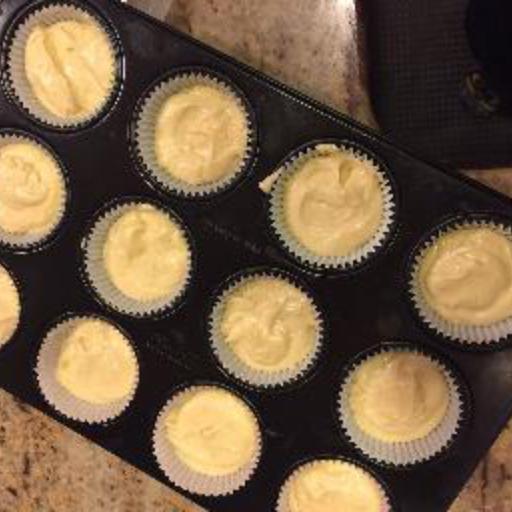}
    \end{minipage}}
    &&\multicolumn{5}{c}{
    \begin{minipage}{0.45\textwidth}
      \includegraphics[width=0.19\textwidth]{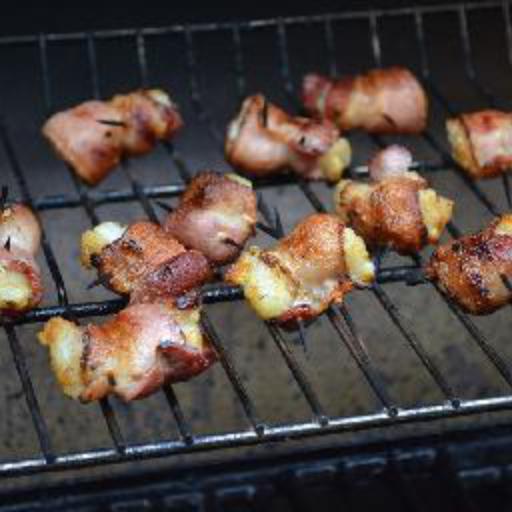}
      \includegraphics[width=0.19\textwidth]{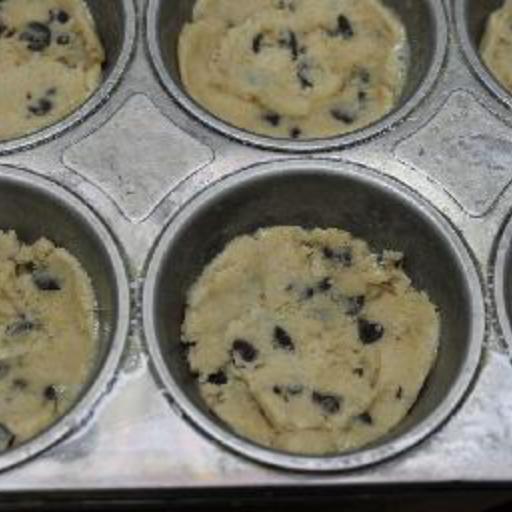}
      \includegraphics[width=0.19\textwidth]{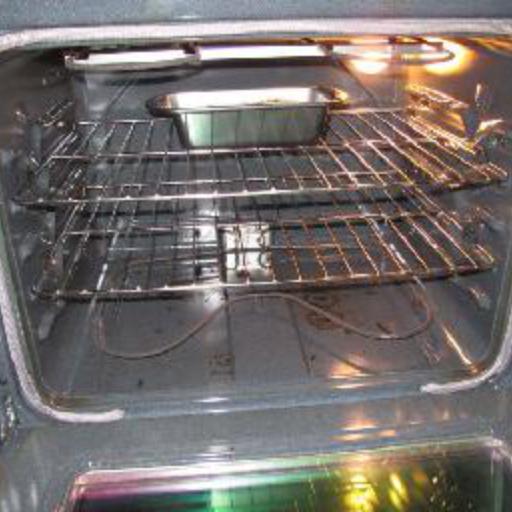}
      \includegraphics[width=0.19\textwidth]{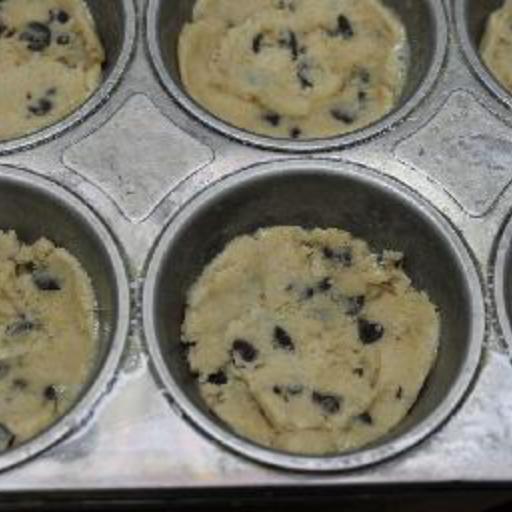}
      \includegraphics[width=0.19\textwidth]{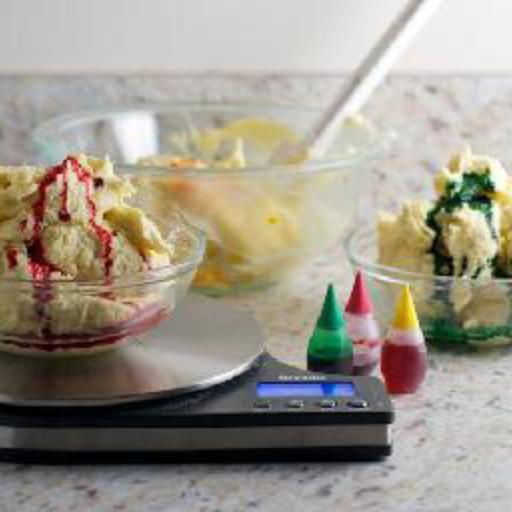}
    \end{minipage}} \\ \\
    \hline \\
     (d) \textcolor{blue}{Q5}&\multicolumn{11}{c}{The cabbage need to be diced.}\\ \\
      \begin{minipage}{0.08\textwidth}
      \includegraphics[width=\textwidth]{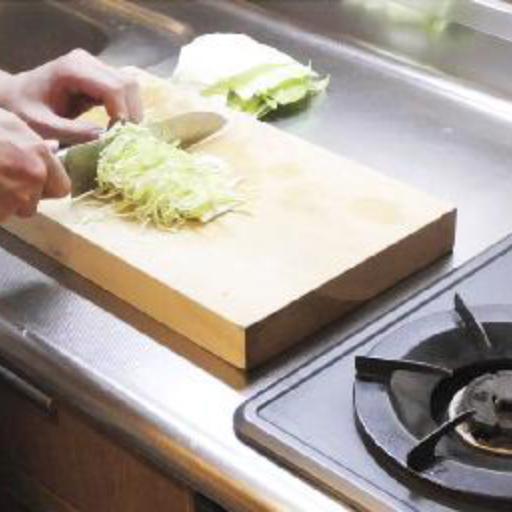}
    \end{minipage}&\multicolumn{5}{c}{
    \begin{minipage}{0.45\textwidth}
      \includegraphics[width=0.19\textwidth]{appendix/q5/q5_top0.jpg}
      \includegraphics[width=0.19\textwidth]{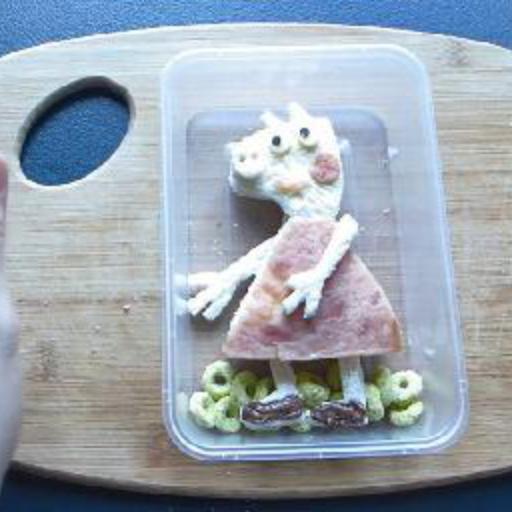}
      \includegraphics[width=0.19\textwidth]{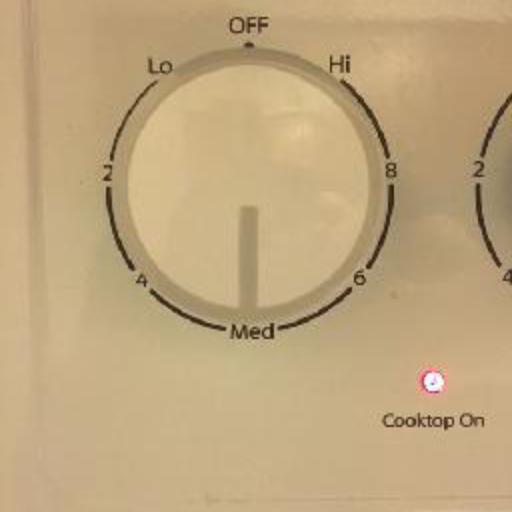}
      \includegraphics[width=0.19\textwidth]{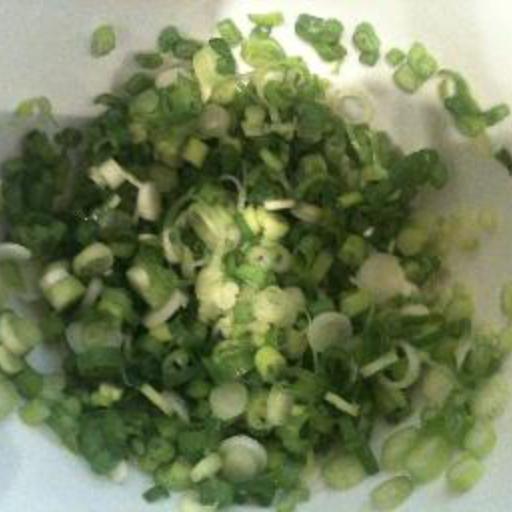}
      \includegraphics[width=0.19\textwidth]{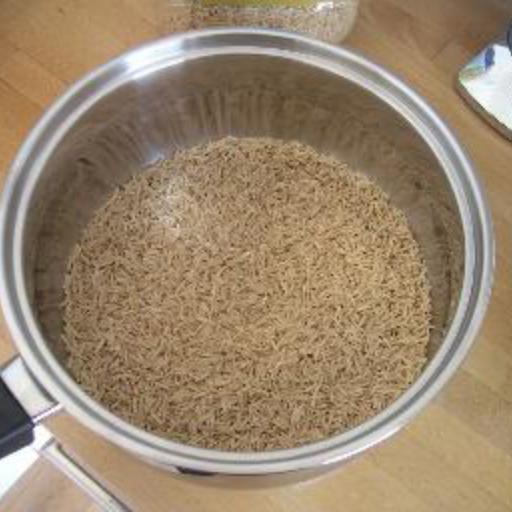}
    \end{minipage}}
    &&\multicolumn{5}{c}{
    \begin{minipage}{0.45\textwidth}
      \includegraphics[width=0.19\textwidth]{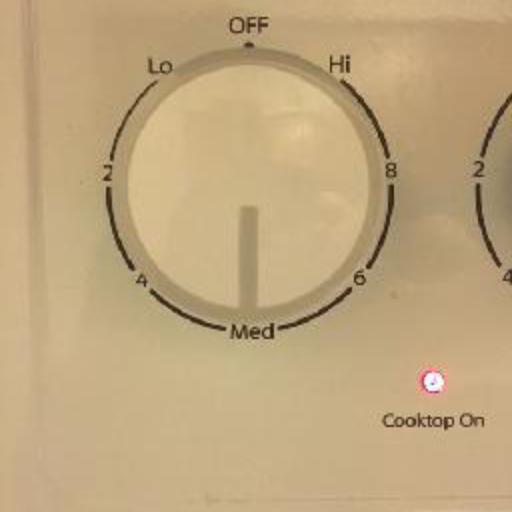}
      \includegraphics[width=0.19\textwidth]{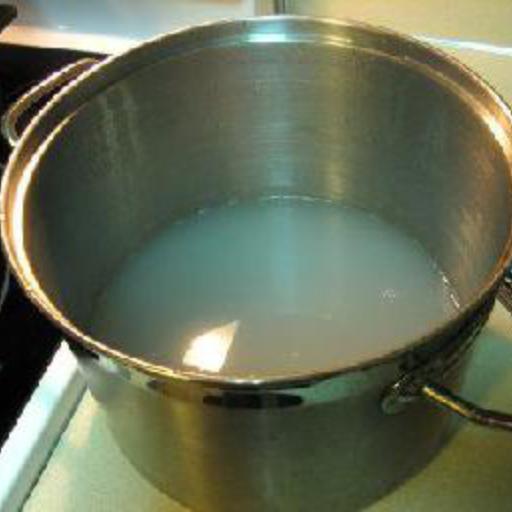}
      \includegraphics[width=0.19\textwidth]{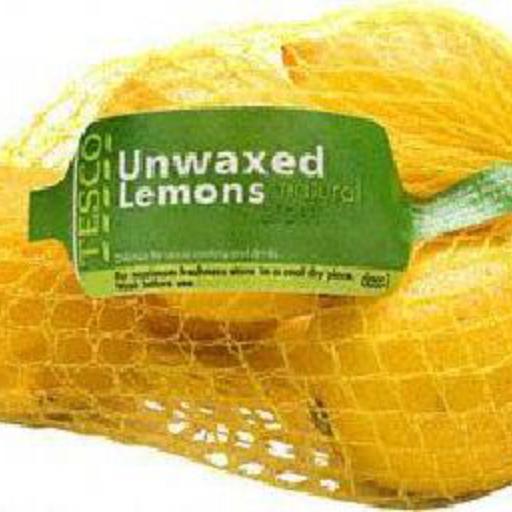}
      \includegraphics[width=0.19\textwidth]{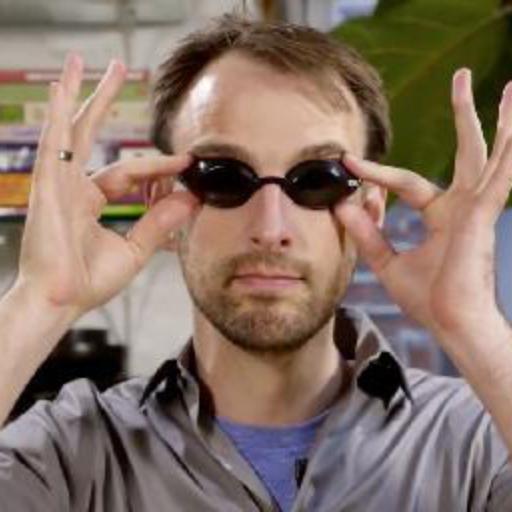}
      \includegraphics[width=0.19\textwidth]{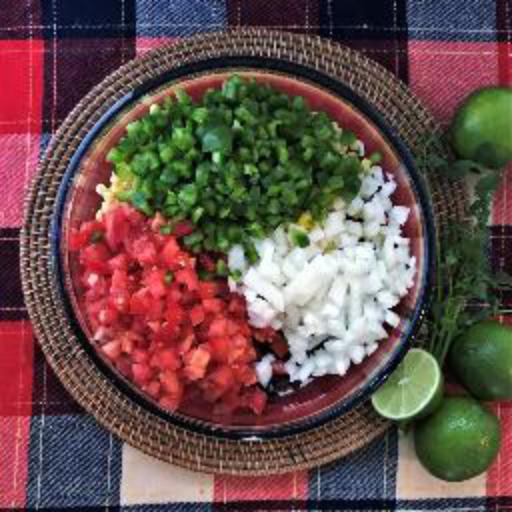}
    \end{minipage}} \\ \\
    \hline \\
     (e) \textcolor{blue}{Q6}&\multicolumn{11}{c}{A little dab will do ya. Now you have a little dab (1-2 tbsp) of coffee in the sugar. }\\
     &\multicolumn{11}{c}{Now we will make the paste. Place the coffee back on the burner to finish cooking.}\\ \\
      \begin{minipage}{0.08\textwidth}
      \includegraphics[width=\textwidth]{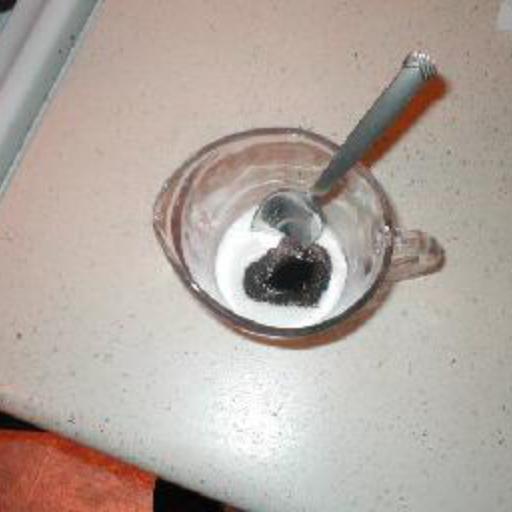}
    \end{minipage}&\multicolumn{5}{c}{
    \begin{minipage}{0.45\textwidth}
      \includegraphics[width=0.19\textwidth]{appendix/q6/q6_top0.jpg}
      \includegraphics[width=0.19\textwidth]{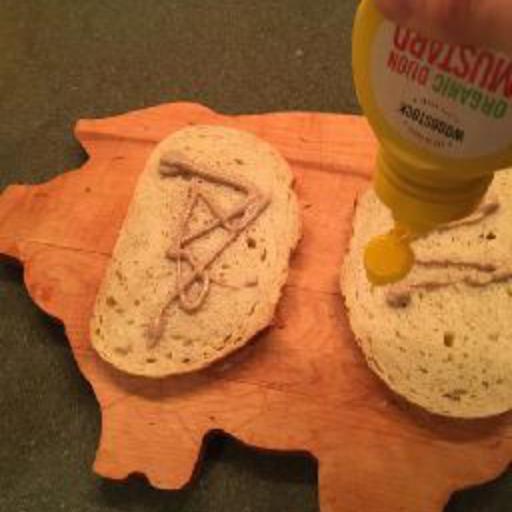}
      \includegraphics[width=0.19\textwidth]{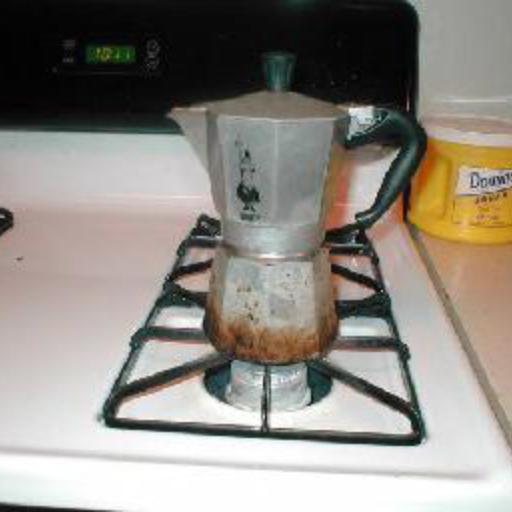}
      \includegraphics[width=0.19\textwidth]{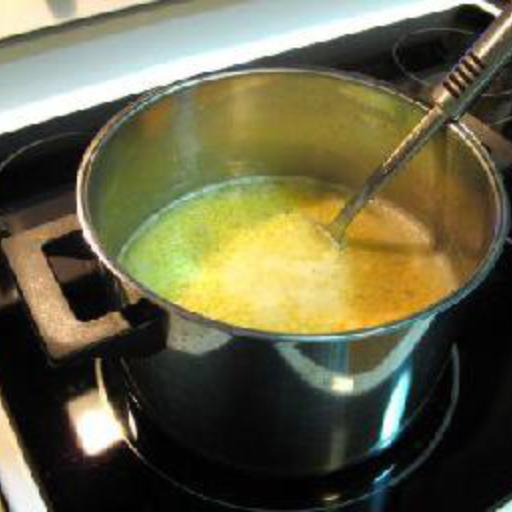}
      \includegraphics[width=0.19\textwidth]{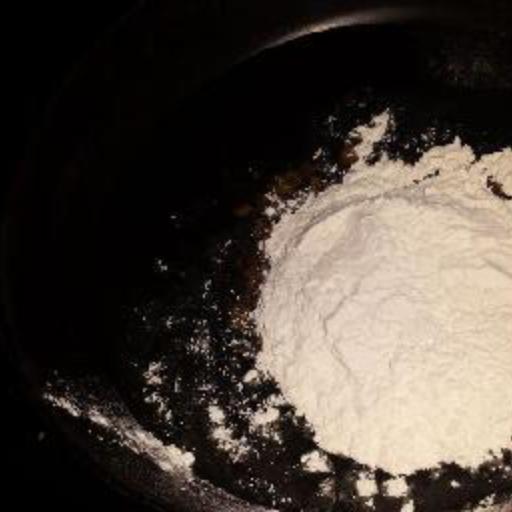}
    \end{minipage}}
    &&\multicolumn{5}{c}{
    \begin{minipage}{0.45\textwidth}
      \includegraphics[width=0.19\textwidth]{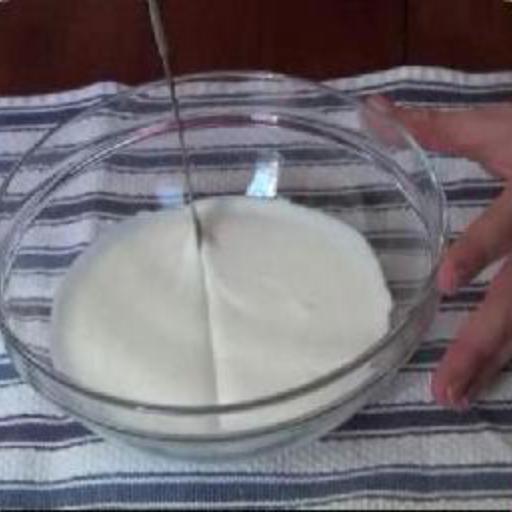}
      \includegraphics[width=0.19\textwidth]{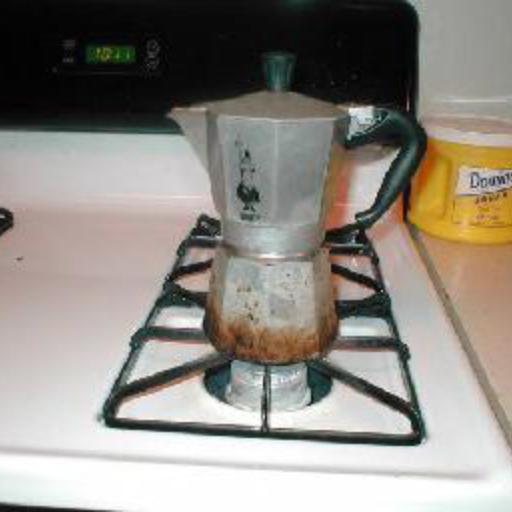}
      \includegraphics[width=0.19\textwidth]{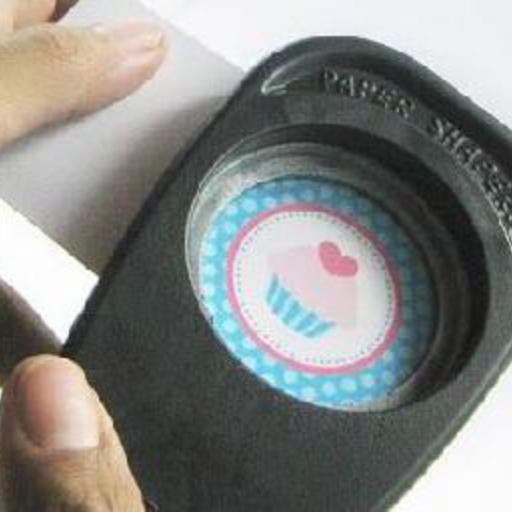}
      \includegraphics[width=0.19\textwidth]{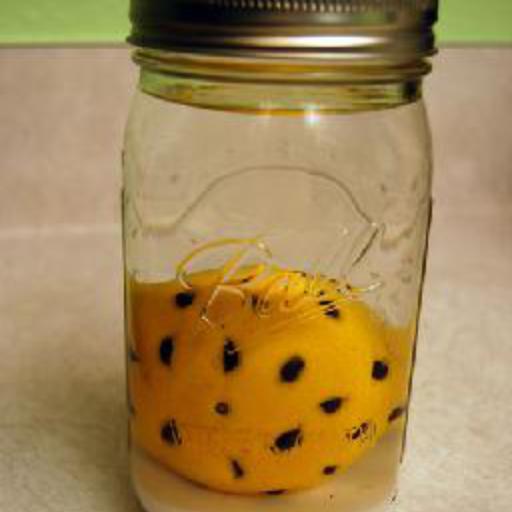}
      \includegraphics[width=0.19\textwidth]{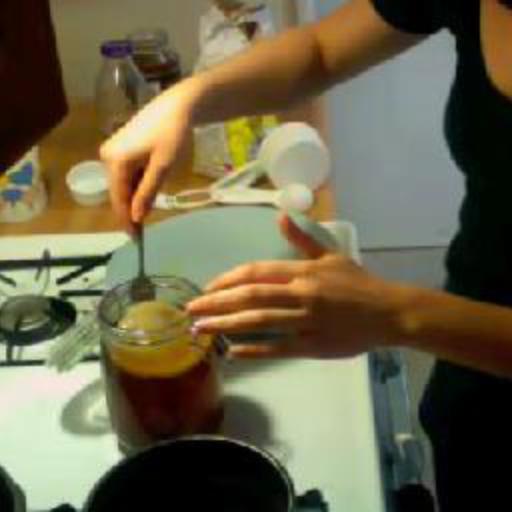}
    \end{minipage}} \\ \\
    \hline \\
     (e) \textcolor{blue}{Q7}&\multicolumn{11}{c}{After a few hours your popsicle should now be completely frozen }\\
     &\multicolumn{11}{c}{so take it out of the freezer. Now remove the popsicle from the cup. If you cannot }\\ 
    &\multicolumn{11}{c}{run the cup under hot water for a few seconds and it should slip.}\\ \\
      \begin{minipage}{0.08\textwidth}
      \includegraphics[width=\textwidth]{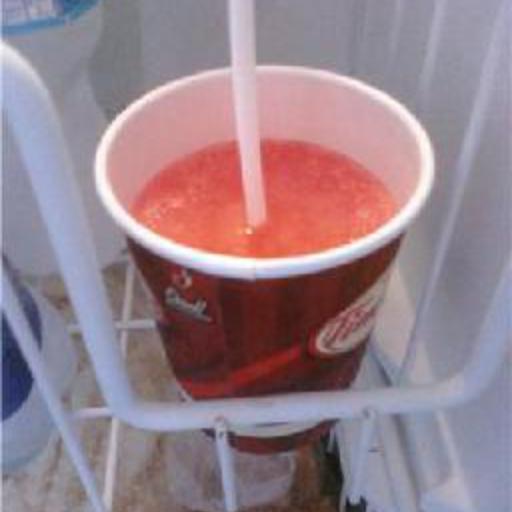}
    \end{minipage}&\multicolumn{5}{c}{
    \begin{minipage}{0.45\textwidth}
      \includegraphics[width=0.19\textwidth]{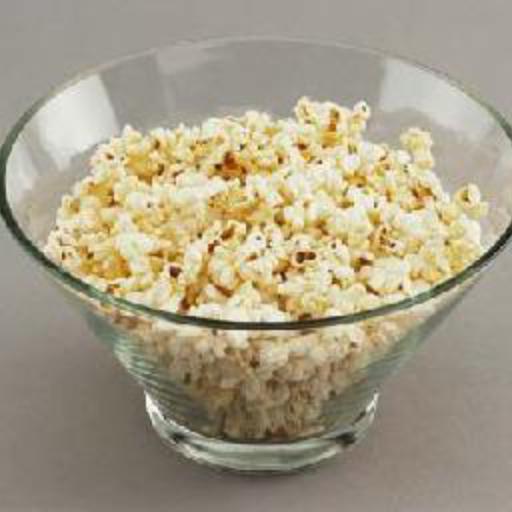}
      \includegraphics[width=0.19\textwidth]{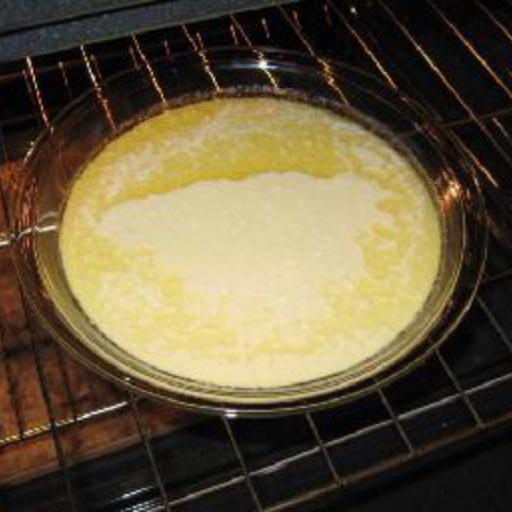}
      \includegraphics[width=0.19\textwidth]{appendix/q7/q7_top2.jpg}
      \includegraphics[width=0.19\textwidth]{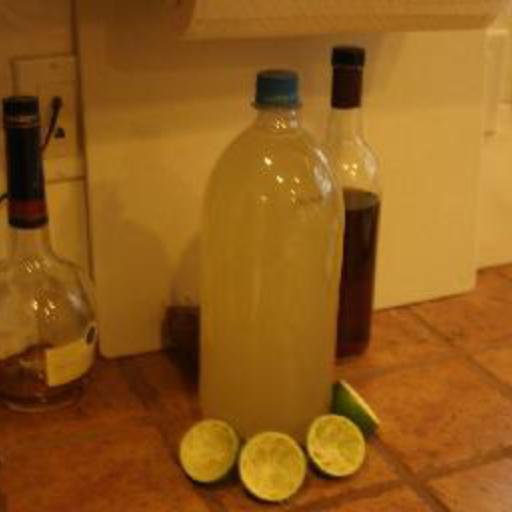}
      \includegraphics[width=0.19\textwidth]{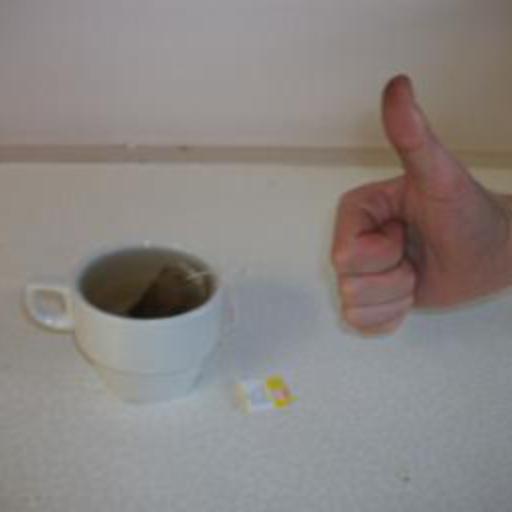}
    \end{minipage}}
    &&\multicolumn{5}{c}{
    \begin{minipage}{0.45\textwidth}
      \includegraphics[width=0.19\textwidth]{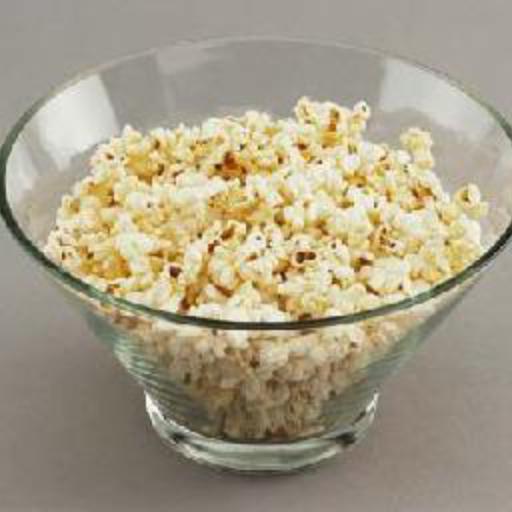}
      \includegraphics[width=0.19\textwidth]{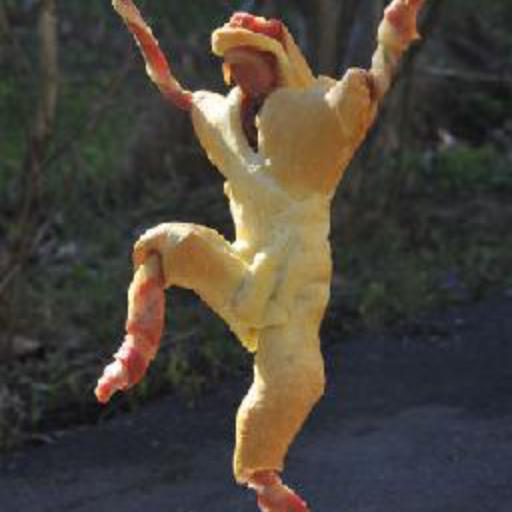}
      \includegraphics[width=0.19\textwidth]{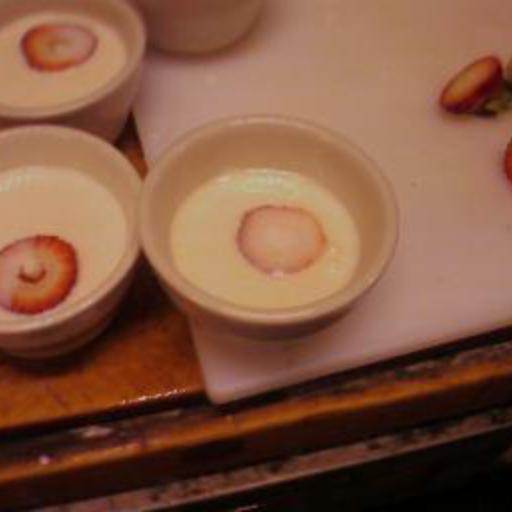}
      \includegraphics[width=0.19\textwidth]{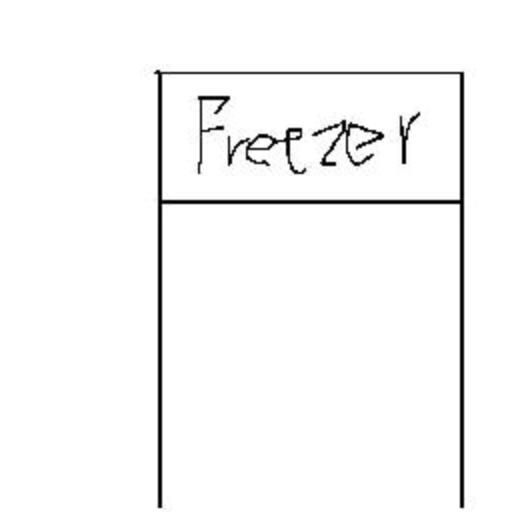}
      \includegraphics[width=0.19\textwidth]{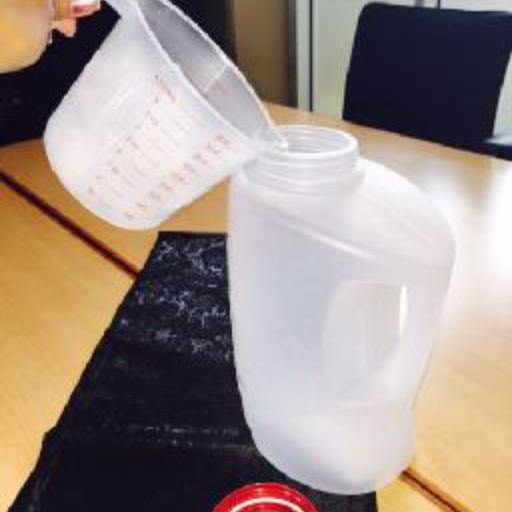}
    \end{minipage}} \\
    \end{tabular}
    \caption{Example input text, ground truth image, ground truth image-text coherence relation and the top 5 retrieved images by the proposed \texttt{CMCM} algorithm and \texttt{CMCA} for comparison on the CITE++ dataset.}
    \label{fig:appendix_cite}
    \end{figure*}

More example predictions by the proposed coherence aware and the agnostic model are given in \autoref{fig:appendix_cite}. We can see not only from the correctly retrieved images but in most cases, all the top 5 images are quite relevant to both the input text and the coherence relation. Note that the coherence relation is not shown to the model. Moreover, these are examples of when the ground truth image has the depicted relation as TRUE. Consequently in all the examples, the probability of presence of the depicted coherence relation is above $0.5$. We also provide the average precision of coherence relation prediction by the auxillary \cam in \autoref{tab:coh_classification_clue} 
\begin{table*}[ht!]
\centering
\begin{tabular}{l|llllll}
&Visible&Subjective&Action&Story&Meta&Irrelevant \\
\hline
\texttt{CMCM}-NoAttn&0.82&0.24&0.47&0.51&0.71&0.36\\
\texttt{CMCM}&0.82&0.28&0.47&0.50&0.71&0.36\\
\texttt{CMCM}$_{Visible}$ &0.83&--&--&--&--&--  \\
\texttt{CMCM}$_{Subjective}$ &--&0.18&--&--&--&-- \\
\texttt{CMCM}$_{Action}$ &--&--&0.52&--&--&-- \\
\texttt{CMCM}$_{Story}$ &--&--&--&0.55&--&-- \\
\texttt{CMCM}$_{Meta}$ &--&--&--&--&0.72&--  \\
\texttt{CMCM}$_{Irrelevant}$ &--&--&--&--&--&0.41\\
\end{tabular}
\caption{Average Precision of Coherence relation prediction using probabilities from the \cam.}
\label{tab:coh_classification_clue}
\end{table*}

\section{Clue Examples}
\begin{figure*}[h]
  \centering
  \footnotesize
  \begin{tabular}{cccccccccccc}
    %GT & Retrieved & GT & Retrieved\\ 
    GT & \multicolumn{5}{c}{\texttt{CMCM}} && \multicolumn{5}{c}{\texttt{CMCA}} \\
    \hline\\
    (a) \textcolor{blue}{Action}&\multicolumn{11}{c}{Young woman drinking coffee to go on the street.}\\ \\
    \begin{minipage}{0.08\textwidth}
      \includegraphics[width=\textwidth]{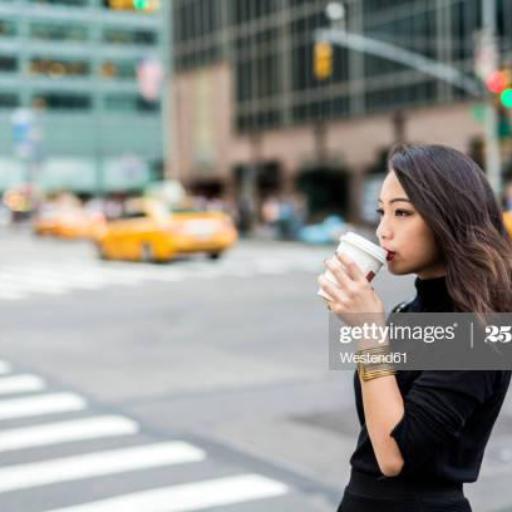}
    \end{minipage}&\multicolumn{5}{c}{
    \begin{minipage}{0.45\textwidth}
      \includegraphics[width=0.19\textwidth]{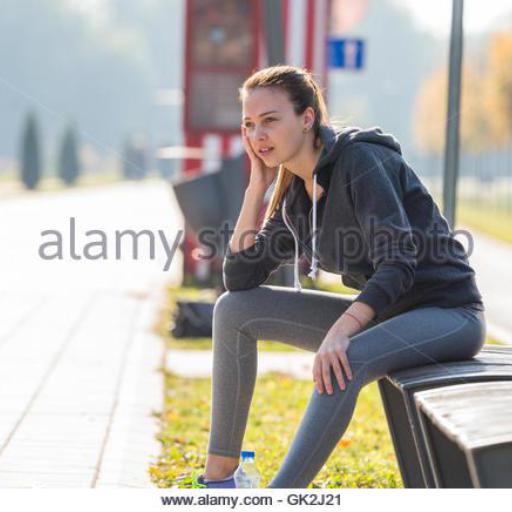}
      \includegraphics[width=0.19\textwidth]{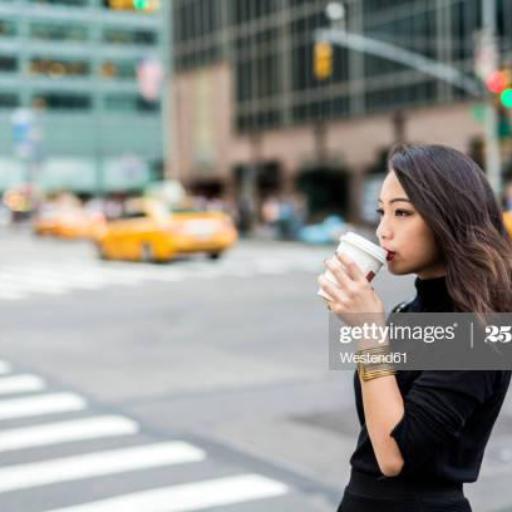}
      \includegraphics[width=0.19\textwidth]{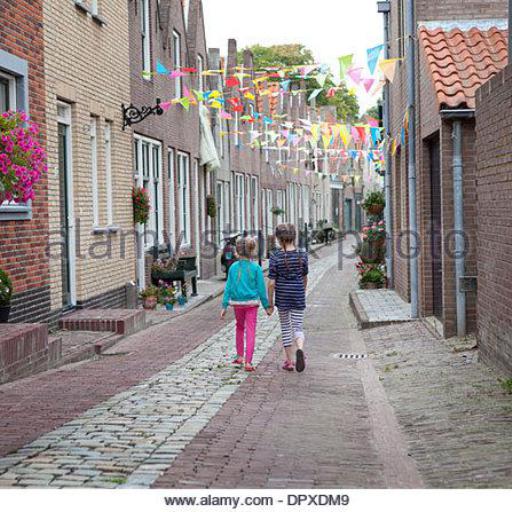}
      \includegraphics[width=0.19\textwidth]{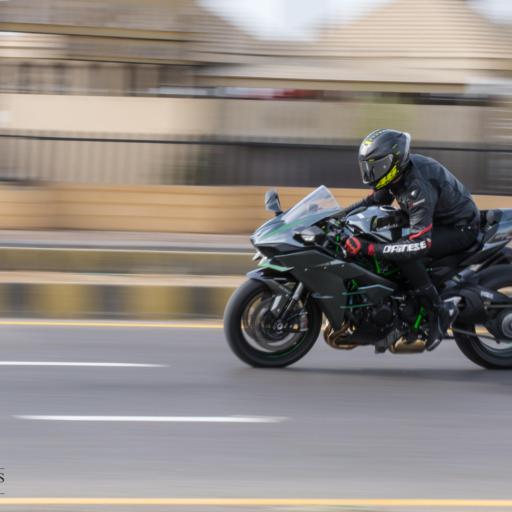}
      \includegraphics[width=0.19\textwidth]{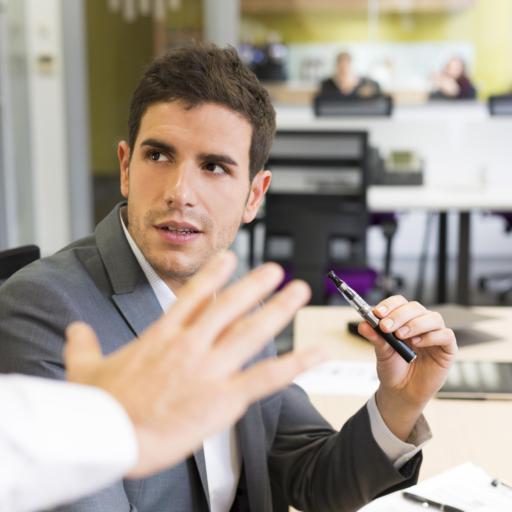}
    \end{minipage}}
    &&\multicolumn{5}{c}{
    \begin{minipage}{0.45\textwidth}
      \includegraphics[width=0.19\textwidth]{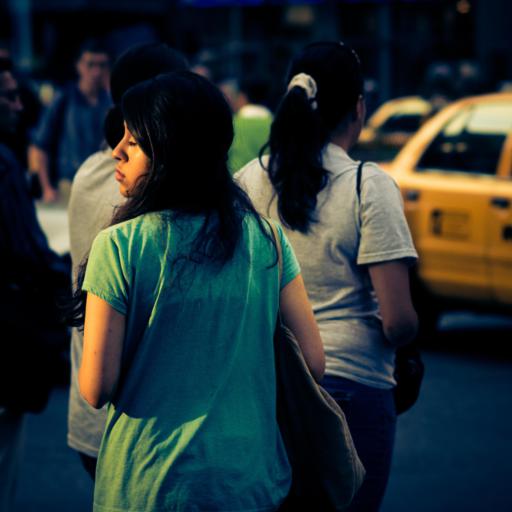}
      \includegraphics[width=0.19\textwidth]{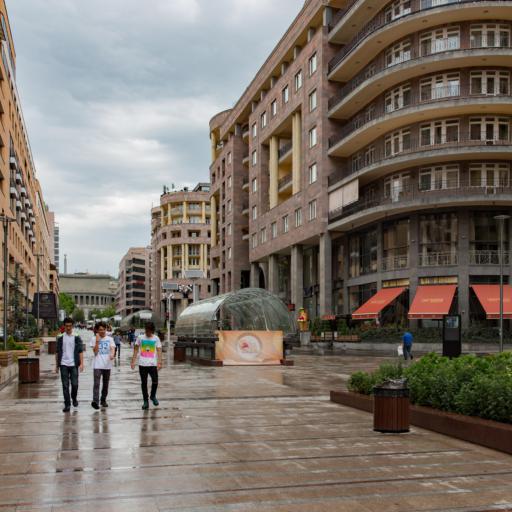}
      \includegraphics[width=0.19\textwidth]{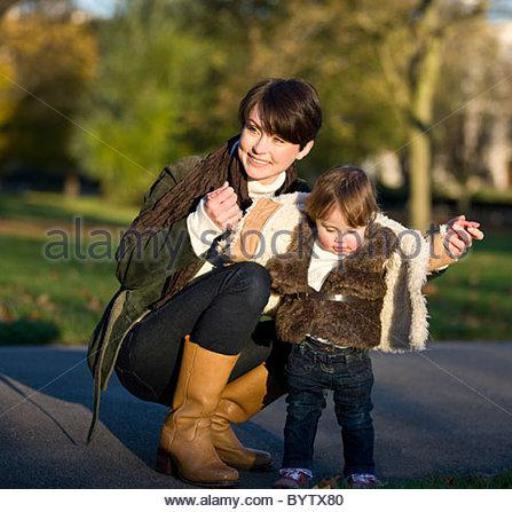}
      \includegraphics[width=0.19\textwidth]{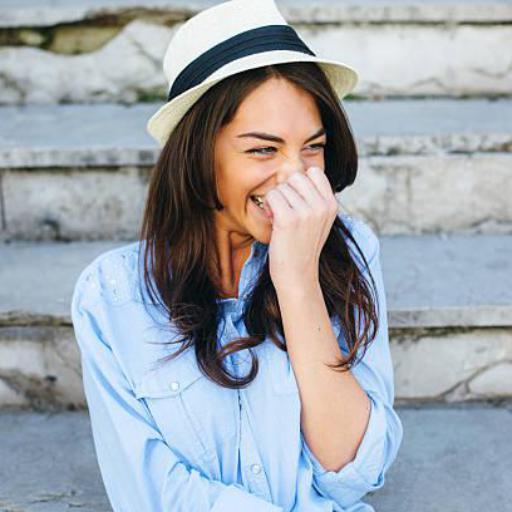}
      \includegraphics[width=0.19\textwidth]{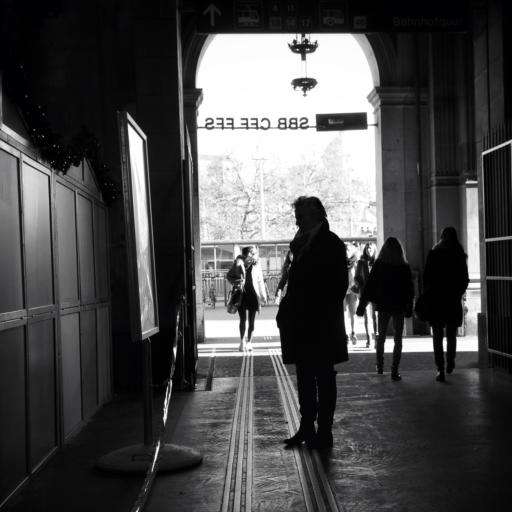}
    \end{minipage}} \\ \\
    \hline\\
    (a) \textcolor{blue}{Story}&\multicolumn{11}{c}{The flower that blooms in adversity is the most rare and beautiful of all.}\\ \\
    \begin{minipage}{0.08\textwidth}
      \includegraphics[width=\textwidth]{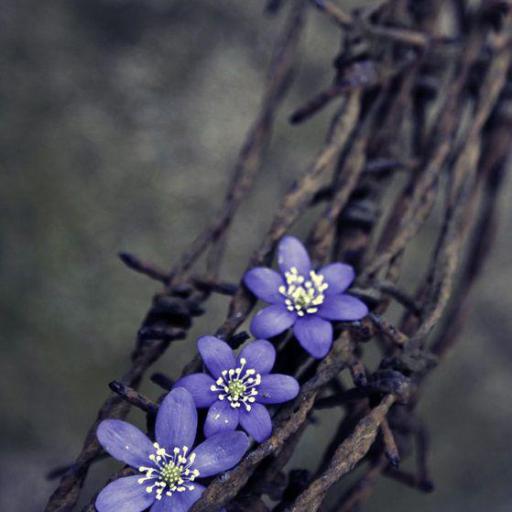}
    \end{minipage}&\multicolumn{5}{c}{
    \begin{minipage}{0.45\textwidth}
      \includegraphics[width=0.19\textwidth]{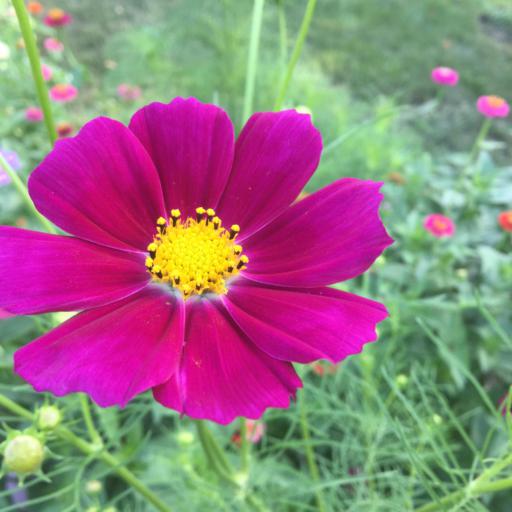}
      \includegraphics[width=0.19\textwidth]{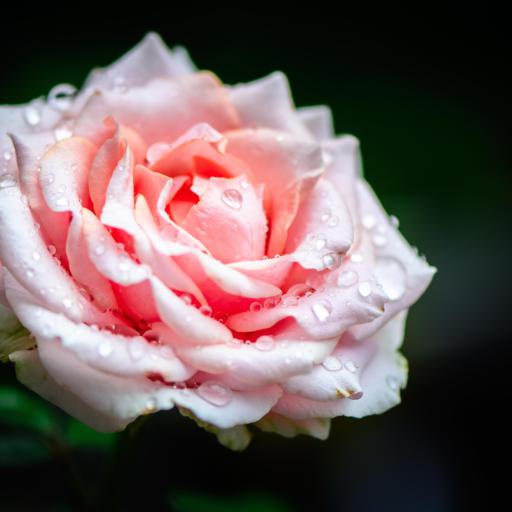}
      \includegraphics[width=0.19\textwidth]{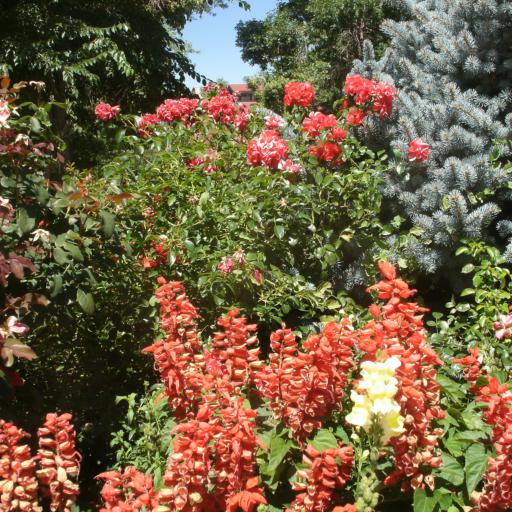}
      \includegraphics[width=0.19\textwidth]{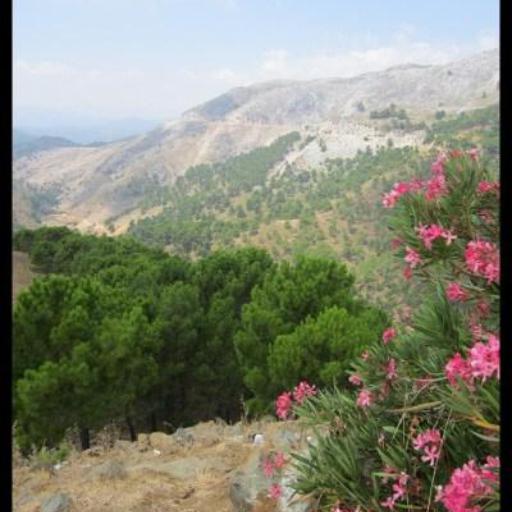}
      \includegraphics[width=0.19\textwidth]{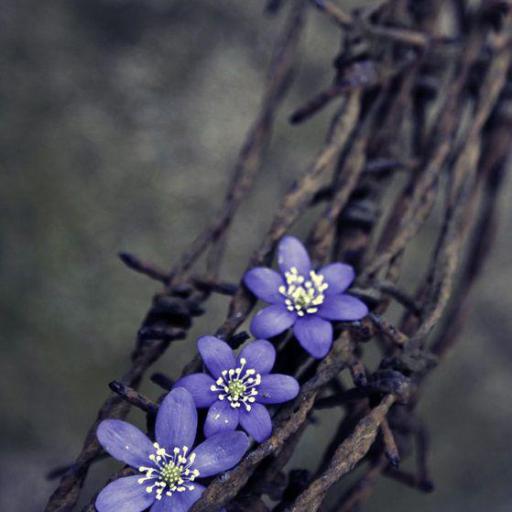}
    \end{minipage}}
    &&\multicolumn{5}{c}{
    \begin{minipage}{0.45\textwidth}
      \includegraphics[width=0.19\textwidth]{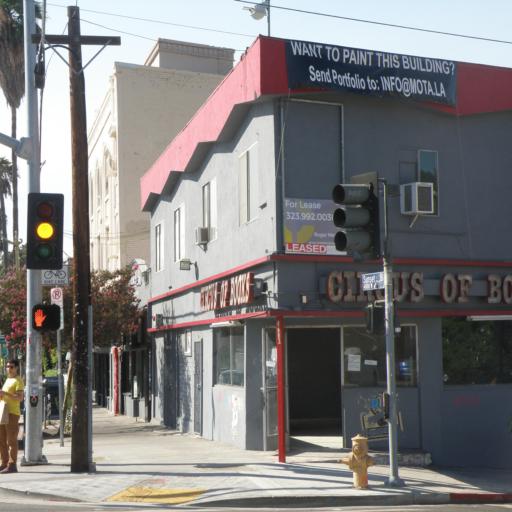}
      \includegraphics[width=0.19\textwidth]{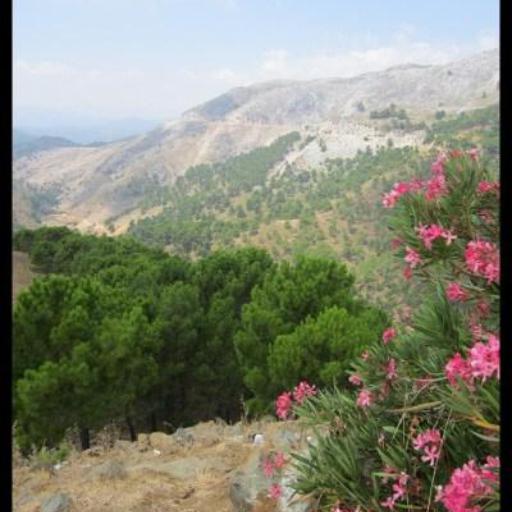}
      \includegraphics[width=0.19\textwidth]{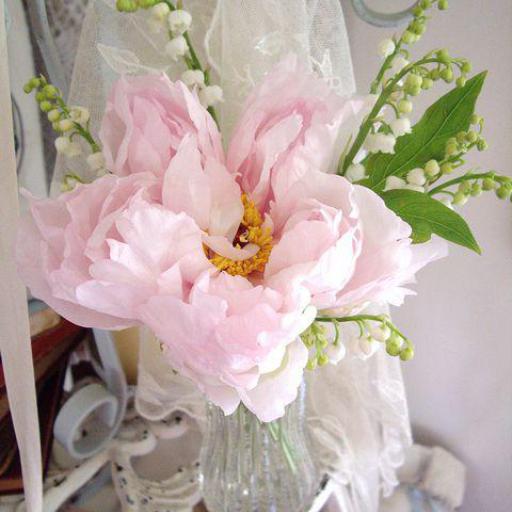}
      \includegraphics[width=0.19\textwidth]{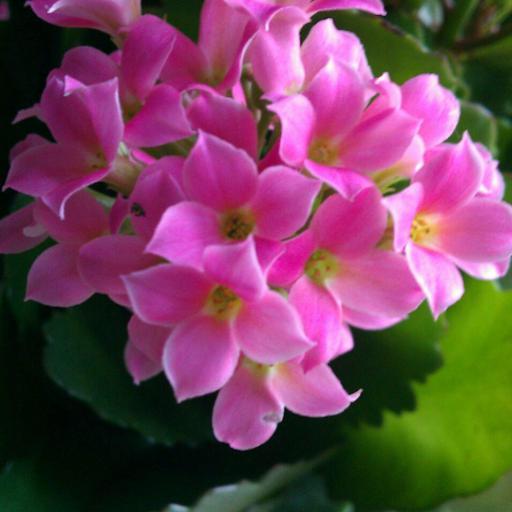}
      \includegraphics[width=0.19\textwidth]{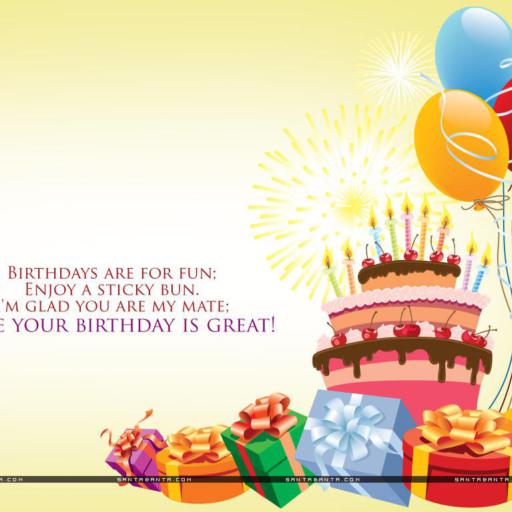}
    \end{minipage}} \\ \\
    \hline\\
    (a) \textcolor{blue}{Subjective}&\multicolumn{11}{c}{newly built small house next to the sea and the beach.}\\ \\
    \begin{minipage}{0.08\textwidth}
      \includegraphics[width=\textwidth]{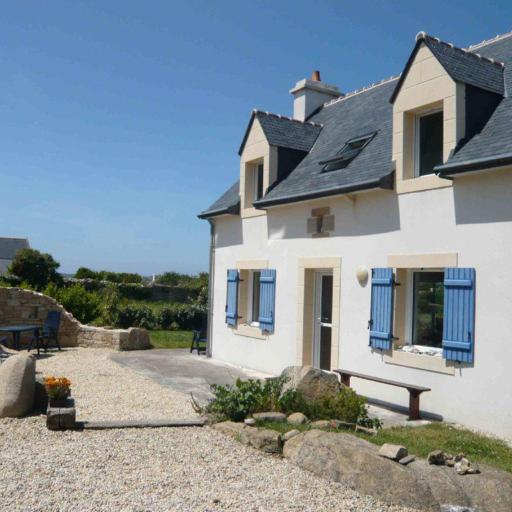}
    \end{minipage}&\multicolumn{5}{c}{
    \begin{minipage}{0.45\textwidth}
      \includegraphics[width=0.19\textwidth]{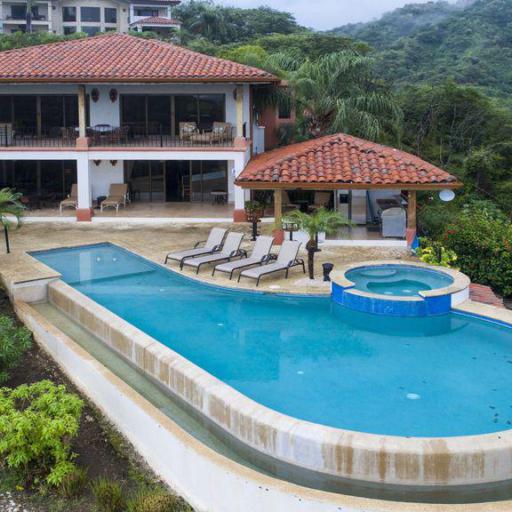}
      \includegraphics[width=0.19\textwidth]{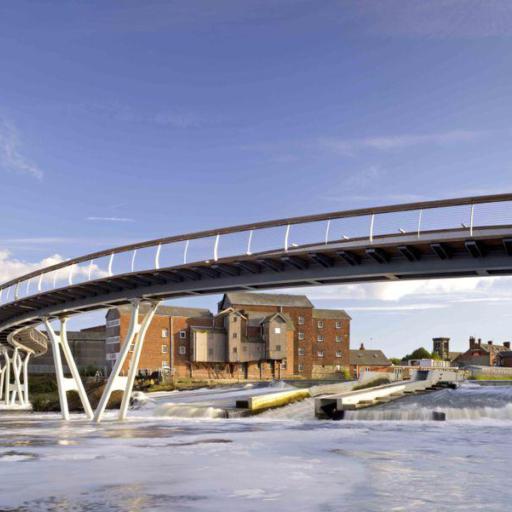}
      \includegraphics[width=0.19\textwidth]{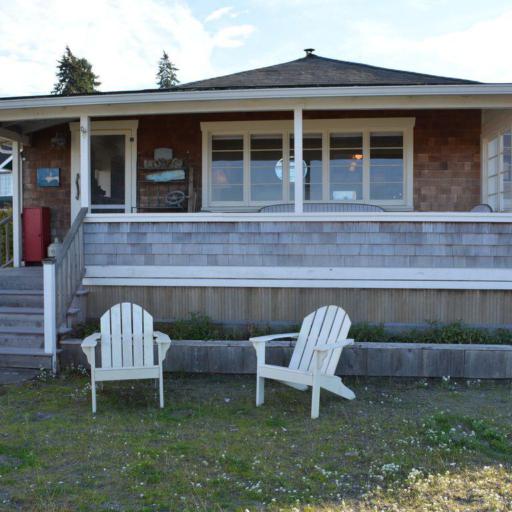}
      \includegraphics[width=0.19\textwidth]{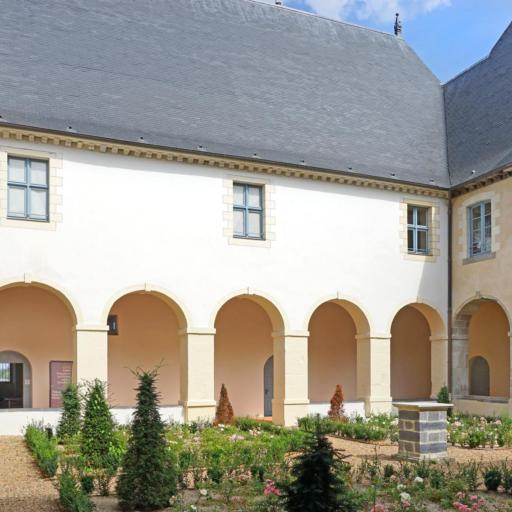}
      \includegraphics[width=0.19\textwidth]{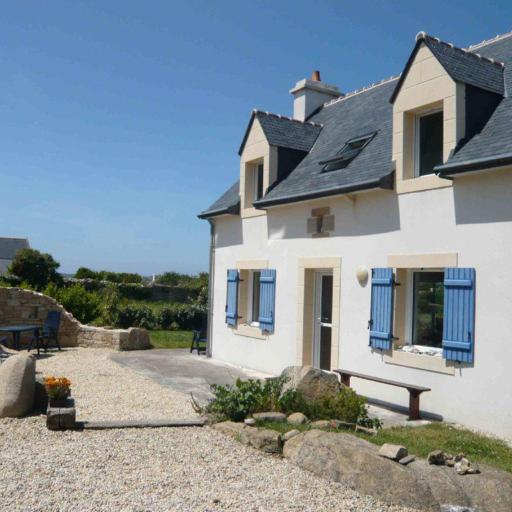}
    \end{minipage}}
    &&\multicolumn{5}{c}{
    \begin{minipage}{0.45\textwidth}
      \includegraphics[width=0.19\textwidth]{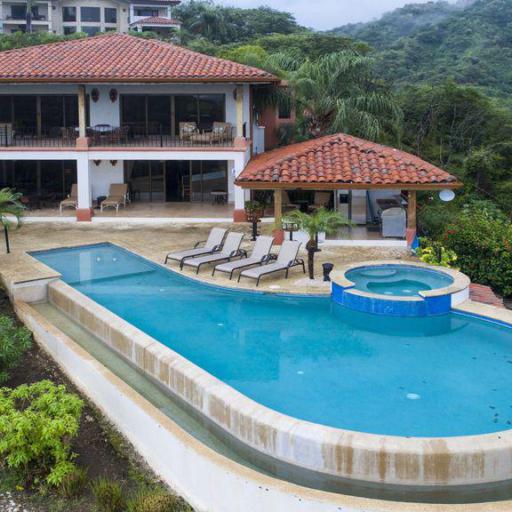}
      \includegraphics[width=0.19\textwidth]{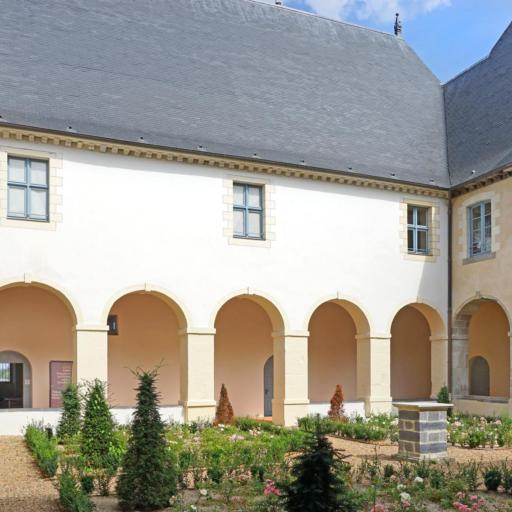}
      \includegraphics[width=0.19\textwidth]{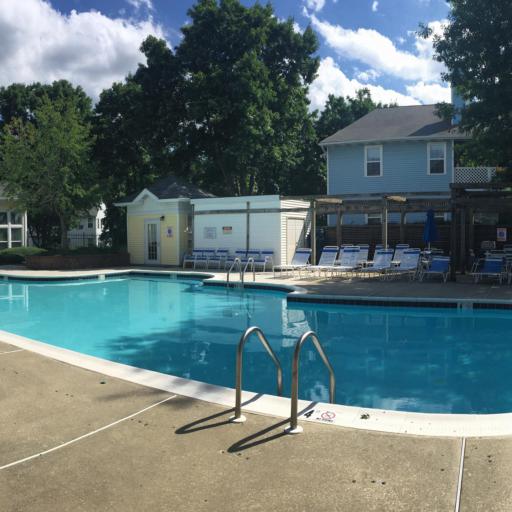}
      \includegraphics[width=0.19\textwidth]{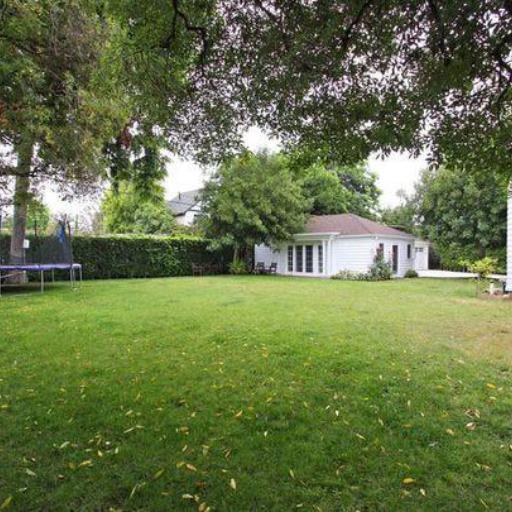}
      \includegraphics[width=0.19\textwidth]{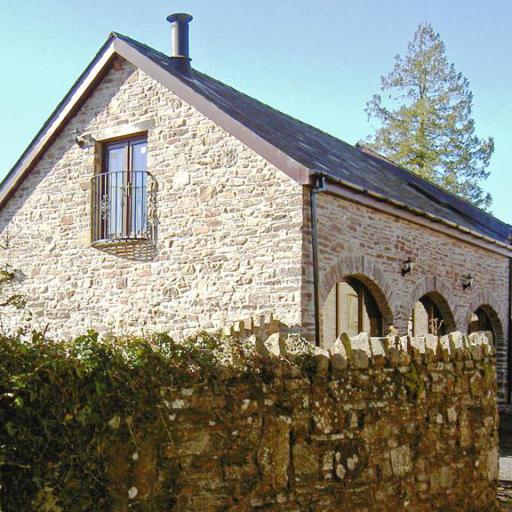}
    \end{minipage}} \\ \\
    \hline\\
    (a) \textcolor{blue}{Visible}&\multicolumn{11}{c}{Illustration of a magnifying glass over a blue background.}\\ \\
    \begin{minipage}{0.08\textwidth}
      \includegraphics[width=\textwidth]{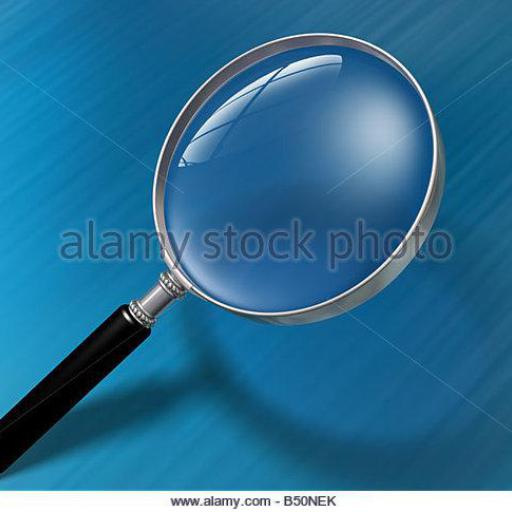}
    \end{minipage}&\multicolumn{5}{c}{
    \begin{minipage}{0.45\textwidth}
      \includegraphics[width=0.19\textwidth]{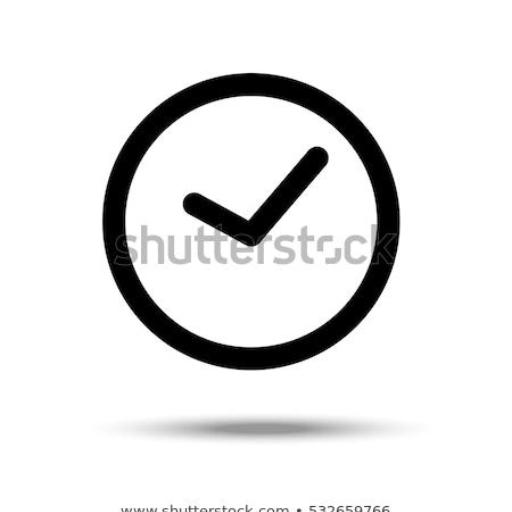}
      \includegraphics[width=0.19\textwidth]{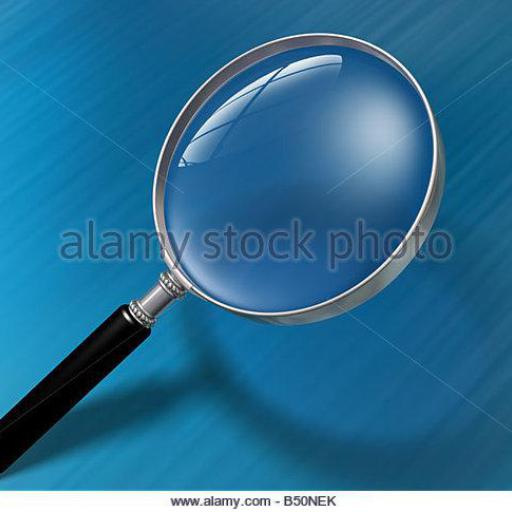}
      \includegraphics[width=0.19\textwidth]{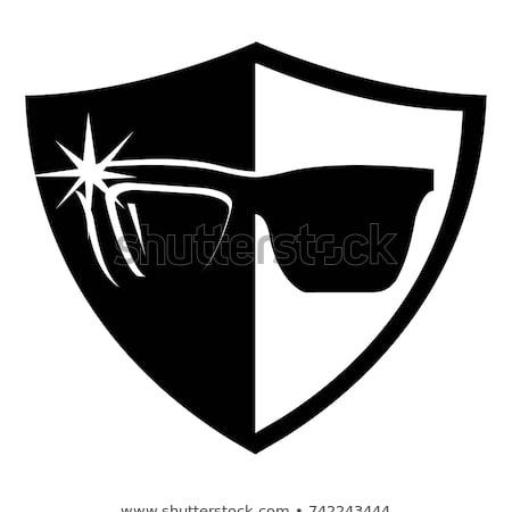}
      \includegraphics[width=0.19\textwidth]{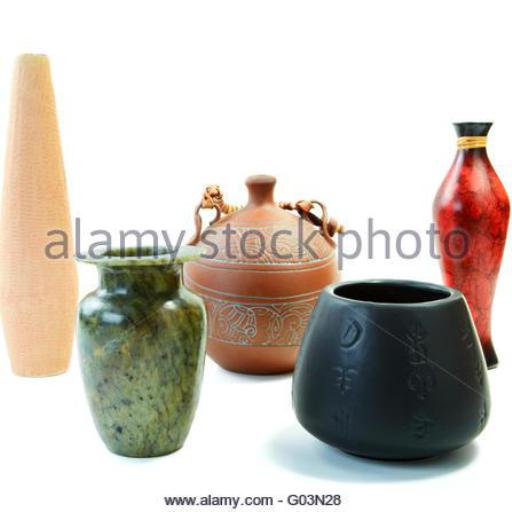}
      \includegraphics[width=0.19\textwidth]{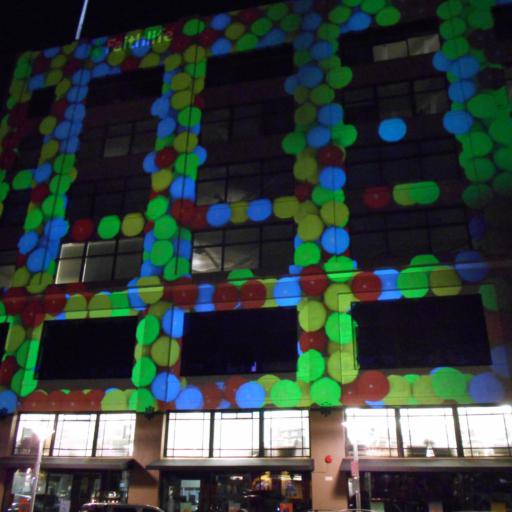}
    \end{minipage}}
    &&\multicolumn{5}{c}{
    \begin{minipage}{0.45\textwidth}
      \includegraphics[width=0.19\textwidth]{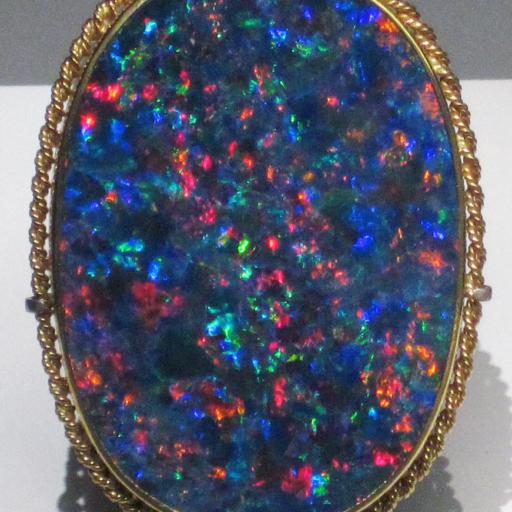}
      \includegraphics[width=0.19\textwidth]{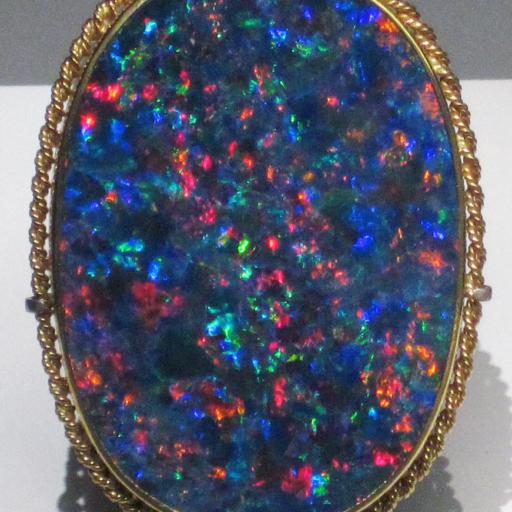}
      \includegraphics[width=0.19\textwidth]{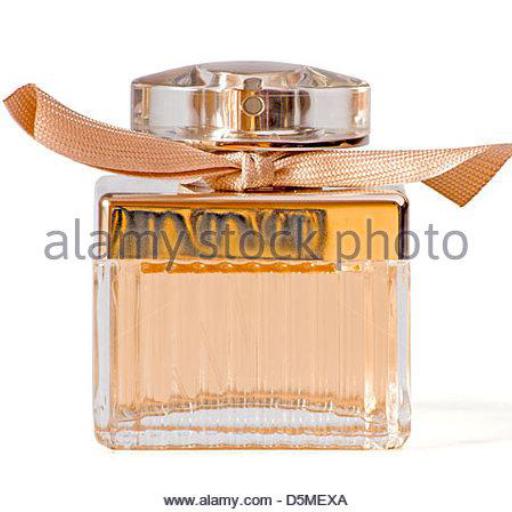}
      \includegraphics[width=0.19\textwidth]{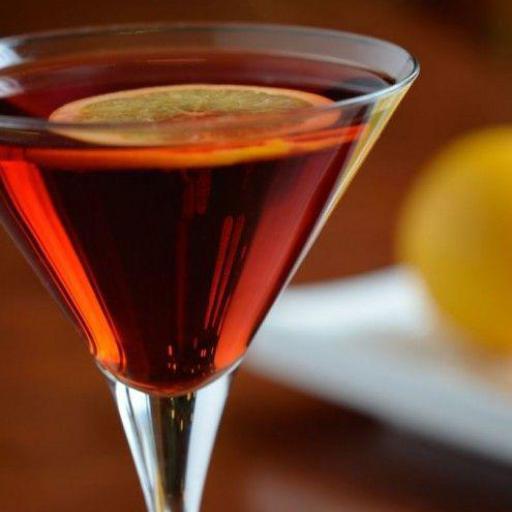}
      \includegraphics[width=0.19\textwidth]{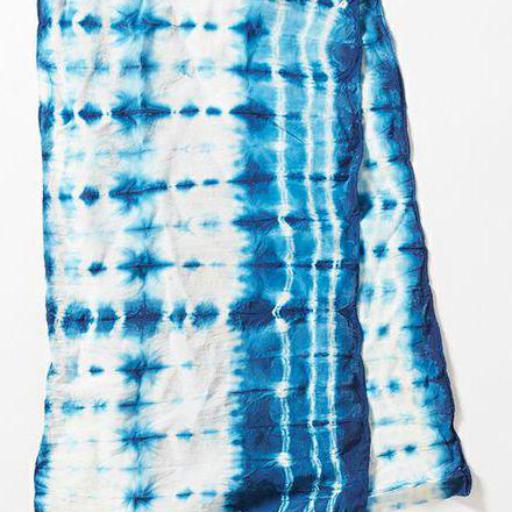}
    \end{minipage}} \\ \\
    \end{tabular}
    \caption{Example input text, ground truth image, ground truth image-text coherence relation and the top 5 retrieved images by the proposed \texttt{CMCM} algorithm and \texttt{CMCA} for comparison on the Clue dataset.}
    \label{fig:appendix_clue}
    \end{figure*}

We show more examples from the Clue dataset in \autoref{fig:appendix_clue}. The coherence aware property of the proposed model is clearly seen from the examples. For the agnostic model, we believe there exists an inherent bias towards the most frequent relation. In the Clue dataset this is the \emph{Visible} relation and hence most of the retrieved images by the agnostic model concentrates on words that can be visually grounded.
\newpage

\end{document}